\documentclass{bmvc2k}
\pdfoutput=1
\usepackage{multirow}
\usepackage{booktabs}
\usepackage{amsmath}
\usepackage{amssymb}
\usepackage{bbm}
\usepackage{wrapfig}
\usepackage{bm}

\definecolor{red}{RGB}{255, 0, 0}
\definecolor{blue}{RGB}{0, 0, 255}

\title{Semi-Supervised Object Detection with Object-wise Contrastive Learning and Regression Uncertainty}

\addauthor{Honggyu Choi}{honggyuchoi@kaist.ac.kr}{1}
\addauthor{Zhixiang Chen}{zhixiang.chen@sheffield.ac.uk}{2}
\addauthor{Xuepeng Shi}{x.shi19@imperial.ac.uk}{3}  
\addauthor{Tae-Kyun Kim}{kimtaekyun@kaist.ac.kr}{1,3}

\addinstitution{
KAIST
}
\addinstitution{
The University of Sheffield
}
\addinstitution{
Imperial College London
}

\runninghead{Choi \etal{}}{Semi-Supervised Object Detection with OCL and RUPL}

\def\etal{\emph{et al}\bmvaOneDot}

\begin{document}

\maketitle
\begin{abstract}
  Semi-supervised object detection~(SSOD) aims to boost detection performance by leveraging extra unlabeled data.
  The teacher-student framework has been shown to be promising for SSOD, in which a teacher network generates pseudo-labels for unlabeled data to assist the training of a student network.
  Since the pseudo-labels are noisy, filtering the pseudo-labels
  is crucial to exploit the potential of such framework.
  Unlike existing suboptimal methods,
  we propose a two-step pseudo-label filtering
  for the classification and regression heads in a teacher-student framework.
  For the classification head, \textbf{OCL}~(\textbf{O}bject-wise \textbf{C}ontrastive \textbf{L}earning) regularizes the object representation learning that utilizes unlabeled data to improve pseudo-label filtering by enhancing the discriminativeness of the classification score.
  This is designed to pull together objects in the same class and push away objects from different classes.
  For the regression head, we further propose \textbf{RUPL}~(\textbf{R}egression-\textbf{U}ncertainty-guided \textbf{P}seudo-\textbf{L}abeling) to learn the aleatoric uncertainty of object localization for label filtering.
  By jointly filtering the pseudo-labels for the classification and regression heads, the student network receives better guidance from the teacher network for object detection task. 
  Experimental results on Pascal VOC and MS-COCO datasets demonstrate the superiority of our proposed method with competitive performance compared to existing methods.

\end{abstract}
\section{Introduction}
\label{sec:intro}
Semi-Supervised Object Detection~(SSOD) leverages both labeled data and extra unlabeled data to learn object detectors.
Previous works~\cite{STAC,NEURIPS2019_csd,instant_teaching,yang2021interactive,liu2021unbiased, humbleteacher,wang2021data,rethink,adaptiveclass,zheng2022dual,combatnoise,xu2021soft, label_matching,activeteacher,scale-equivalent,denselearning,MUM} have proposed various methods to exploit the unlabeled data. Among them, one promising solution is to generate pseudo-labels for the unlabeled data by a pre-trained model and then use them as labeled data along with the original labeled data to train the target object detector.
By applying different augmentations to an image, the consistency between the labels of them can serve as extra knowledge to train the target model.
The consistency can be exploited by using the generated pseudo-labels from weakly augmented images to regularize the model's predictions of strongly augmented images~\cite{STAC}.
A teacher-student framework~\cite{meanteacher} further learns the pseudo-labels by mutual learning teacher and student networks, where the teacher network generates pseudo-labels of unlabeled data to assist the training of the student network, and the student network transfers the updated knowledge to the weights of the teacher network.

\begin{figure}[t]
\begin{tabular}{cc}
\qquad
\qquad
\bmvaHangBox{\includegraphics[trim={0cm 4.0cm 0cm 0},clip,width=0.35\columnwidth]{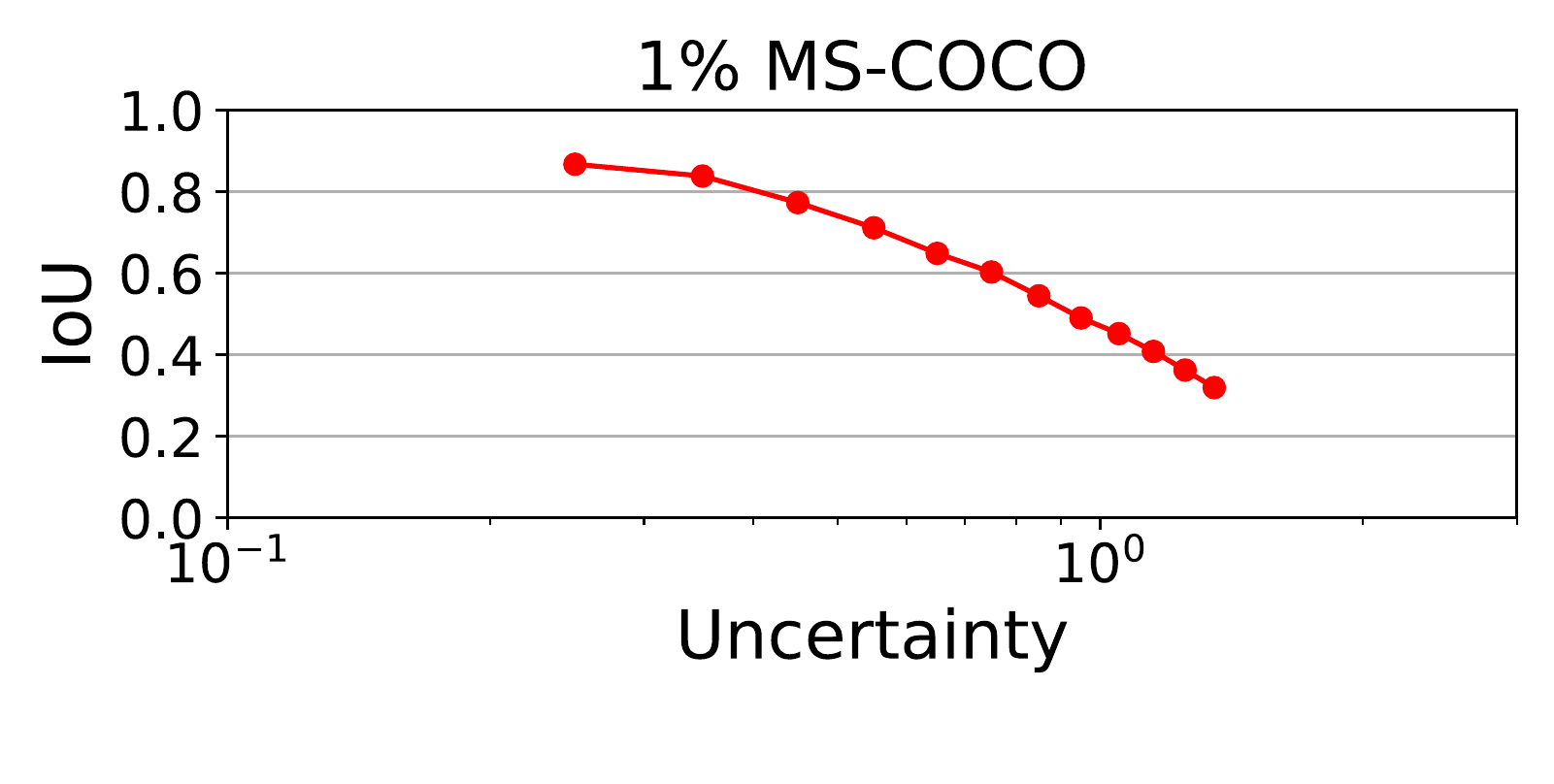}}&
\qquad
\bmvaHangBox{\includegraphics[trim={0cm 4.0cm 0cm 0},clip,width=0.35\columnwidth]{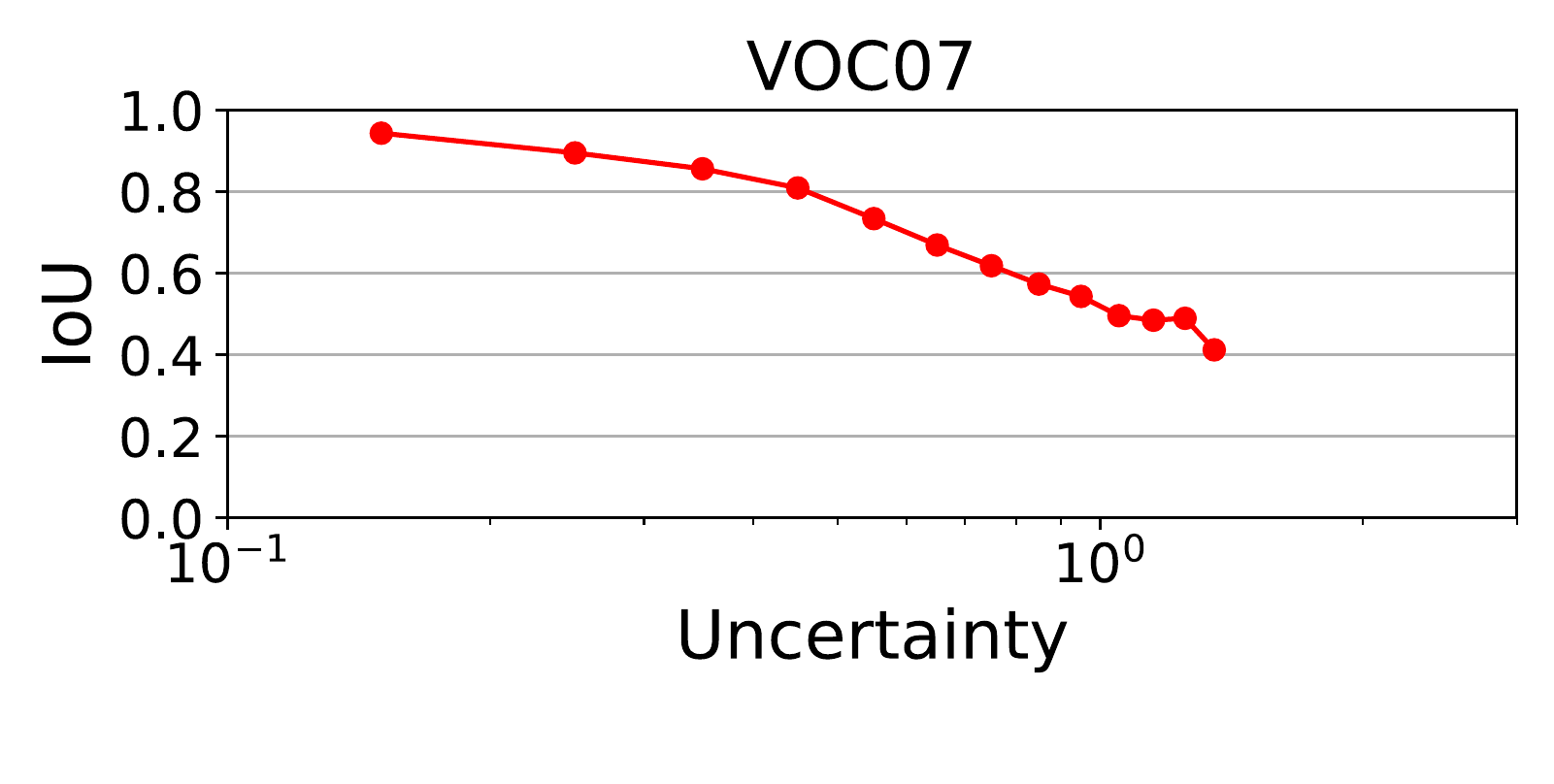}}
\\
\end{tabular}
\caption{The correlation between the regression uncertainties and the ground-truth IoUs of bounding boxes. The ground-truth IoU represents the localization quality. Low uncertainty suggests accurate localization of the bounding box.
}
\label{fig:iou_uncertainty}
\end{figure}

To train the detector with the pseudo-labels, the quality of the generated pseudo-labels is critical to the detection performance.
For the detection task, the pseudo-labels are defined for each region of interest and include the class labels and bounding boxes.
Hence, the qualities of both the classification and the localization are important to filter out unreliable pseudo-labels.
The classification score is widely adopted to select region proposals with confidence higher than a pre-defined threshold. The selected labeled regions are used to train both the classification and regression heads in the detector.
There are some heuristic designs to measure the localization quality of the generated pseudo-labels, e.g., prediction consistency~\cite{xu2021soft}, interval classification uncertainty~\cite{rethink}.
While these techniques together with the classification score are effective, we argue the existing classification scores and localization optimization is suboptimal, since the lack of labeled data makes classification score less discriminative, and the measurement of pseudo labels' localization quality is less investigated. 
The classification scores mainly distill knowledge from labeled data to evaluate the quality of unlabeled data, which under exploits the unlabeled data. 
To tackle these challenges, we extend existing methods in self-supervised representation learning to SSOD task. Instead of using heuristic designs, we further adopt uncertainty to measure the localization quality, which turns out to be an effective indicator as shown in Fig.~\ref{fig:iou_uncertainty}.

In this paper, we propose an effective method to generate pseudo-labels for semi-supervis-ed object detection. 
Specifically, we present a two-step pseudo-label filtering for the classification and regression heads in a teacher-student framework. For the classification head, the classification score is used to filter out unreliable pseudo class labels. 
We aim to improve the classification filtering through enhancing the capacity of the classification branch by taking into consideration of the unlabeled data. We introduce an object-wise contrastive learning (\textbf{OCL}) loss for the feature extractor in the classification head. This contrastive loss is defined on object regions with semantic similar regions defined as positive pairs. This loss pull together object representations of the positive pairs and push away representations of negative pairs. 
For the regression head, we propose to use uncertainty as the indicator to filter out pseudo-labels with regression-uncertainty-guided pseudo-labeling (\textbf{RUPL}). We design an uncertainty head in parallel to the classification and regression heads to learn the aleatoric uncertainty for bounding boxes. We only select pseudo-labels with low uncertainties to train our models.
We combine OCL and RUPL in our framework to improve the quality of pseudo-labels for the detection task.
To demonstrate effectiveness of our framework, we perform the experiments on standard object detection benchmark: PASCAL VOC~\cite{voc} and MS-COCO~\cite{mscoco}. We experimentally show that OCL and RUPL are complementary and have a synergetic effect. Our framework also achieves competitive performance compared to existing methods, without geometry or improved augmentations.

Our contributions are summarized as follows: 
(1) We propose an integrated framework that addresses filtering pseudo-labels for classification and regression in SSOD. 
(2) We propose OCL that improves the discriminativeness of the classification score to enhance the filtering the pseudo-labels for classification.
(3) We propose RUPL that models bounding box localization quality via  uncertainty and removes misplaced pseudo-labels for regression.
(4) We experimentally show the synergetic effect between OCL and RUPL and demonstrate the effectiveness of the overall framework on standard object detection benchmarks.

\section{Related Work}
\label{sec:related_work}
\noindent\textbf{Semi-Supervised Object Detection.}
SSOD methods can be divided into two categories: pseudo-labeling~\cite{STAC, wang2021data} methods and consistency regularization~\cite{NEURIPS2019_csd, ISD, proposal_learning} methods. Pseudo-labeling methods regularize the model using predictions generated from the pre-trained model utilizing unlabeled data.
Recent pseudo-labeling-based works~\cite{liu2021unbiased, instant_teaching, humbleteacher, yang2021interactive} adopt the teacher-student framework~\cite{meanteacher}. In this framework, a teacher's predictions guide a student and the student weight parameters evolve the teacher, which leads to remarkable performance improvement. Since pseudo-labels may have noise that degrades the model performance, classification scores are used to eliminate unreliable ones. To improve the pseudo-label quality, IT~\cite{instant_teaching} and ISTM~\cite{yang2021interactive} propose ensemble-based method, and ACRST~\cite{adaptiveclass} adopts a multi-label classifier at the image level to use high level information. Furthermore, to eliminate misaligned pseudo-labels, 3DIoUMatch~\cite{wang20213dioumatch} adopts IoU prediction~\cite{iounet} in semi-supervised 3D object detection, and ST~\cite{xu2021soft} utilizes regression prediction consistency, and RPL~\cite{rethink} reformulates regression as classification. Concurrent work~\cite{ubteacherv2} adopts uncertainty estimation and guides the student if pseudo-labels have lower uncertainty than the student's predictions. In contrast to existing works, we propose an integrated framework that improves pseudo-label quality by improving the discriminativeness of object-wise features and modeling regression uncertainty. 

\noindent\textbf{Contrastive Learning.}
Contrastive Learning~(CL) decreases the distance between positive paired samples and increases the distance between negative ones. Representation learning~\cite{mocov3, simclr, infonce} has succeeded through self-supervised CL that treats different views of the same image as a positive pair. By extension, SupCon~\cite{khosla2020supcont} proposes supervised CL that makes a positive pair for images from the same class. Recent works~\cite{class_aware, pointcontrasitve} apply CL that makes positive pairs if unlabeled images or points have the same predicted class. In SSOD, PL~\cite{proposal_learning} adopts CL that treats overlapped region proposals as positive pairs and otherwise as negative pairs even if they have the same category, which hinders the representation learning. By contrast, in SSOD, we first introduce CL that leverages predictions to identify object regions from unlabeled images and makes positive pairs for objects from the same class to improve the discriminativeness.

\noindent\textbf{
Uncertainty Estimation.}
A seminal work~\cite{whatUncertainty} captures epistemic and aleatoric uncertainties in the deep learning frameworks of computer vision. Recent works~\cite{box_regression_uncertainty, gaussianyolo} employ aleatoric uncertainty in bounding box regression to identify well-localized bounding boxes. UPS~\cite{rizve2021in} applies aleatoric uncertainty in semi-supervised image classification to remove noisy pseudo-labels. In contrast, our work adopts aleatoric uncertainty in SSOD to model the localization quality of pseudo-labels and select reliable pseudo-labels for regression.

\section{Proposed Method}
\label{sec:methods}
We first define the problem, then show the overall framework in Sec.~\ref{sec:overview}.
Finally we detail the proposed OCL in Sec.~\ref{sec:contrastive}
and RUPL in Sec.~\ref{sec:uncertainty}.

\begin{figure}[t]
    \centering
    \includegraphics[trim=0cm 7cm 10cm 2cm, clip=true, width=\columnwidth]{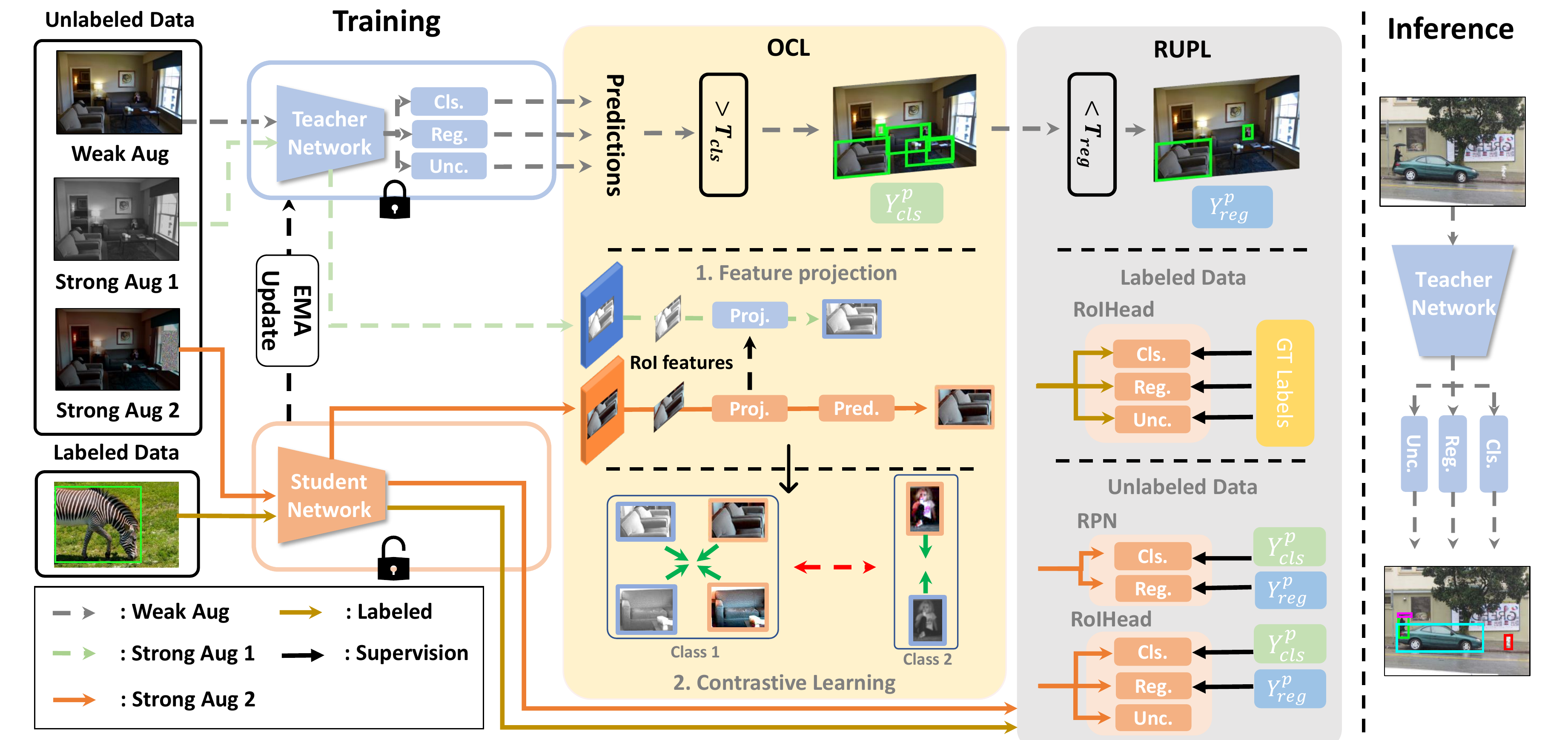}
    \caption{The overview of our proposed framework.
    }
    \label{fig:overview}
\end{figure}

\subsection{Problem Definition}
Compared with supervised object detection, semi-supervised object detection aims to utilize additional unlabeled images to improve the object detection accuracy. In the teacher-student framework, our task is to generate pseudo-labels from unlabeled data using the teacher network for to train the student network.  Specifically, there are a labeled dataset $\{\bm{X}^l,\bm{Y}^l\}=\{(\bm{x}^l_i, \bm{y}^l_i)\}^{N_l}_{i=1}$ and an unlabeled dataset $\bm{X}^u =\{\bm{x}^u_i\}^{N_u}_{i=1}$, where $N_l$, $N_u$, $\bm{x}^l$, $\bm{y}^l$ and $\bm{x}^u$  denote the size of labeled dataset, the size of unlabeled dataset, labeled images, ground-truth labels, and unlabeled images, respectively. The ground-truth label~$\bm{y}^l$ consists of each object's class category~$c^l$ and bounding box $b^l$.
The goal of pseudo-label filtering is to generate $\{\bm{X}^{p}_{cls},\bm{Y}^{p}_{cls}\}$ and  $\{\bm{X}^{p}_{reg},\bm{Y}^{p}_{reg}\}$ from $\bm{X}^u$ to train the classification and regression heads in the detection model, respectively.

\subsection{Framework Overview}
\label{sec:overview}
To utilize the unlabeled data in
semi-supervised object detection, we propose object-wise contrastive learning (OCL) and regression-uncertainty-guided pseudo-labeling (RUPL) to select reliable pseudo-labels for the training of classification and regression, respectively. 
We apply our proposed modules to a simple baseline~\cite{liu2021unbiased}. This baseline utilizes the teacher-student framework and pseudo-labeling, which are commonly adopted in SSOD frameworks. Fig.~\ref{fig:overview} shows our proposed framework with OCL and RUPL. \\
\textbf{Training stage}. 
The teacher-student framework has a student network and a teacher network.
At the pre-training stage, the student network is trained with a supervised loss (Eq.~\ref{eq:sup}) using the labeled dataset $\{\bm{X}^l,\bm{Y}^l\}$ .
\begin{equation}
  \mathcal{L}_{sup} = \mathcal{L}_{rpn,cls} + \mathcal{L}_{rpn,reg} + \mathcal{L}_{roi,cls} + \mathcal{L}_{roi,reg}
    \label{eq:sup}
\end{equation}
The supervised loss consists of classification and regression losses for each RPN and ROI.
Specifically, we follow~\cite{liu2021unbiased} to use cross-entropy loss for $\mathcal{L}_{rpn,cls}$, smooth L1 loss~\cite{girshick2015fast} for $\mathcal{L}_{rpn,reg}$, focal loss~\cite{focalloss} for $\mathcal{L}_{roi,cls}$. We adopt uncertainty-aware regression loss~\cite{whatUncertainty} for $\mathcal{L}_{roi,reg}$.
At the mutual learning stage, the teacher network is initialized by the pre-trained student network.
Both labeled and unlabeled images are used at this stage. The teacher network generates the pseudo-labels of unlabeled images to train the student network. The overall loss of the student network is
\begin{equation}
    \mathcal{L} = \mathcal{L}_{sup} + \lambda_{unsup} \mathcal{L}_{unsup} + \lambda_{ocl} \mathcal{L}_{ocl},
    \label{eq:semi-sup}
\end{equation}
where $\mathcal{L}_{unsup}$ and $\mathcal{L}_{ocl}$ denote the unsupervised loss and object-wise contrastive loss, respectively.
The unsupervised and object-wise contrastive losses are computed on unlabeled images.
Our unsupervised loss consists of similar loss terms as the supervised loss. 
Here, $\mathcal{L}_{rpn,cls}$ and $\mathcal{L}_{roi,cls}$ are computed on the set of pseudo-labels for classification $\{\bm{X}^p_{cls}, \bm{Y}^p_{cls}\}$, $\mathcal{L}_{rpn,reg}$ and $\mathcal{L}_{roi,reg}$ are computed on the set of pseudo-labels for regression $\{\bm{X}^p_{reg}, \bm{Y}^p_{reg}\}$. 
And we use smoothL1 loss for $\mathcal{L}_{roi,reg}$.
The teacher network is iteratively updated by the student network's weight parameters via exponential moving average~\cite{liu2021unbiased}.
\\
\textbf{Inference Stage}. The teacher
network is used at the inference stage to detect objects for a given image. With the uncertainty branch, our model also predicts the aleatoric uncertainty of a bounding box, which can be interpreted as the confidence of the boundary location.

\subsection{Object-wise Contrastive Learning}
\label{sec:contrastive}
\begin{figure}[t]
    \centering
    \includegraphics[trim=0cm 0cm 0cm 0cm, clip=true, width=\columnwidth]{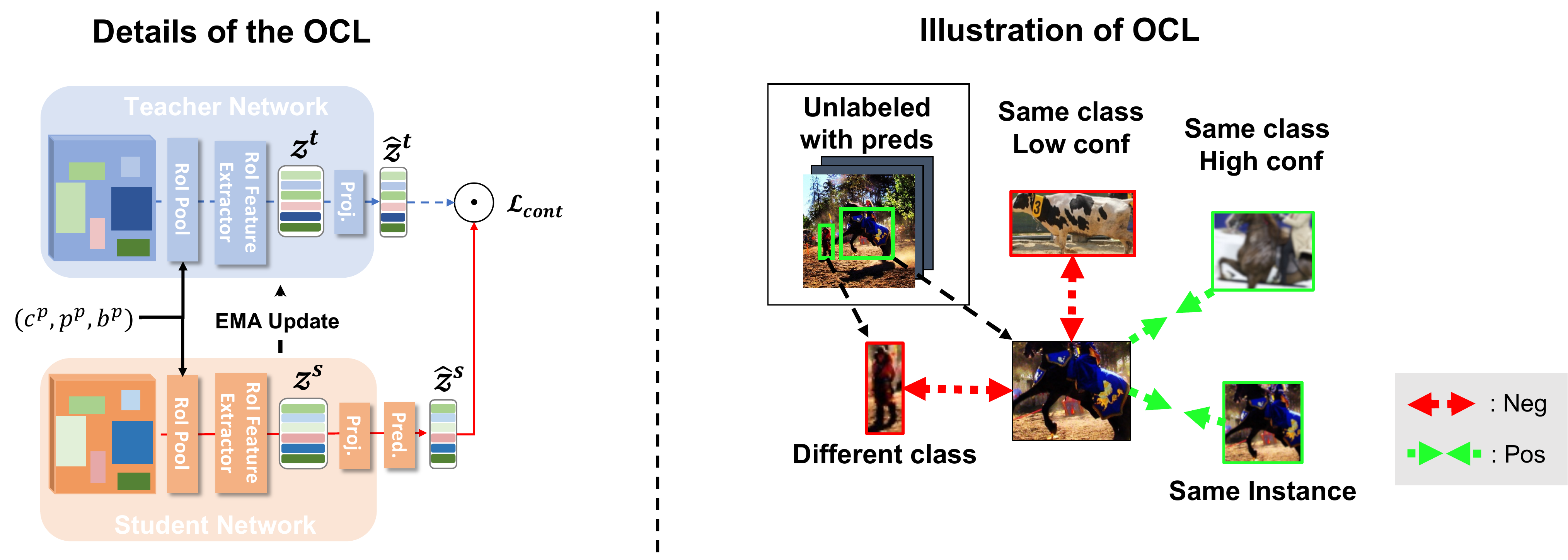}
    \caption{Left: Training details of OCL. Right: Intuitive illustration of the OCL.
    }
    \label{fig:ocl_toy_example}
\end{figure}

In the classification head, we propose the object-wise contrastive learning (OCL) to regularize the feature representation of ROI objects. 
We visualize the details of OCL and show an illustration in Fig.~\ref{fig:ocl_toy_example}. 

To use the contrastive loss, we construct positive and negative pairs of ROI objects within each batch during training.
We first predict the class category ${c}^p$, classification score ${p}^p$, and bounding box ${b}^p$ for each ROI object detected by the teacher network taking as input the weakly augmented unlabeled images $\mathcal{A}_0(\bm{x}^u)$. We denote the number of detected ROI objects in a batch as $N_{o}$.
We then input two strong augmented images $\mathcal{A}_1(\bm{x}^u)$ and $\mathcal{A}_2(\bm{x}^u)$ to the teacher and student networks to get the ROI features $\bm{z}^t$ and $\bm{z}^s$ using predicted bounding boxes ${b}^p$, respectively.
Then, $\bm{z}^t$ and $\bm{z}^s$ are projected to the low-dimensional feature space
and normalized with L2-normalization to generate $\bm{\hat{z}}^t$ and $\bm{\hat{z}}^s$, respectively.

As the predicted class is noisy, we take the classification score into consideration to define ROI object pairs, as 
shown in Eq.~\ref{eq:confidence-aware pairing}. 
A pair is positive if the $n$-th object feature $\bm{\hat{z}}^{s}_n$ and the $m$-th object feature $\bm{\hat{z}}^{t}_m$ belong to the same ROI.
We further measure the similarity if the two features are not from the same ROI but have the same class with high confidence.
\begin{equation}
    {w}_{nm} = \left\{
        \begin{array}{rcl}
            &1 & \mathrm{if}\; n=m,    \\ 
            &{p}^{p}_{n} \cdot {p}^{p}_{m} & \mathrm{if}\; n \neq m \;\mathrm{and} \;{c}^{p}_{n} = {c}^{p}_{m}\;\mathrm{and}\;{p}^{p}_{n} > T_{cont} \;\mathrm{and}\; {p}^{p}_{m} > T_{cont}, \\
            &0 & \mathrm{otherwise}.
        \end{array}
    \right.
    \label{eq:confidence-aware pairing}
\end{equation}
$T_{cont}$ is a classification score threshold for contrastive learning. Our contrastive loss is based on the supervised contrastive loss~\cite{khosla2020supcont}, and is formulated as 
\begin{equation}
    \mathcal{L}_{cont}
    = - \cfrac{1}{\mathcal{N}_o}\; \sum_{n=1}^{\mathcal{N}_o}\; \cfrac{1}{\sum_{m=1}^{\mathcal{N}_o} \mathbbm{1}_{({w}_{nm} > 0)}}\; \sum_{m=1}^{\mathcal{N}_o}\; {w}_{nm} \cdot \log{\cfrac{\exp({\bm{\hat{z}}}^{s}_{n} \cdot {\bm{\hat{z}}}^{t}_{m}/\tau)}{\sum_{l=1}^{\mathcal{N}_o} \mathbbm{1}_{(l \neq n)}\exp({\bm{\hat{z}}}^{s}_{n} \cdot {\bm{\hat{z}}}^{t}_{l}/\tau)}},
\label{eq:semi_cont_loss}
\end{equation}
where $\tau$ is the temperature parameter. $\mathbbm{1}(\cdot)$ equals to 1 if the condition is true, otherwise zero.
Note that, for the $n$-th ROI object, we normalize the loss by the number of positive pairs rather than the similarity. This aims to reduce the influence of mis-classified predictions.
We adopt the symmetrized loss~\cite{mocov3} to compute our object-wise contrastive loss as 
\begin{equation}
    \mathcal{L}_{ocl} = \cfrac{1}{2}(\mathcal{L}_{cont}  + \mathcal{L}'_{cont}),
\label{eq:total_semi_cont_loss}
\end{equation}
where $\mathcal{L}'_{cont}$ is the same loss as $\mathcal{L}_{cont}$ but with the augmentations for the teacher and student networks swapped.
Through the OCL (Eq.~\ref{eq:total_semi_cont_loss}), intra-class objects are encouraged to have similar feature representations, and inter-class objects are encouraged to have different feature representations. 
As a result, we can select pseudo-labels for classification task as
\begin{equation}
    \{\bm{X}^p_{cls}, \bm{Y}^p_{cls}\} = \{(\bm{x}_i^p, \bm{y}_i^p) | {p}^p_{i,\cdot} > T_{cls}\}_{i=1}^{N_u},
\end{equation}
where
$\bm{y}^p_i$  is the  pseudo-labels for the $i$-th image and consists of the class category
and bounding box
for objects,
${p}^p_{i,\cdot}$ is the classification score for the corresponding object in the $i$-th image, and $T_{cls}$ is a predefined threshold. 

\subsection{Regression-Uncertainty-guided Pseudo-Labeling}
\label{sec:uncertainty}

We propose regression-uncertainty-guided pseudo-labeling (RUPL) to utilize regression uncertainty to filter out unreliable bounding boxes. Following~\cite{box_regression_uncertainty}, we add an uncertainty head in parallel to classification and regression heads to predict regression uncertainty $\sigma$.
We use the uncertainty-aware regression loss~\cite{whatUncertainty} as the the student network's bounding box regression loss~\footnote{This loss is only applied to labeled images.},
\begin{equation}
    \mathcal{L}_{roi,reg} = 
    \cfrac{smoothL_1(\hat{t}, t)}{\sigma^2} + \lambda_{unc}  \log(\sigma^2),
    \label{eq:uncer}
\end{equation}
where $\hat{t}$ and $t$ denote the predicted bounding box offset and ground truth bounding box offset.
$\lambda_{unc}$ is a hyperparameter to control the effect of the uncertainty term.
During training, we apply this loss to each of the four boundaries of a bounding box. During inference, we use the average of these four uncertainties (denoted as $\bar{\sigma}$) as the regression uncertainty of the bounding box. 
We can then derive the selection of pseudo labels for regression as
\begin{equation}
    \{\bm{X}^p_{reg}, \bm{Y}^p_{reg}\} = \{(\bm{x}_i^p, \bm{y}_i^p) | {p}^p_{i,\cdot} > T_{cls}\; \mathrm{and}\;
    \bar{\sigma}^p_{i,\cdot} < T_{reg}\}_{i=1}^{N_u},
\end{equation}
where
$\bm{y}^p_i$  is the  pseudo-labels for the $i$-th image and consists of the class category and bounding box for objects,
${p}^p_{i,\cdot}$ and $\bar{\sigma}^p_{i,\cdot}$ are the classification score and regression uncertainty for the corresponding object in the $i$-th image, and $T_{cls}$ $T_{reg}$ are predefined thresholds. 
By learning the regression uncertainty, we provide an alternative way to capture the reliability of pseudo-labels without additional forwarding or reformulation. 

\section{Experiments}
\label{sec:exper}
\subsection{Experimental Settings}
Following previous works~\cite{liu2021unbiased}, we evaluate our method on Pascal VOC~\cite{voc} and MS-COCO~\cite{mscoco}. We conduct experiments using different settings. (1) VOC~\cite{liu2021unbiased}: VOC07-trainval and VOC12-trainval are used as labeled and unlabeled datasets, respectively.
We also show results using COCO20cls~\cite{liu2021unbiased} as an additional unlabeled dataset. VOC07-test is used to evaluate. (2) COCO-standard~\cite{liu2021unbiased}: We randomly select 1\%/5\%/10\% samples from COCO2017 train dataset as our labeled datasets, and the rest of them as unlabeled datasets. COCO2017 validation set is used to evaluate. (3) COCO-35k~\cite{wang2021data}: We use a subset of COCO2014 validation set as a labeled dataset and COCO2014 training set as an unlabeled dataset. COCO2014 minival is used to evaluate. (4) COCO-additional~\cite{liu2021unbiased}: We use COCO2017 train dataset and COCO2017 unlabeled dataset as labeled and unlabeled datasets, respectively. COCO2017 validation set is used to evaluate.

\subsection{Implementation details}
Following~\cite{STAC, liu2021unbiased}, our object detector is Faster-RCNN~\cite{ren2015faster} with ResNet-50 backbone and FPN~\cite{fpn}. For VOC and VOC with COCO20cls settings, we train our model by 60k and 90k iterations, respectively, including 12k iterations for pre-training. For the coco-standard setting, we pre-train 5k/20k/40k iterations for 1\%/5\%/10\% of COCO-standard and persist the training until 180k iterations. For the COCO-35k and COCO-additional settings, models are pre-trained/trained for 12k/180k and 90k/360k iterations, respectively. More training details, including augmentation strategies and model architecture, are in the supplementary.

\subsection{Results}
\noindent \textbf{VOC. }
Experimental results on Pascal VOC~\cite{voc} are shown in Tab.~\ref{tab:voc_exp}. We add our proposed OCL and RUPL to UBT~\cite{liu2021unbiased} without additional augmentation strategies. When using VOC12-trainval as unlabeled data, our model outperforms UBT~\cite{liu2021unbiased} by 3.35 mAP and 0.89 mAP on $\text{AP}_{50:95}$ and $\text{AP}_{50}$, respectively. When using COCO20cls~\cite{liu2021unbiased} as additional unlabeled data, our model has larger improvements and outperforms UBT~\cite{liu2021unbiased} by 3.54 mAP and 1.25 mAP on $\text{AP}_{50:95}$ and $\text{AP}_{50}$, respectively. This shows that our proposed method can better benefit from unlabeled data.
Moreover, our method outperforms other existing models~\cite{STAC,combatnoise,instant_teaching,humbleteacher,adaptiveclass,rethink,zheng2022dual,MUM,globalclassprototypes,ubteacherv2,scale-equivalent} on both experiment settings, which shows the superiority of our method. 
 \begin{table}[t]
    \centering
    \scriptsize
    \begin{tabular}{c|c|c|cc||c|cc}
        \toprule
         Method & Conference & Unlabeled & $\text{AP}_{50:95}$  & $\text{AP}_{50}$  & Unlabeled & $\text{AP}_{50:95}$  & $\text{AP}_{50}$\\
        \midrule
        $\mathrm{Supervised}^\dagger$\cite{liu2021unbiased} & - & None & 42.13 & 72.63 & None & 42.13 & 72.63 \\
        \midrule
        STAC\cite{STAC} & - & {} & 44.64 & 77.45 & {} & 46.01 & 79.08 \\
        $\mathrm{UBT}^\dagger$\cite{liu2021unbiased} & ICLR2021 & {} & 48.69 & 77.37 & {} & 50.34 & 78.82 \\
        IT~\cite{instant_teaching} & CVPR2021 & {} & 50.00 & 79.20 & {} & 50.80 & 79.90 \\
        HT\cite{humbleteacher} & CVPR2021 &  {} & 53.04 & 80.94  & {} & 54.41 & 81.29 \\
        CN~\cite{combatnoise} & NIPS2021 & {} & 49.3 & 80.6 & {} & 50.2 & 81.4 \\
        ST~\cite{globalclassprototypes} & ICCV2021 & {} & - & 80.32 & {} & - & - \\
        ACRST\cite{adaptiveclass}  & AAAI2022 & {} & 54.30 & 81.11 & VOC12-trainval & - & - \\
        RPL\cite{rethink} & AAAI2022 & VOC12-trainval & 54.6 & 79.0 & + & 56.1 & 79.6 \\
        DDT\cite{zheng2022dual} & AAAI2022 & {} & 54.7 & \textbf{82.4} & COCO20cls & 55.9 & 82.5 \\
        $\mathrm{MUM}^\dagger$\cite{MUM} & CVPR2022 & {} & 50.22 & 78.94 & {} & 52.31 & 80.45 \\
        MA-GCP~\cite{globalclassprototypes} & CVPR2022 & {} & - & 81.72 & {} & - & - \\
        SED~\cite{scale-equivalent} & CVPR2022 & {} & - & 80.60  & {} & - & - \\
        $\mathrm{UBTv2}^\ddagger$~\cite{ubteacherv2} & CVPR2022 & {} & 56.87 & 81.29 & {} & 58.08\ & 82.04 \\
        $\mathrm{Ours}^\dagger$ & - & {} & 52.04 & 78.29 & {} & 53.88 & 80.07 \\
        $\mathrm{Ours}^\ddagger$ & - & {} & \textbf{57.34} & 82.18 & {} & \textbf{58.99} & \textbf{82.98} \\
        \bottomrule
    \end{tabular}
    \caption{Experimental results on Pascal VOC dataset~\cite{voc}. Our method can outperform existing works. $\dagger$ and $\ddagger$ denote results that are evaluated by the COCOevaluator and VOCevaluator in Detectron2~\cite{wu2019detectron2}, respectively. Unmarked works are not implemented on Detectron2~\cite{wu2019detectron2}, or are hard to check through the paper or released code. We give more information about the evaluator in our supplementary Sec.~\textcolor{red}{3}.
    }
    \label{tab:voc_exp}
\end{table}

\begin{table}[h]
    \centering
    \scriptsize
    \begin{tabular}{c|c}
        \toprule
        \multicolumn{2}{c}{COCO-35k} \\
        \midrule
        Method & $\text{AP}_{50:95}$ \\
        \midrule
        Supervised~\cite{wang2021data} & 31.3 \\
        DD~\cite{wang2021data} & 33.1 \\
        MP~\cite{wang2021data} & 34.8 \\
        MP + DD~\cite{wang2021data} & 35.2 \\
        $\mathrm{UBT}^\ddagger$ & 36.36 \\
        \midrule
        Ours & \textbf{37.13} \\
        \bottomrule
    \end{tabular}
    \qquad
    \begin{tabular}{c|c|c|c}
        \toprule
        \multicolumn{4}{c}{COCO-standard} \\
        \midrule
        Method &  1\% & 5\% & 10\% \\
        \midrule
        Supervised~\cite{liu2021unbiased} & 9.05 & 18.47 & 23.86 \\
        STAC~\cite{STAC} & 13.97 & 24.38 & 28.64 \\
        UBT~\cite{liu2021unbiased} & 20.75 & 28.27 & 31.50 \\
        IT~\cite{instant_teaching} & 18.05 & 26.75 & 30.4 \\
        RPL~\cite{rethink} & 18.21 & 27.78 & 31.67 \\ 
        CN~\cite{combatnoise} & 18.41 & 28.96 & 32.43 \\
        $\mathrm{ST}^\dagger$~\cite{xu2021soft} & 20.46 & 30.74 & 34.04 \\
        DDT~\cite{zheng2022dual} & 18.62 & 29.24 & 32.80 \\
        MUM~\cite{MUM} & 21.88 & 28.52 & 31.87 \\
        $\mathrm{UBTv2}^\dagger$~\cite{ubteacherv2} & \textbf{25.40} & \textbf{31.85} & \textbf{35.08} \\
        \midrule
        Ours & 21.63 & 30.66 & 33.53 \\
        \bottomrule
    \end{tabular}
    \qquad
    \begin{tabular}{c|c}
        \toprule
        \multicolumn{2}{c}{COCO-additional} \\
        \midrule
        Method &  $\text{AP}_{50:95}$ \\
        \midrule
        Supervised~\cite{liu2021unbiased} & 40.20 \\
        STAC~\cite{STAC}& 39.21 \\
        PL~\cite{proposal_learning} & 38.40 \\
        UBT~\cite{liu2021unbiased}  & 41.30 \\
        IT~\cite{instant_teaching} & 40.20 \\
        RPL~\cite{rethink} & 43.30 \\ 
        CN~\cite{combatnoise} & 43.20 \\
        $\mathrm{ST}^\dagger$~\cite{xu2021soft} & 44.50 \\
        DDT~\cite{zheng2022dual} & 41.90 \\
        MUM~\cite{MUM} & 42.11 \\
        $\mathrm{UBTv2}^\dagger$~\cite{ubteacherv2} & \textbf{44.75} \\
        \midrule
        Ours & 41.89 \\
        \bottomrule
    \end{tabular}
    \caption{Experimental results on MS-COCO. We report the $\text{AP}_{50:95}$ of different methods. $\dagger$ denotes using extra augmentation such as a scale-jittering augmentation strategy to improve the accuracy, while we do not use. It significantly improves the mAP by about 1.37AP~(We refer the reader to Tab. 6 of the supplementary of UBTv2~\cite{ubteacherv2}).
    $\ddagger$ denotes our implemented result based on the official repository.
    }
    \label{tab:coco}
\end{table}
\makeatletter
\newcommand\tabcaption{\def\@captype{table}\caption}
\newcommand\figcaption{\def\@captype{figure}\caption}
\makeatother

\noindent \textbf{MS-COCO.}
We also conduct experiments on MS-COCO~\cite{mscoco} to verify the effectiveness of our method, shown in Tab.~\ref{tab:coco}. Comparing with UBT~\cite{liu2021unbiased}, our model outperform it by 0.88/2.39/2.03 mAP, 0.77 mAP, and 0.59 mAP on COCO-standard
,COCO-35k, and COCO-additional. respectively. 
Moreover, our method achieves comparable results on COCO-standard, COCO-35k, and COCO-additional compared to other state-of-the-arts.
These results consistently support the effectiveness of our method.

\subsection{Ablation studies}
We conduct ablation studies to investigate the effectiveness of our model using 1\% COCO-standard.
Because of the limitation of computing resources, all experiments in this section are conducted with batch size 12/12 (labeled/unlabeled) and training iteration 45k, as in~\cite{MUM}.

\begin{figure}[t]
\begin{minipage}[b]{0.47\textwidth}
\centering
\includegraphics[trim=0.5cm 0.2cm 0.5cm 0.5cm, clip=true, width=0.7\columnwidth]{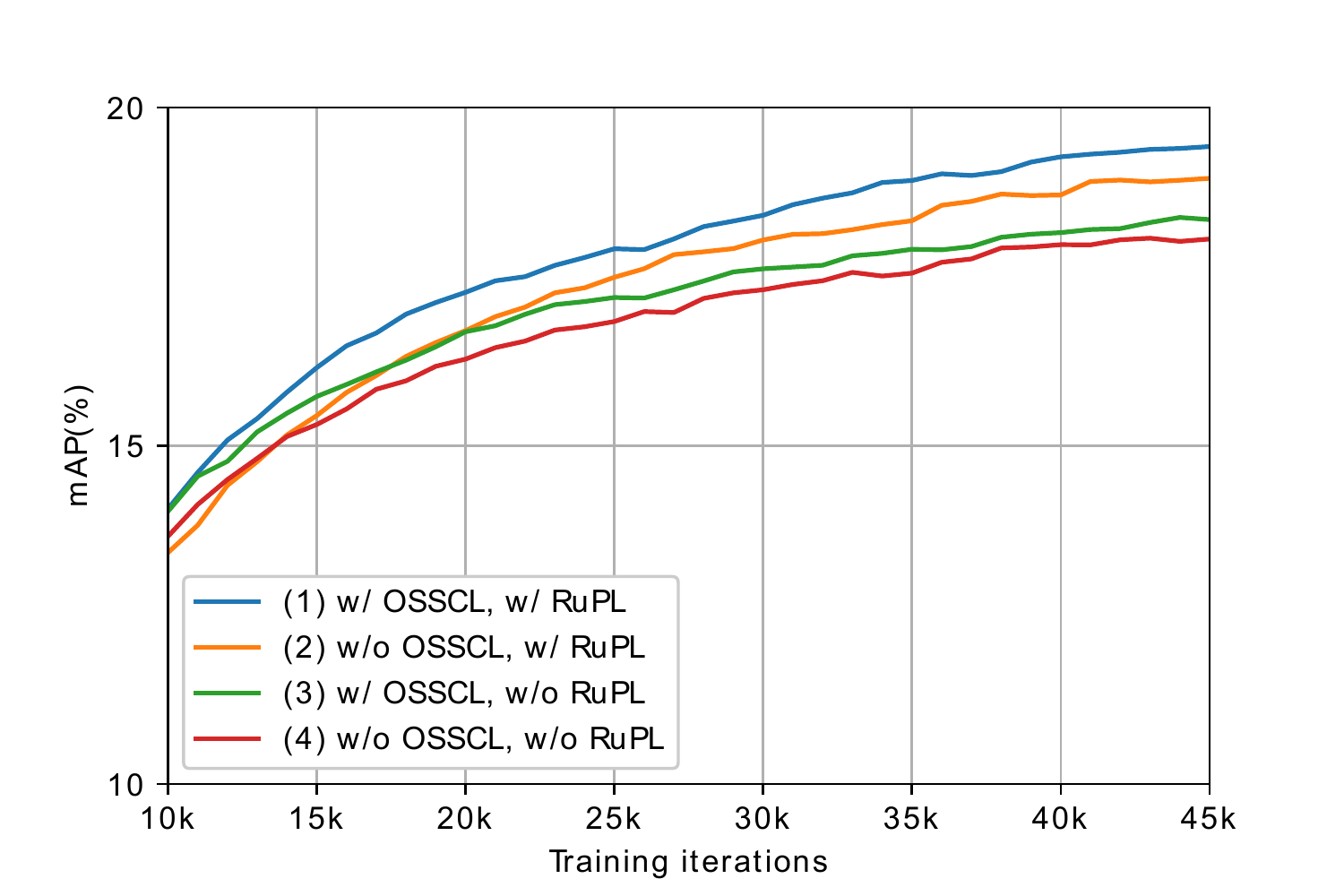}
\caption{mAP ($\mathrm{AP}_{50:95}$) curves during training of the experiments in Tab~\ref{tab:ablation_exp}.}
\label{fig:ablation}
\end{minipage}%
\hfill%
\begin{minipage}[b]{0.47\textwidth}
\scriptsize
    \begin{tabular}{ccc|ccc}
        \toprule
        {} & OCL & RUPL & $\text{AP}_{50:95}$  & $\text{AP}_{50}$ & $\text{AP}_{75}$ \\ 
        \midrule
        \multicolumn{1}{c|}{(1)} & \checkmark & \checkmark  & \textbf{19.42} & 34.65 & \textbf{19.36} \\
        \multicolumn{1}{c|}{(2)} & {} & \checkmark  & 18.95 & 33.82 & 19.07 \\
        \multicolumn{1}{c|}{(3)} & \checkmark &  {} & 18.34 & \textbf{35.21} & 17.16 \\
        \multicolumn{1}{c|}{(4)} & {} & {} & 18.05 & 34.45 & 17.09 \\
        \bottomrule
    \end{tabular}
\tabcaption{Ablation studies of different modules.}
\label{tab:ablation_exp}
\end{minipage}%
\end{figure}

\begin{table}[t]
\begin{minipage}[b]{0.47\textwidth}\centering
    \scriptsize
    \begin{tabular}{c|ccc}
    \toprule
         Method &  $\text{AP}_{50:95}$ & $\text{AP}_{50}$ & $\text{AP}_{75}$ \\
         \midrule
         w/o CL &  18.05 & 34.45 & 17.09\\
         Self-sup CL &  18.24 & 34.89 & \textbf{17.22}\\
         OCL (Ours) &  \textbf{18.34} & \textbf{35.21} & 17.16\\
    \bottomrule
    \end{tabular}
    \caption{Ablation studies of OCL. CL denotes contrastive learning. A pair in Self-sup CL is positive if and only if they are the same instance(i = j in Eq.~\ref{eq:confidence-aware pairing}). We exclude RUPL to facilitate the comparison.}
    \label{tab:self_semi_cont}
\end{minipage}%
\hfill%
\begin{minipage}[b]{0.47\textwidth}\centering
    \centering
    \tiny
    \begin{tabular}{c|ccc}
        \toprule
        Method & $\text{AP}_{50:95}$ & $\text{AP}_{50}$ & $\text{AP}_{75}$ \\
        \midrule
        Box jittering~\cite{xu2021soft} & 18.15 & \textbf{34.84} & 17.37 \\
        Predicted IoU~\cite{wang20213dioumatch} & 18.85 & 34.76 & 18.26 \\
        Aleatoric uncertainty~\cite{whatUncertainty} (Ours) & \textbf{18.95} & 33.81 & \textbf{19.06} \\
        \bottomrule
    \end{tabular}
    \caption{Different methods capturing the localization quality of pseudo-labels. Aleatoric uncertainty~\cite{whatUncertainty} (Ours) is the best. We exclude OCL to facilitate comparison.
    }
    \label{tab:measurement}
\end{minipage}%
\end{table}
\noindent\textbf{Effectiveness of proposed modules.} 
We remove each proposed module from our framework and report results in Tab.~\ref{tab:ablation_exp}.
Comparing (3) to (4), we can see introducingOCL can improve both $\text{AP}_{50}$ and $\text{AP}_{75}$. This result demonstrates that the model benefits from more accurate pseudo-labels w.r.t classification. Comparing (2) to (4), we can see introducing RUPL can improve $\text{AP}_{75}$ but decrease $\text{AP}_{50}$. We assume the reason is that RUPL makes the model focus more on regression than classification during training. Comparing (1) to (2-4), we can see the accuracy of (1) is significantly higher than that of (2-4), which shows applying both OCL and RUPL can lead to the best accuracy. We argue that the model can generate more precise pseudo-labels with more discriminative classification scores and the introduced localization uncertainties, which leads to the improvement of accuracy. We also show the mAP curves of different experiments during training in Fig.~\ref{fig:ablation}, which further supports the effectiveness of our proposed modules.
We emphasize that the proposed OCL and RUPL can complement each other and synergistically improve the model performance.

\noindent\textbf{Ablation studies of OCL.}
To verify the effectiveness of the class information of our OCL,  we show the results of three models in Tab.~\ref{tab:self_semi_cont}, i.e., without, self-supervised, and, object-wise semi-supervised contrastive learning (our OCL).
In Tab.~\ref{tab:self_semi_cont}, self-sup CL improves the model performance compared to w/o CL and
OCL further improves the performance since it helps the model to learn more discriminative feature representation for objects from different classes. This supports the effectiveness of our OCL.

\noindent\textbf{Different localization quality measurements.} In Tab.~\ref{tab:measurement}, we compare different localization quality measurements. 
Specifically, we compare box jittering~\cite{xu2021soft}, predicted IoU~\cite{wang20213dioumatch}, and aleatoric uncertainty~\cite{whatUncertainty} in our RUPL. We use grid-search to find the best thresholds of box jittering~\cite{xu2021soft} and predicted IoU~\cite{wang20213dioumatch} to filter pseudo-labels, and set them as 0.01 and 0.8, respectively.
The model with aleatoric uncertainty~\cite{whatUncertainty} achieves the highest performance on the $\mathrm{AP}_{50:95}$.

\noindent\textbf{Different Regression Thresholds of RUPL.}
In this section, we show experimental results on how different thresholds of RUPL ($T_{reg}$) affect the detection accuracy. As shown in Tab. \ref{tab:thres_exp}, the model achieves the highest performance when we set the threshold as 0.5. The detection accuracy decreases when we set a larger threshold or small threshold. With a larger threshold, the selected samples become more diverse but unreliable. While with a smaller threshold, the selected samples become more reliable but monotonous.
\begin{table}[h]
    \centering
    \scriptsize
    \begin{tabular}{c|ccc}
        \toprule
        Threshold & $\text{AP}_{50:95}$ & $\text{AP}_{50}$ & $\text{AP}_{75}$ \\ 
        \midrule
        0.3       & 19.18          & 34.53          & 19.13          \\
        0.4       & 19.37       & 34.49               & 19.28               \\
        0.5       & \textbf{19.42} & \textbf{34.65} & \textbf{19.36} \\
        0.6       & 19.26          & 34.45          & \textbf{19.36}          \\
        0.7       & 18.78          & 33.50          & 18.93          \\
        \bottomrule
    \end{tabular}
    \caption{Detection accuracy with different regression thresholds of RUPL.}
    \label{tab:thres_exp}
\end{table}

\section{Conclusion}
\label{sec:con}
In this paper, we propose a two-step pseudo-label filtering for SSOD. 
We deal with both the classification and regression heads in the detection model. For the classification head, we propose an object-wise contrastive learning loss to exploit the unlabeled data to enhance the discriminativeness of classification score for pseudo label filtering. 
For the regression head, we design an uncertainty branch to learn regression uncertainty to measure the localization quality for bounding box filtering. 
We experimentally show that the two components create a synergistic effect when integrated into the teacher-student framework. 
Our framework achieves remarkable performance gain against our baseline on both PASCAL VOC and MS-COCO without additional augmentation, and shows competitive results compared to other state-of-the-arts.

\section*{Acknowledgement}
This work is in part sponsored by KAIA grant (22CTAP-C163793-02, MOLIT), NST grant (CRC 21011, MSIT), KOCCA grant (R2022020028, MCST) and the Samsung Display corporation.

\bibliography{ms}

\begin{thebibliography}{42}
\providecommand{\natexlab}[1]{#1}
\providecommand{\url}[1]{\texttt{#1}}
\expandafter\ifx\csname urlstyle\endcsname\relax
  \providecommand{\doi}[1]{doi: #1}\else
  \providecommand{\doi}{doi: \begingroup \urlstyle{rm}\Url}\fi

\bibitem[Chen et~al.(2022{\natexlab{a}})Chen, Chen, Yang, Xuan, Song, Xie, Pu,
  Song, and Zhuang]{label_matching}
Binbin Chen, Weijie Chen, Shicai Yang, Yunyi Xuan, Jie Song, Di~Xie, Shiliang
  Pu, Mingli Song, and Yueting Zhuang.
\newblock Label matching semi-supervised object detection.
\newblock In \emph{CVPR}, pages 14381--14390, June 2022{\natexlab{a}}.

\bibitem[Chen et~al.(2022{\natexlab{b}})Chen, Li, Chen, Wang, Zhang, and
  Hua]{denselearning}
Binghui Chen, Pengyu Li, Xiang Chen, Biao Wang, Lei Zhang, and Xian-Sheng Hua.
\newblock Dense learning based semi-supervised object detection.
\newblock In \emph{CVPR}, pages 4815--4824, June 2022{\natexlab{b}}.

\bibitem[Chen et~al.(2020)Chen, Kornblith, Norouzi, and Hinton]{simclr}
Ting Chen, Simon Kornblith, Mohammad Norouzi, and Geoffrey~E. Hinton.
\newblock A simple framework for contrastive learning of visual
  representations.
\newblock In \emph{ICML}, volume 119, pages 1597--1607, 2020.

\bibitem[Chen et~al.(2021)Chen, Xie, and He]{mocov3}
Xinlei Chen, Saining Xie, and Kaiming He.
\newblock An empirical study of training self-supervised vision transformers.
\newblock In \emph{ICCV}, pages 9620--9629, 2021.

\bibitem[Choi et~al.(2019)Choi, Chun, Kim, and Lee]{gaussianyolo}
Jiwoong Choi, Dayoung Chun, Hyun Kim, and Hyuk{-}Jae Lee.
\newblock Gaussian yolov3: An accurate and fast object detector using
  localization uncertainty for autonomous driving.
\newblock In \emph{ICCV}, pages 502--511, 2019.

\bibitem[Everingham et~al.(2010)Everingham, Gool, Williams, Winn, and
  Zisserman]{voc}
Mark Everingham, Luc~Van Gool, Christopher K.~I. Williams, John~M. Winn, and
  Andrew Zisserman.
\newblock The pascal visual object classes {(VOC)} challenge.
\newblock \emph{Int. J. Comput. Vis.}, 88\penalty0 (2):\penalty0 303--338,
  2010.

\bibitem[Girshick(2015)]{girshick2015fast}
Ross~B. Girshick.
\newblock Fast {R-CNN}.
\newblock In \emph{ICCV}, pages 1440--1448, 2015.

\bibitem[Grill et~al.(2020)Grill, Strub, Altch{\'{e}}, Tallec, Richemond,
  Buchatskaya, Doersch, Pires, Guo, Azar, Piot, Kavukcuoglu, Munos, and
  Valko]{byol}
Jean{-}Bastien Grill, Florian Strub, Florent Altch{\'{e}}, Corentin Tallec,
  Pierre~H. Richemond, Elena Buchatskaya, Carl Doersch, Bernardo~{\'{A}}vila
  Pires, Zhaohan Guo, Mohammad~Gheshlaghi Azar, Bilal Piot, Koray Kavukcuoglu,
  R{\'{e}}mi Munos, and Michal Valko.
\newblock Bootstrap your own latent - {A} new approach to self-supervised
  learning.
\newblock In \emph{NeurIPS}, 2020.

\bibitem[Guo et~al.(2022)Guo, Mu, Chen, Wang, Yu, and Luo]{scale-equivalent}
Qiushan Guo, Yao Mu, Jianyu Chen, Tianqi Wang, Yizhou Yu, and Ping Luo.
\newblock Scale-equivalent distillation for semi-supervised object detection.
\newblock In \emph{CVPR}, pages 14522--14531, June 2022.

\bibitem[He et~al.(2019)He, Zhu, Wang, Savvides, and
  Zhang]{box_regression_uncertainty}
Yihui He, Chenchen Zhu, Jianren Wang, Marios Savvides, and Xiangyu Zhang.
\newblock Bounding box regression with uncertainty for accurate object
  detection.
\newblock In \emph{CVPR}, pages 2888--2897, 2019.

\bibitem[Jeong et~al.(2019)Jeong, Lee, Kim, and Kwak]{NEURIPS2019_csd}
Jisoo Jeong, Seungeui Lee, Jeesoo Kim, and Nojun Kwak.
\newblock Consistency-based semi-supervised learning for object detection.
\newblock In \emph{NeurIPS}, pages 10758--10767, 2019.

\bibitem[Jeong et~al.(2021)Jeong, Verma, Hyun, Kannala, and Kwak]{ISD}
Jisoo Jeong, Vikas Verma, Minsung Hyun, Juho Kannala, and Nojun Kwak.
\newblock Interpolation-based semi-supervised learning for object detection.
\newblock In \emph{CVPR}, pages 11602--11611, 2021.

\bibitem[Jiang et~al.(2018)Jiang, Luo, Mao, Xiao, and Jiang]{iounet}
Borui Jiang, Ruixuan Luo, Jiayuan Mao, Tete Xiao, and Yuning Jiang.
\newblock Acquisition of localization confidence for accurate object detection.
\newblock In \emph{ECCV}, volume 11218, pages 816--832, 2018.

\bibitem[Jiang et~al.(2021)Jiang, Shi, Tian, Lai, Liu, Fu, and
  Jia]{pointcontrasitve}
Li~Jiang, Shaoshuai Shi, Zhuotao Tian, Xin Lai, Shu Liu, Chi{-}Wing Fu, and
  Jiaya Jia.
\newblock Guided point contrastive learning for semi-supervised point cloud
  semantic segmentation.
\newblock In \emph{ICCV}, pages 6403--6412, 2021.

\bibitem[Kendall and Gal(2017)]{whatUncertainty}
Alex Kendall and Yarin Gal.
\newblock What uncertainties do we need in bayesian deep learning for computer
  vision?
\newblock In \emph{NeurIPS}, pages 5574--5584, 2017.

\bibitem[Khosla et~al.(2020)Khosla, Teterwak, Wang, Sarna, Tian, Isola,
  Maschinot, Liu, and Krishnan]{khosla2020supcont}
Prannay Khosla, Piotr Teterwak, Chen Wang, Aaron Sarna, Yonglong Tian, Phillip
  Isola, Aaron Maschinot, Ce~Liu, and Dilip Krishnan.
\newblock Supervised contrastive learning.
\newblock In \emph{NeurIPS}, 2020.

\bibitem[Kim et~al.(2022)Kim, Jang, Seo, Jeong, Na, and Kwak]{MUM}
JongMok Kim, JooYoung Jang, Seunghyeon Seo, Jisoo Jeong, Jongkeun Na, and Nojun
  Kwak.
\newblock Mum: Mix image tiles and unmix feature tiles for semi-supervised
  object detection.
\newblock In \emph{CVPR}, pages 14512--14521, June 2022.

\bibitem[Li et~al.(2022{\natexlab{a}})Li, Yuan, and Li]{globalclassprototypes}
Aoxue Li, Peng Yuan, and Zhenguo Li.
\newblock Semi-supervised object detection via multi-instance alignment with
  global class prototypes.
\newblock In \emph{CVPR}, pages 9809--9818, June 2022{\natexlab{a}}.

\bibitem[Li et~al.(2022{\natexlab{b}})Li, Wu, Shrivastava, and Davis]{rethink}
Hengduo Li, Zuxuan Wu, Abhinav Shrivastava, and Larry~S. Davis.
\newblock Rethinking pseudo labels for semi-supervised object detection.
\newblock In \emph{AAAI}, pages 1314--1322, 2022{\natexlab{b}}.

\bibitem[Lin et~al.(2014)Lin, Maire, Belongie, Hays, Perona, Ramanan,
  Doll{\'{a}}r, and Zitnick]{mscoco}
Tsung{-}Yi Lin, Michael Maire, Serge~J. Belongie, James Hays, Pietro Perona,
  Deva Ramanan, Piotr Doll{\'{a}}r, and C.~Lawrence Zitnick.
\newblock Microsoft {COCO:} common objects in context.
\newblock In \emph{ECCV}, pages 740--755, 2014.

\bibitem[Lin et~al.(2017)Lin, Doll{\'{a}}r, Girshick, He, Hariharan, and
  Belongie]{fpn}
Tsung{-}Yi Lin, Piotr Doll{\'{a}}r, Ross~B. Girshick, Kaiming He, Bharath
  Hariharan, and Serge~J. Belongie.
\newblock Feature pyramid networks for object detection.
\newblock In \emph{CVPR}, pages 936--944, 2017.

\bibitem[Lin et~al.(2020)Lin, Goyal, Girshick, He, and Doll{\'{a}}r]{focalloss}
Tsung{-}Yi Lin, Priya Goyal, Ross~B. Girshick, Kaiming He, and Piotr
  Doll{\'{a}}r.
\newblock Focal loss for dense object detection.
\newblock \emph{IEEE T-PAMI}, 42\penalty0 (2):\penalty0 318--327, 2020.

\bibitem[Liu et~al.(2021)Liu, Ma, He, Kuo, Chen, Zhang, Wu, Kira, and
  Vajda]{liu2021unbiased}
Yen{-}Cheng Liu, Chih{-}Yao Ma, Zijian He, Chia{-}Wen Kuo, Kan Chen, Peizhao
  Zhang, Bichen Wu, Zsolt Kira, and Peter Vajda.
\newblock Unbiased teacher for semi-supervised object detection.
\newblock In \emph{ICLR}, 2021.

\bibitem[Liu et~al.(2022)Liu, Ma, and Kira]{ubteacherv2}
Yen-Cheng Liu, Chih-Yao Ma, and Zsolt Kira.
\newblock Unbiased teacher v2: Semi-supervised object detection for anchor-free
  and anchor-based detectors.
\newblock In \emph{CVPR}, pages 9819--9828, June 2022.

\bibitem[Mi et~al.(2022)Mi, Lin, Zhou, Shen, Luo, Sun, Cao, Fu, Xu, and
  Ji]{activeteacher}
Peng Mi, Jianghang Lin, Yiyi Zhou, Yunhang Shen, Gen Luo, Xiaoshuai Sun,
  Liujuan Cao, Rongrong Fu, Qiang Xu, and Rongrong Ji.
\newblock Active teacher for semi-supervised object detection.
\newblock In \emph{CVPR}, pages 14482--14491, June 2022.

\bibitem[Ren et~al.(2015)Ren, He, Girshick, and Sun]{ren2015faster}
Shaoqing Ren, Kaiming He, Ross~B. Girshick, and Jian Sun.
\newblock Faster {R-CNN:} towards real-time object detection with region
  proposal networks.
\newblock In \emph{NeurIPS}, pages 91--99, 2015.

\bibitem[Rizve et~al.(2021)Rizve, Duarte, Rawat, and Shah]{rizve2021in}
Mamshad~Nayeem Rizve, Kevin Duarte, Yogesh~S. Rawat, and Mubarak Shah.
\newblock In defense of pseudo-labeling: An uncertainty-aware pseudo-label
  selection framework for semi-supervised learning.
\newblock In \emph{ICLR}, 2021.

\bibitem[Sohn et~al.(2020)Sohn, Zhang, Li, Zhang, Lee, and Pfister]{STAC}
Kihyuk Sohn, Zizhao Zhang, Chun{-}Liang Li, Han Zhang, Chen{-}Yu Lee, and Tomas
  Pfister.
\newblock A simple semi-supervised learning framework for object detection.
\newblock \emph{CoRR}, abs/2005.04757, 2020.

\bibitem[Tang et~al.(2021{\natexlab{a}})Tang, Ramaiah, Wang, Xu, and
  Xiong]{proposal_learning}
Peng Tang, Chetan Ramaiah, Yan Wang, Ran Xu, and Caiming Xiong.
\newblock Proposal learning for semi-supervised object detection.
\newblock In \emph{WACV}, pages 2290--2300, 2021{\natexlab{a}}.

\bibitem[Tang et~al.(2021{\natexlab{b}})Tang, Chen, Luo, and
  Zhang]{humbleteacher}
Yihe Tang, Weifeng Chen, Yijun Luo, and Yuting Zhang.
\newblock Humble teachers teach better students for semi-supervised object
  detection.
\newblock In \emph{CVPR}, pages 3132--3141, 2021{\natexlab{b}}.

\bibitem[Tarvainen and Valpola(2017)]{meanteacher}
Antti Tarvainen and Harri Valpola.
\newblock Mean teachers are better role models: Weight-averaged consistency
  targets improve semi-supervised deep learning results.
\newblock In \emph{NeurIPS}, pages 1195--1204, 2017.

\bibitem[van~den Oord et~al.(2018)van~den Oord, Li, and Vinyals]{infonce}
A{\"{a}}ron van~den Oord, Yazhe Li, and Oriol Vinyals.
\newblock Representation learning with contrastive predictive coding.
\newblock \emph{CoRR}, abs/1807.03748, 2018.

\bibitem[Wang et~al.(2021{\natexlab{a}})Wang, Cong, Litany, Gao, and
  Guibas]{wang20213dioumatch}
He~Wang, Yezhen Cong, Or~Litany, Yue Gao, and Leonidas~J. Guibas.
\newblock 3dioumatch: Leveraging iou prediction for semi-supervised 3d object
  detection.
\newblock In \emph{CVPR}, pages 14615--14624, 2021{\natexlab{a}}.

\bibitem[Wang et~al.(2021{\natexlab{b}})Wang, Li, Guo, and Wang]{combatnoise}
Zhenyu Wang, Ya{-}Li Li, Ye~Guo, and Shengjin Wang.
\newblock Combating noise: Semi-supervised learning by region uncertainty
  quantification.
\newblock In \emph{NeurIPS}, pages 9534--9545, 2021{\natexlab{b}}.

\bibitem[Wang et~al.(2021{\natexlab{c}})Wang, Li, Guo, Fang, and
  Wang]{wang2021data}
Zhenyu Wang, Yali Li, Ye~Guo, Lu~Fang, and Shengjin Wang.
\newblock Data-uncertainty guided multi-phase learning for semi-supervised
  object detection.
\newblock In \emph{CVPR}, pages 4568--4577, 2021{\natexlab{c}}.

\bibitem[Wu et~al.(2019)Wu, Kirillov, Massa, Lo, and
  Girshick]{wu2019detectron2}
Yuxin Wu, Alexander Kirillov, Francisco Massa, Wan-Yen Lo, and Ross Girshick.
\newblock Detectron2.
\newblock \url{https://github.com/facebookresearch/detectron2}, 2019.

\bibitem[Xu et~al.(2021)Xu, Zhang, Hu, Wang, Wang, Wei, Bai, and
  Liu]{xu2021soft}
Mengde Xu, Zheng Zhang, Han Hu, Jianfeng Wang, Lijuan Wang, Fangyun Wei, Xiang
  Bai, and Zicheng Liu.
\newblock End-to-end semi-supervised object detection with soft teacher.
\newblock In \emph{ICCV}, pages 3040--3049, 2021.

\bibitem[Yang et~al.(2022)Yang, Wu, Zhang, Jiang, Liu, Zheng, Zhang, Wang, and
  Zeng]{class_aware}
Fan Yang, Kai Wu, Shuyi Zhang, Guannan Jiang, Yong Liu, Feng Zheng, Wei Zhang,
  Chengjie Wang, and Long Zeng.
\newblock Class-aware contrastive semi-supervised learning.
\newblock \emph{CoRR}, abs/2203.02261, 2022.

\bibitem[Yang et~al.(2021)Yang, Wei, Wang, Hua, and Zhang]{yang2021interactive}
Qize Yang, Xihan Wei, Biao Wang, Xian{-}Sheng Hua, and Lei Zhang.
\newblock Interactive self-training with mean teachers for semi-supervised
  object detection.
\newblock In \emph{CVPR}, pages 5941--5950, 2021.

\bibitem[Zhang et~al.(2022)Zhang, Pan, and Wang]{adaptiveclass}
Fangyuan Zhang, Tianxiang Pan, and Bin Wang.
\newblock Semi-supervised object detection with adaptive class-rebalancing
  self-training.
\newblock In \emph{AAAI}, pages 3252--3261, 2022.

\bibitem[Zheng et~al.(2022)Zheng, Chen, Cai, Ye, and Tan]{zheng2022dual}
Shida Zheng, Chenshu Chen, Xiaowei Cai, Tingqun Ye, and Wenming Tan.
\newblock Dual decoupling training for semi-supervised object detection with
  noise-bypass head.
\newblock In \emph{AAAI}, pages 3526--3534, 2022.

\bibitem[Zhou et~al.(2021)Zhou, Yu, Wang, Qian, and Li]{instant_teaching}
Qiang Zhou, Chaohui Yu, Zhibin Wang, Qi~Qian, and Hao Li.
\newblock Instant-teaching: An end-to-end semi-supervised object detection
  framework.
\newblock In \emph{CVPR}, pages 4081--4090, 2021.

\end{thebibliography}
\newpage
\appendix

In this material, we provide: 
1) Why $\textrm{AP}_{50}$ decreases when RUPL is applied,
2) Justification for using three sets of augmented data,
3) Comparison between two different evaluators,
4) Applying RUPL to the RPN,
5) Applicable to anchor free detector,
6) Comparing with DDT,
7) Qualitative results,
8) Implementation details.

\section{Why $\textrm{AP}_{50}$ decreases when RUPL is applied}
In~\cite{box_regression_uncertainty}, $\textrm{AP}_{50}$ also decreases by up to~1.2 when applying the uncertainty-aware regression loss. We assume that this loss makes models focus more on more achievable training samples, so $\textrm{AP}_{75}$ increases and $\textrm{AP}_{50}$ decreases. Nevertheless, we emphasize that RUPL can improve $\textrm{AP}_{50:95}$, which is the most important metric.

\section{Justification for using three sets of augmented data} 
We generate additional strong augmented images for OCL following~\cite{simclr}, which uses two sets of strong augmented images. We also observe that using weakly augmented images for OCL decreases the overall AP in Tab.~\textcolor{red}{3} (1) by 0.24. 

\section{Comparison between two different evaluators}
\label{cocoVSvoc}
In official repository~\footnote{https://github.com/facebookresearch/unbiased-teacher}, UBT~\cite{liu2021unbiased} authors also notice that VOCevaluator results in higher accuracy than COCOevaluator, shown in Tab.~\ref{tab:eval}. They only report the results of UBTv2~\cite{ubteacherv2} with VOCevaluator without releasing code or models.
\begin{table}[h]
    \centering
    \scriptsize
    \begin{tabular}{c|c|c|c}
        \toprule
        Evaluator& Unlabeled & $\textrm{AP}_{50:95}$ & $\textrm{AP}_{50}$ \\
        \midrule
        COCO & VOC12 & 49.01 & 75.78 \\
        VOC & VOC12 & 54.48(+5.47) & 80.51(+4.73) \\
        \midrule
        COCO & VOC12+COCO20cls & 50.71 & 77.92 \\
        VOC &  VOC12+COCO20cls & 55.79(+5.08) & 81.71(+3.79) \\
        \bottomrule
    \end{tabular}
    \caption{Performance of official released models of UBT~\cite{liu2021unbiased}.}
    \label{tab:eval}
\end{table}

\section{Applying RUPL to the RPN}
Predicting uncertainty in the RPN is unnecessary because only the regression uncertainty in the ROI head can reflect the localization accuracy of pseudo-labels.~\cite{box_regression_uncertainty} also only predicts regression uncertainty in the ROI head. 

\section{Applicable to anchor-free detector} 
Our method can apply to anchor-free detectors. For OCL, we extract instance features from the feature map using detection results and feed them into the projection branch, which is added parallel to the classification branch. Also, RUPL can apply to an anchor-free detector because regression targets are four boundaries of a bounding box, similar to the anchor-free detector's regression targets.

\section{Comparing with DDT}
DDT~\cite{zheng2022dual} computes the localization quality of pseudo-labels through the IoU between outputs of two parallel heads. This method utilizes the output consistency, similar to box jittering in Tab.~\textcolor{red}{5}. We think box jittering is a more precise method because it uses ten samples to compute consistency. Our regression uncertainty is better than box jittering as shown in Tab.~\textcolor{red}{5}. Besides, DDT authors did not release the code.
\section{Qualitative Results}
\begin{figure}[h]
\begin{tabular}{c@{}c@{}c@{}c}
Baseline & Ours & Baseline & Ours \\
\includegraphics[width=0.24\columnwidth]{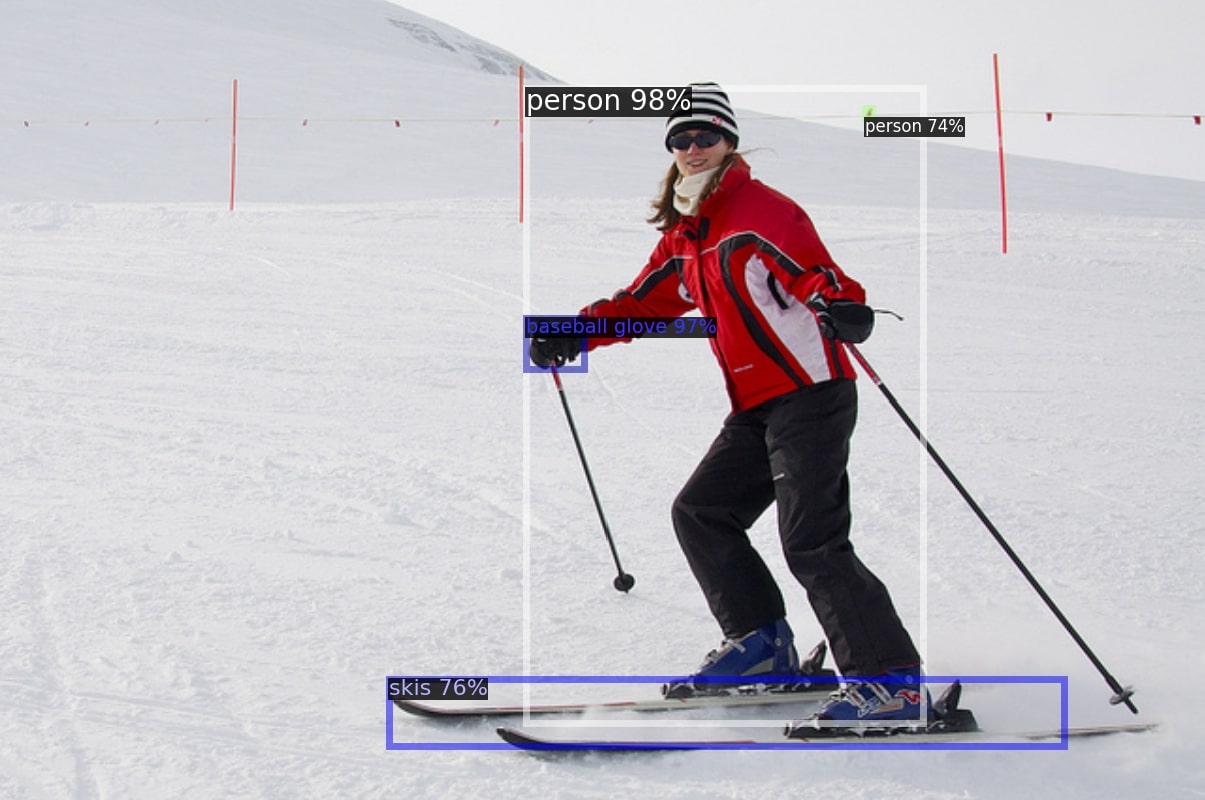}&
\includegraphics[width=0.24\columnwidth]{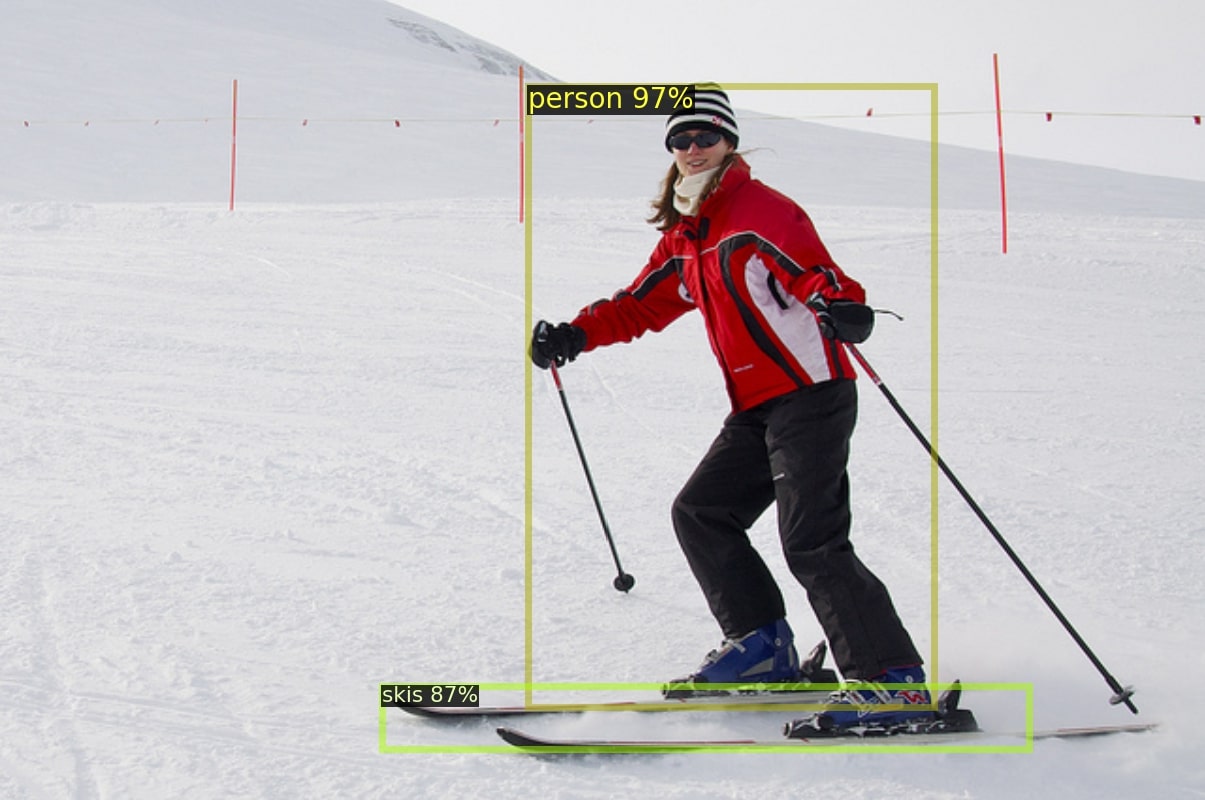} &
\includegraphics[width=0.24\columnwidth]{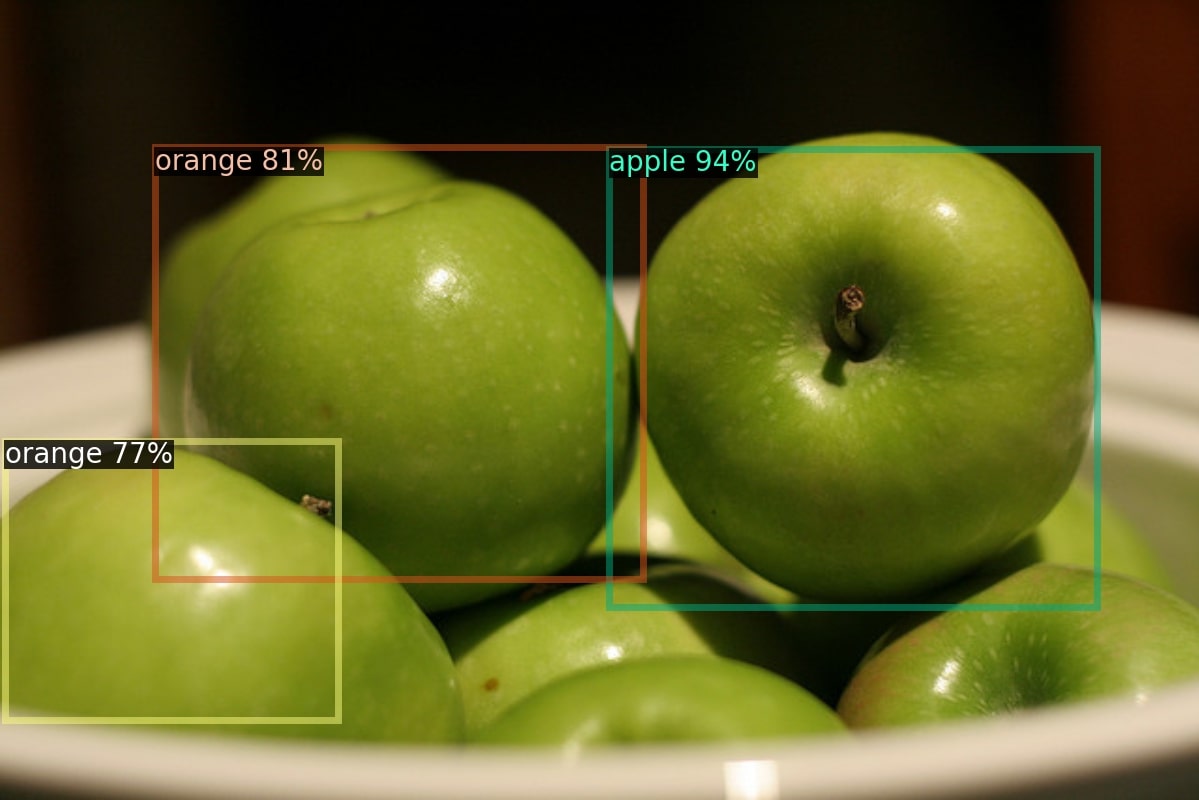}&
\includegraphics[width=0.24\columnwidth]{./ours/23.jpg} \\

\includegraphics[width=0.24\columnwidth]{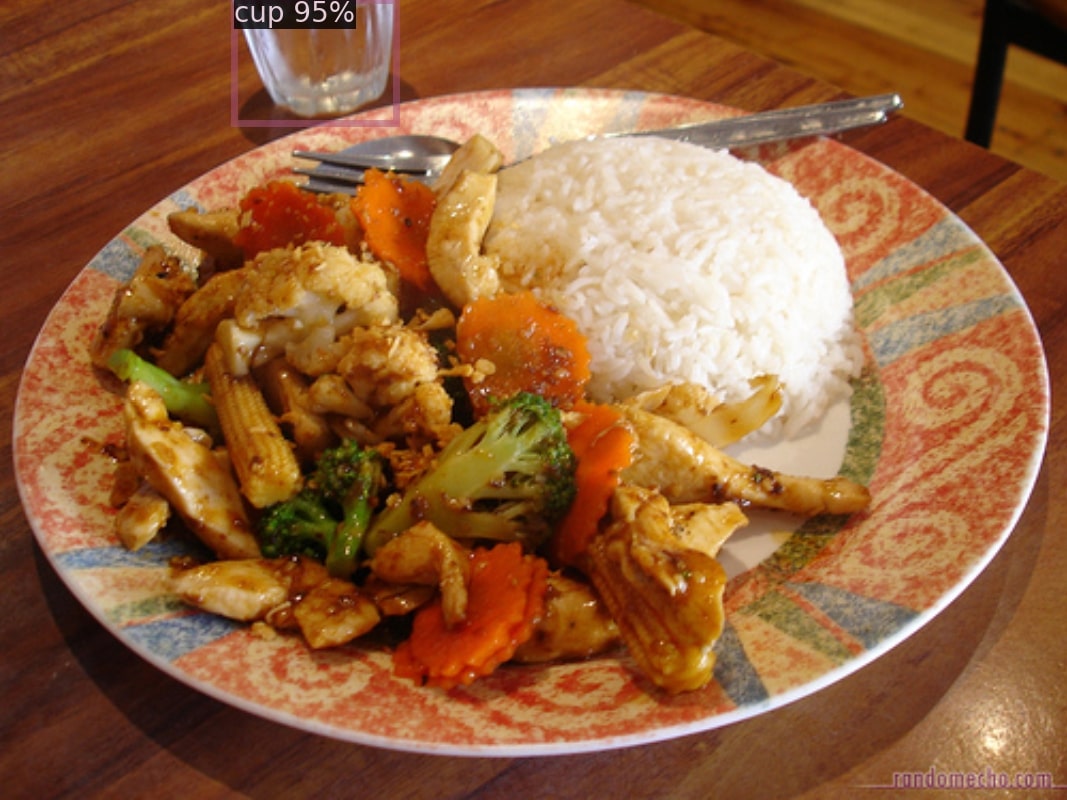}&
\includegraphics[width=0.24\columnwidth]{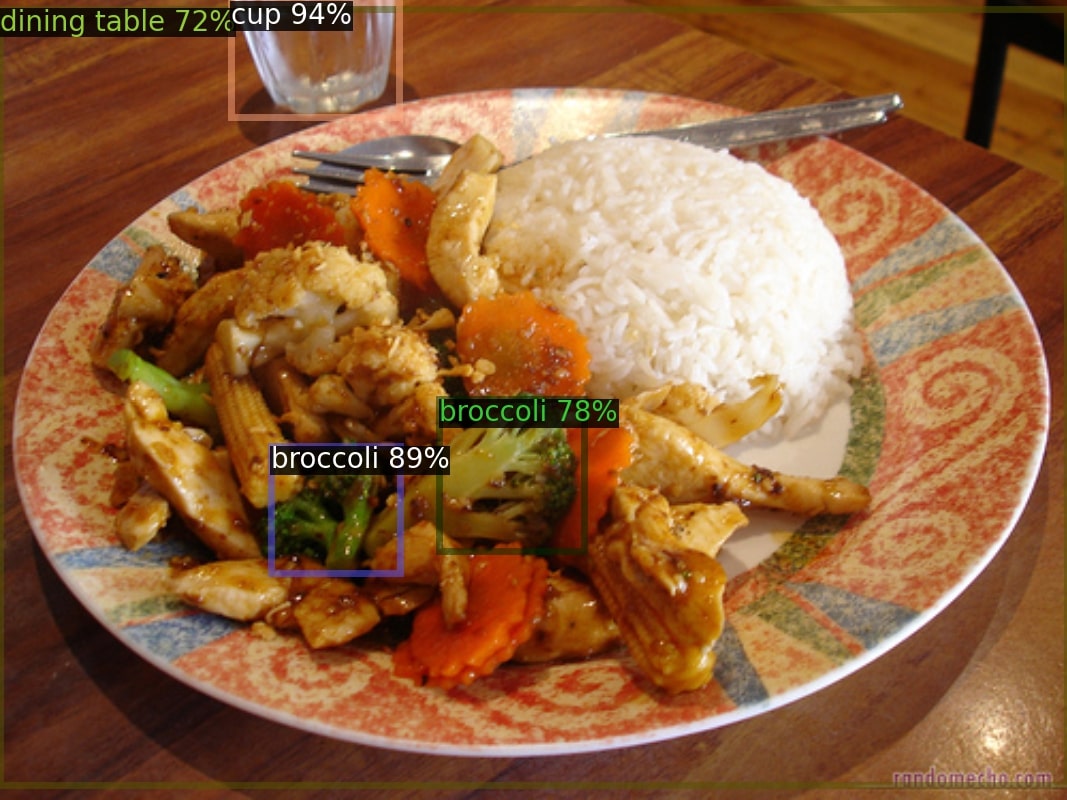} &
\includegraphics[width=0.24\columnwidth]{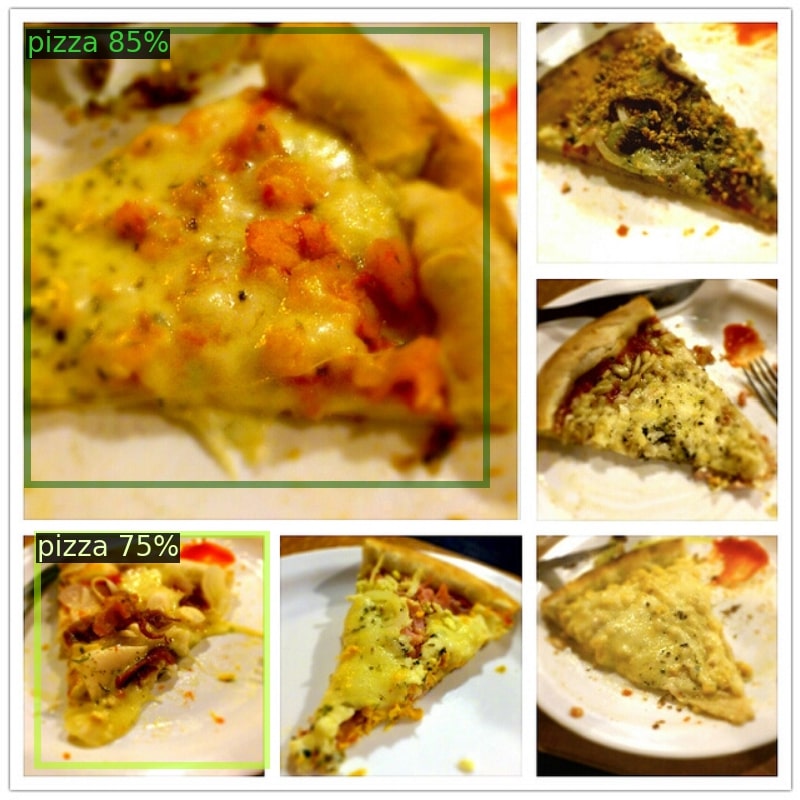}&
\includegraphics[width=0.24\columnwidth]{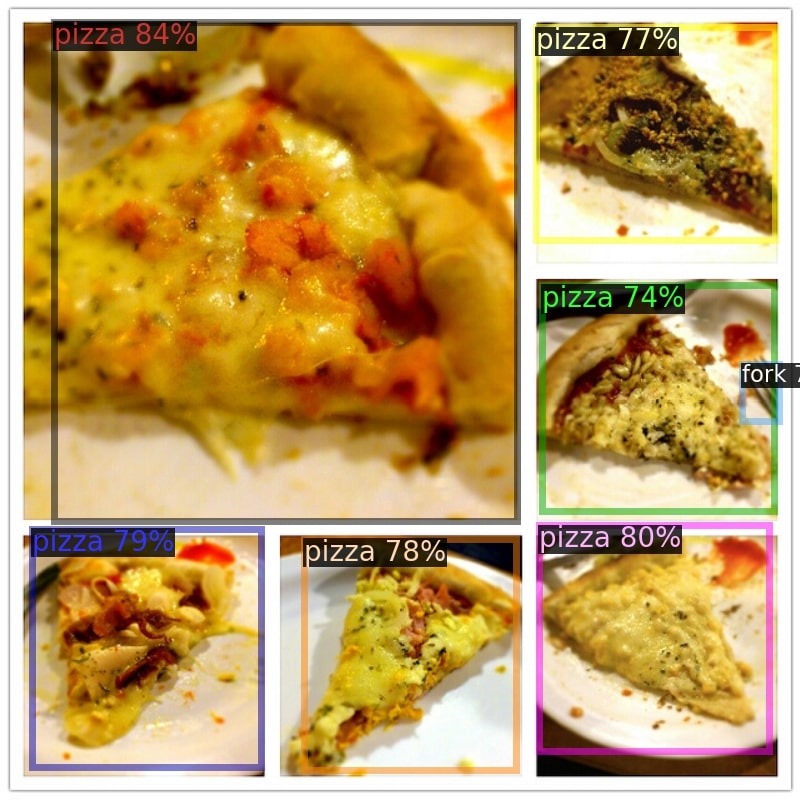} \\

\includegraphics[width=0.24\columnwidth]{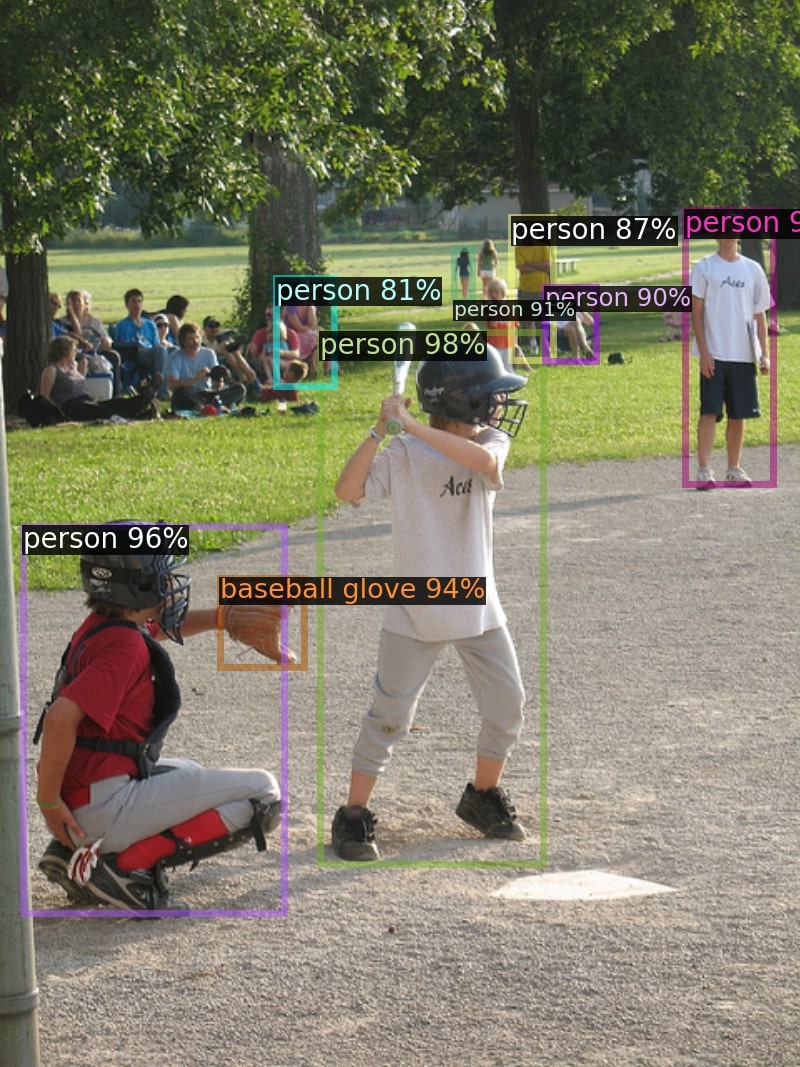}&
\includegraphics[width=0.24\columnwidth]{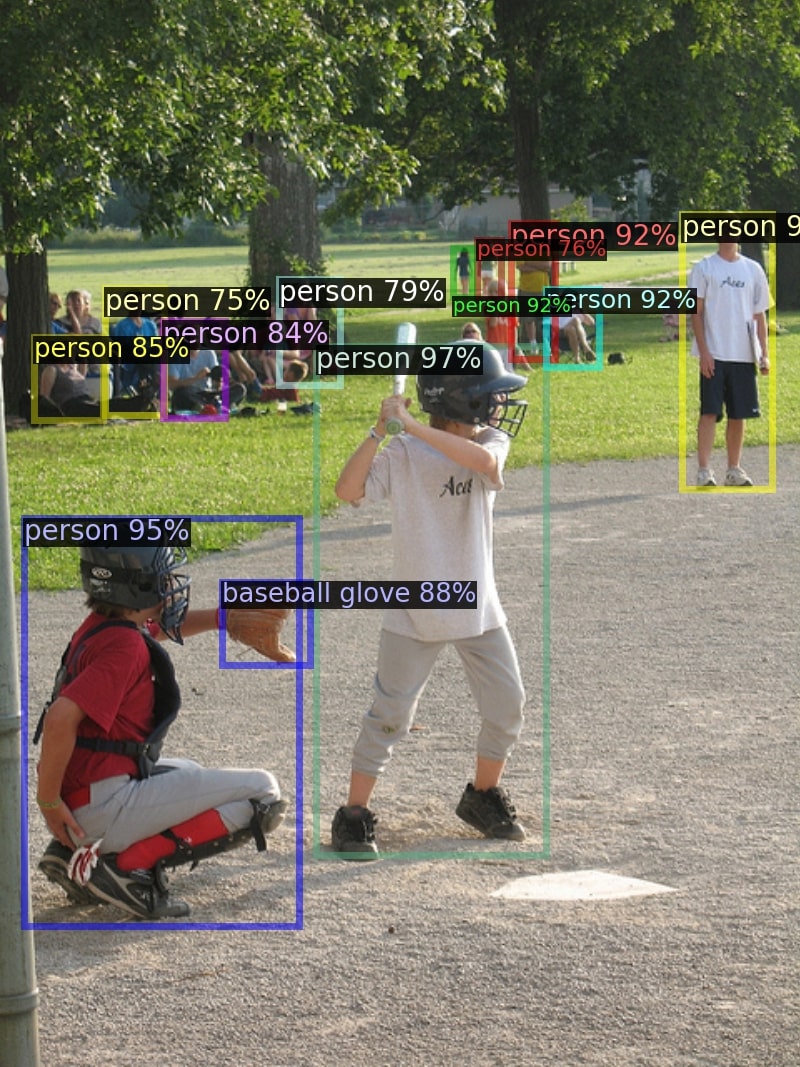} &
\includegraphics[width=0.24\columnwidth]{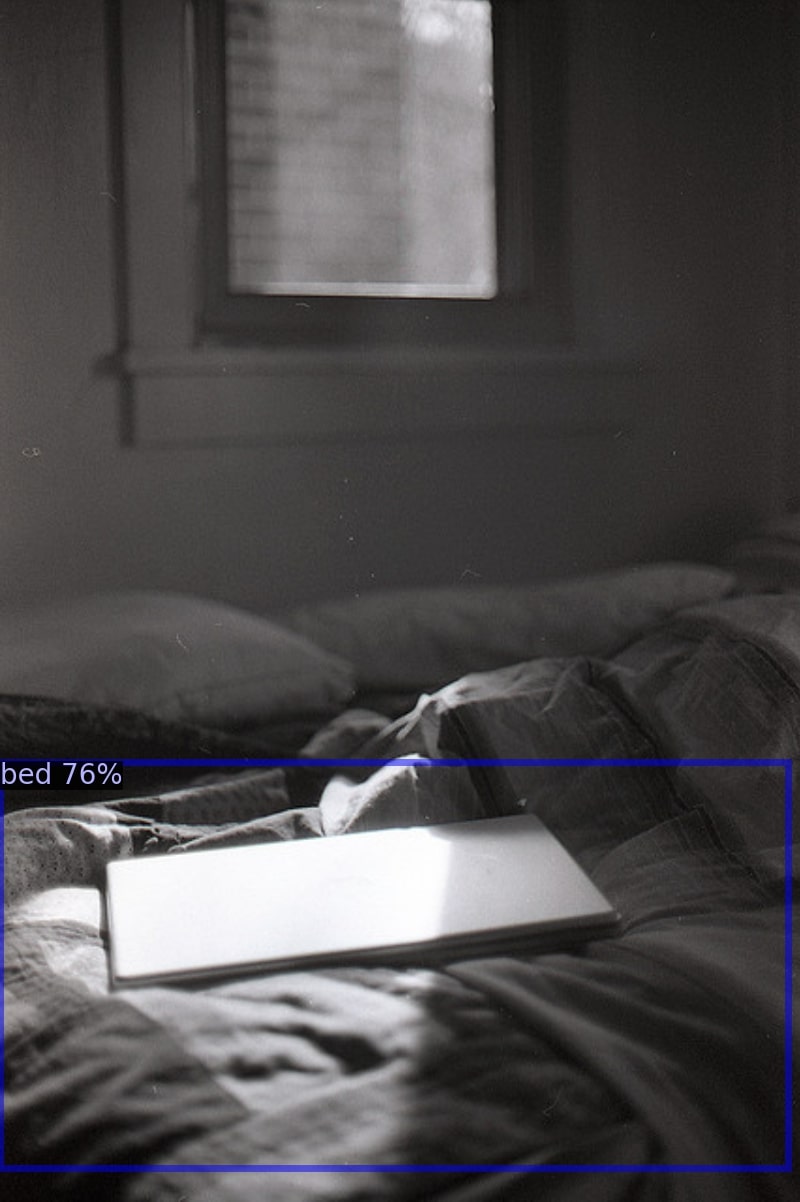}&
\includegraphics[width=0.24\columnwidth]{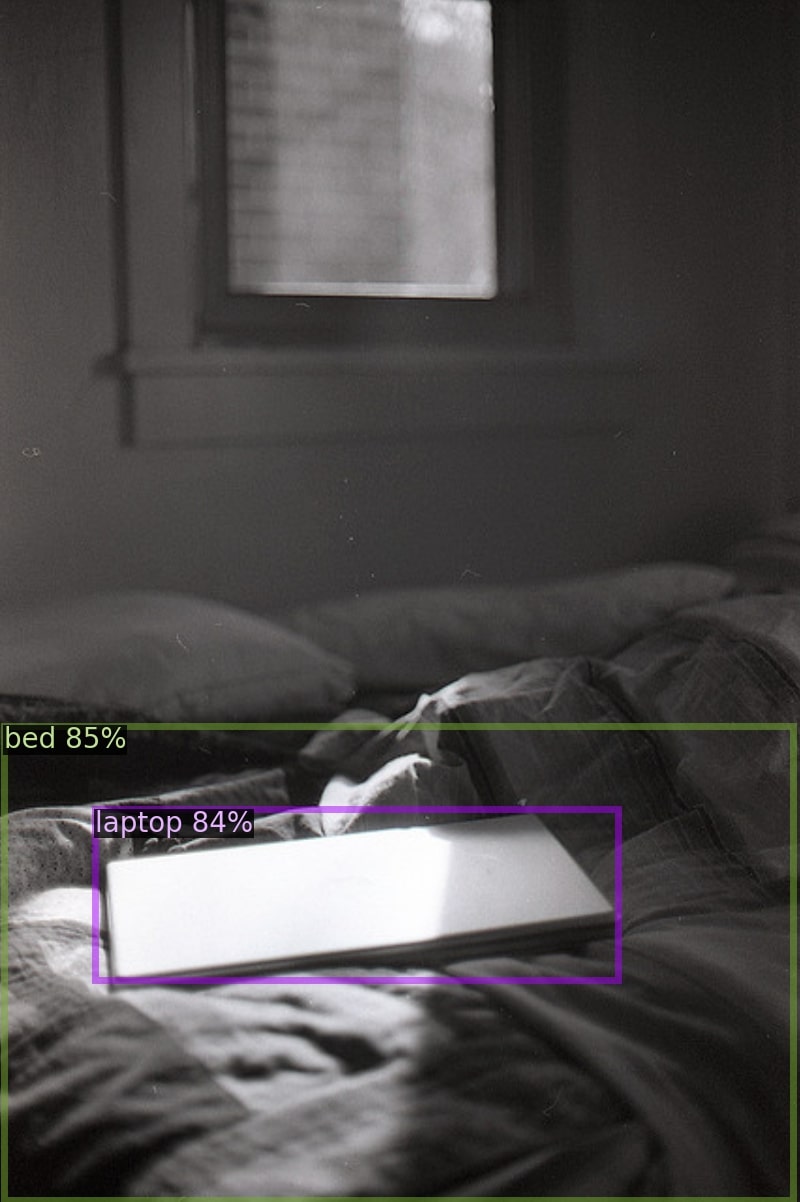} \\

\includegraphics[width=0.24\columnwidth]{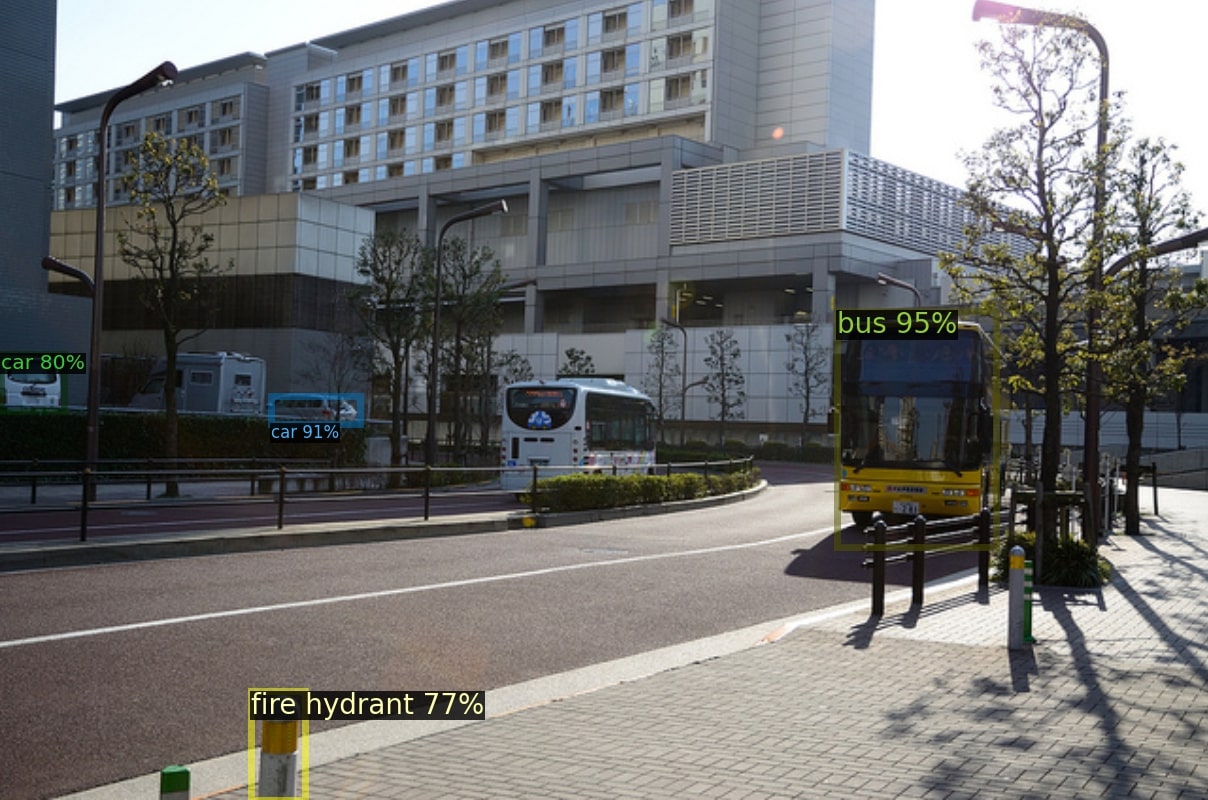}&
\includegraphics[width=0.24\columnwidth]{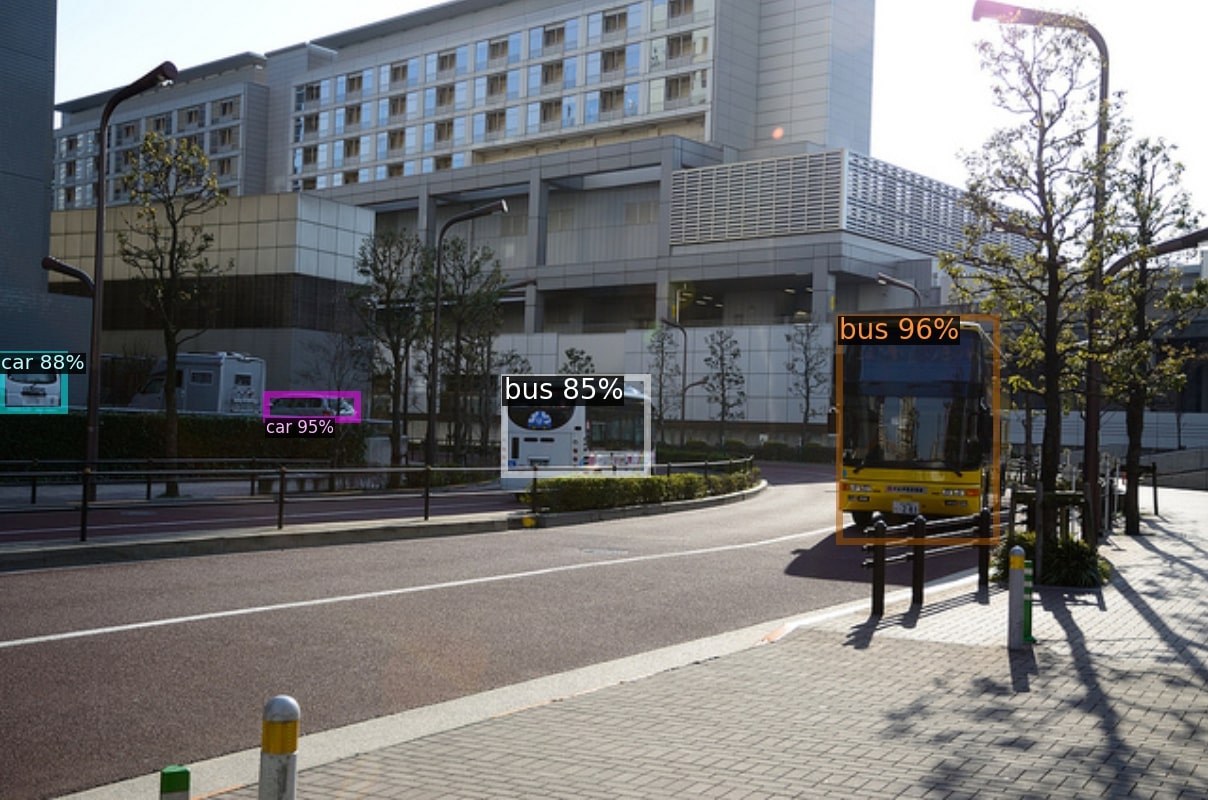} &
\includegraphics[width=0.24\columnwidth]{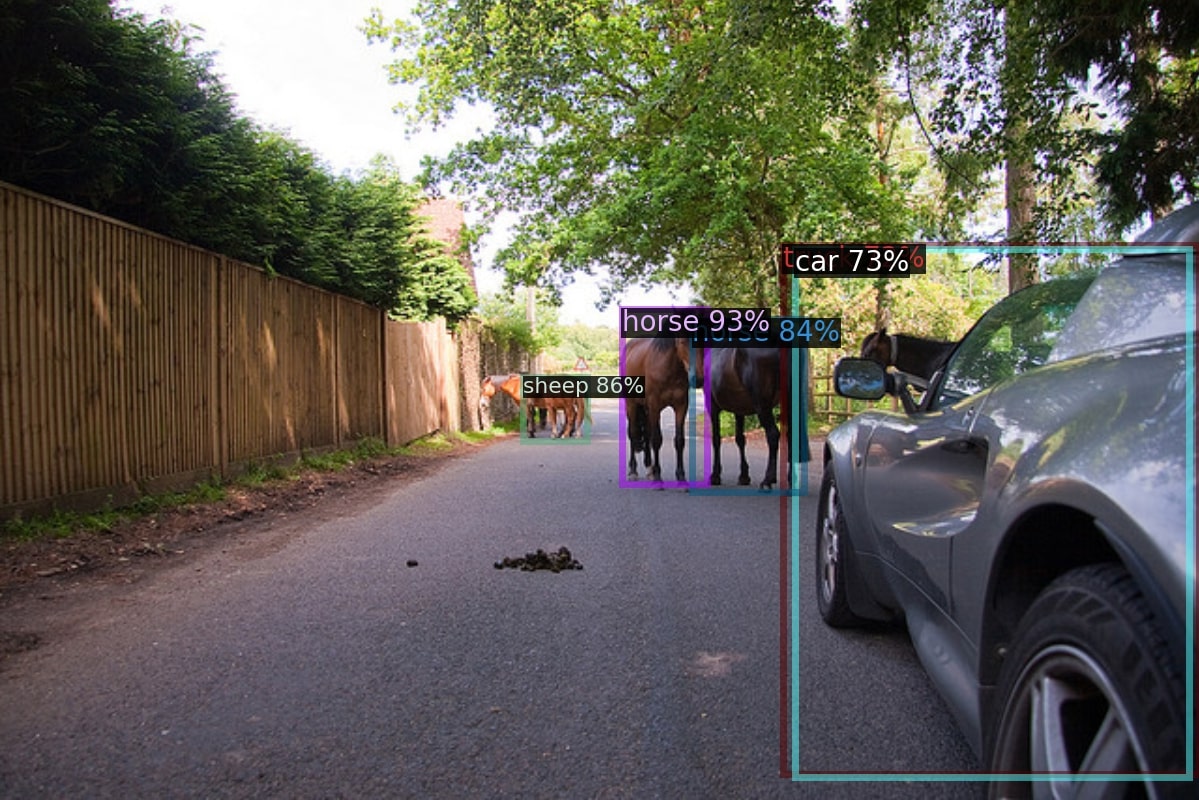}&
\includegraphics[width=0.24\columnwidth]{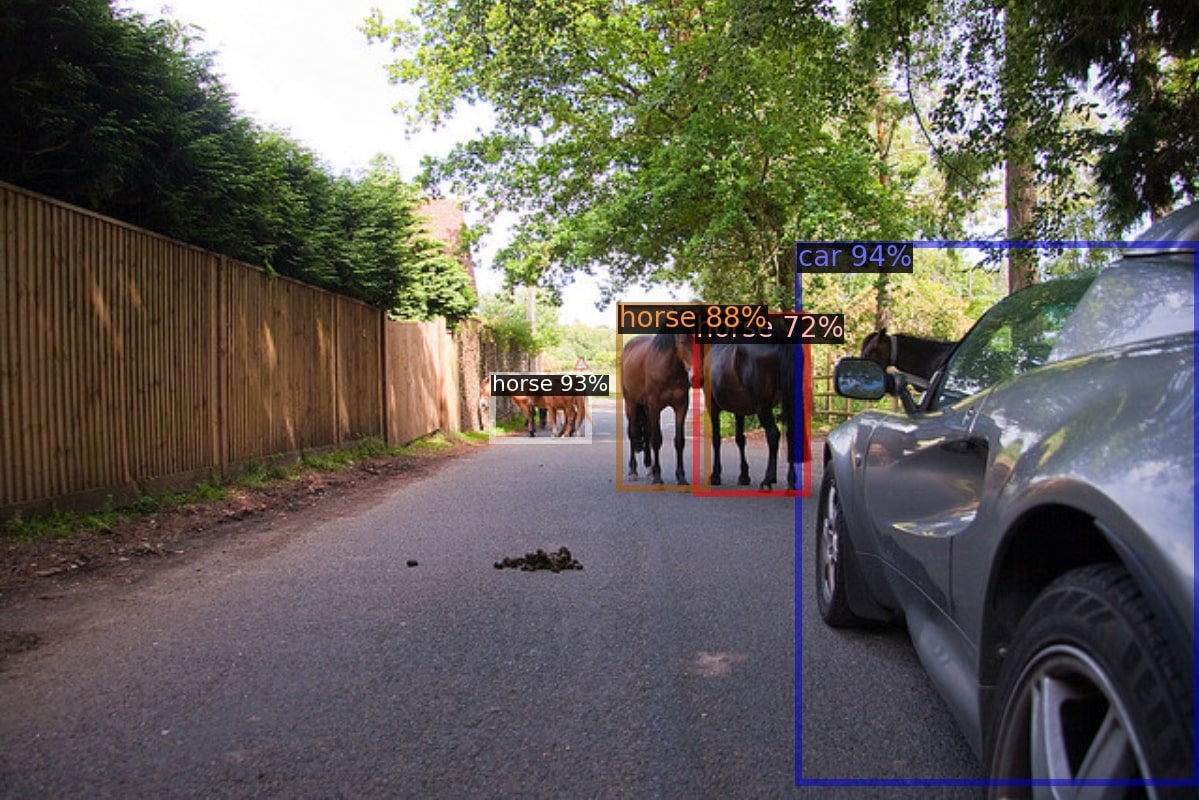} \\

\includegraphics[width=0.24\columnwidth]{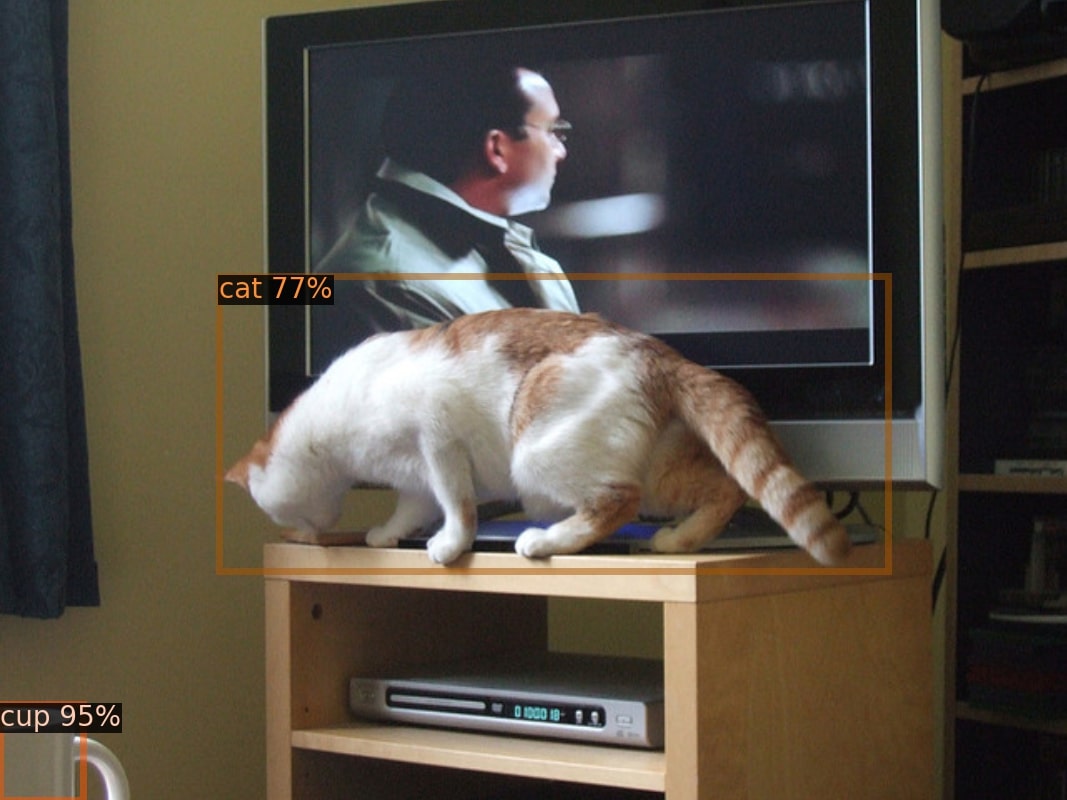}&
\includegraphics[width=0.24\columnwidth]{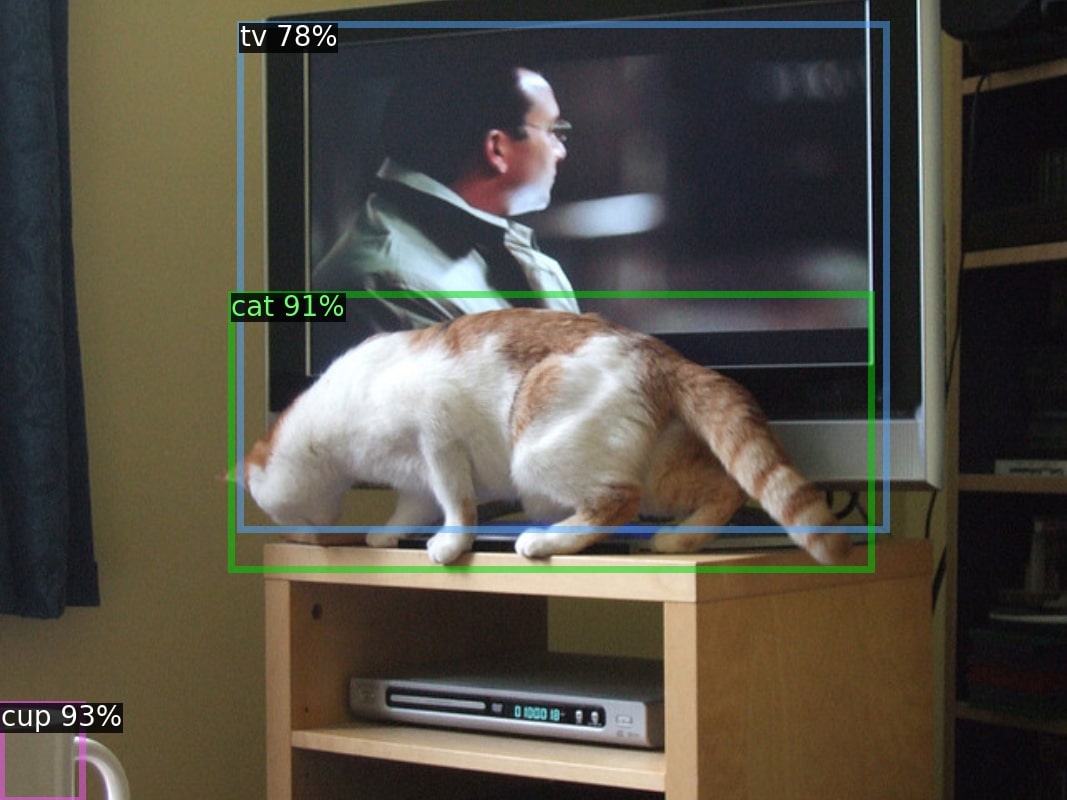} &
\includegraphics[width=0.24\columnwidth]{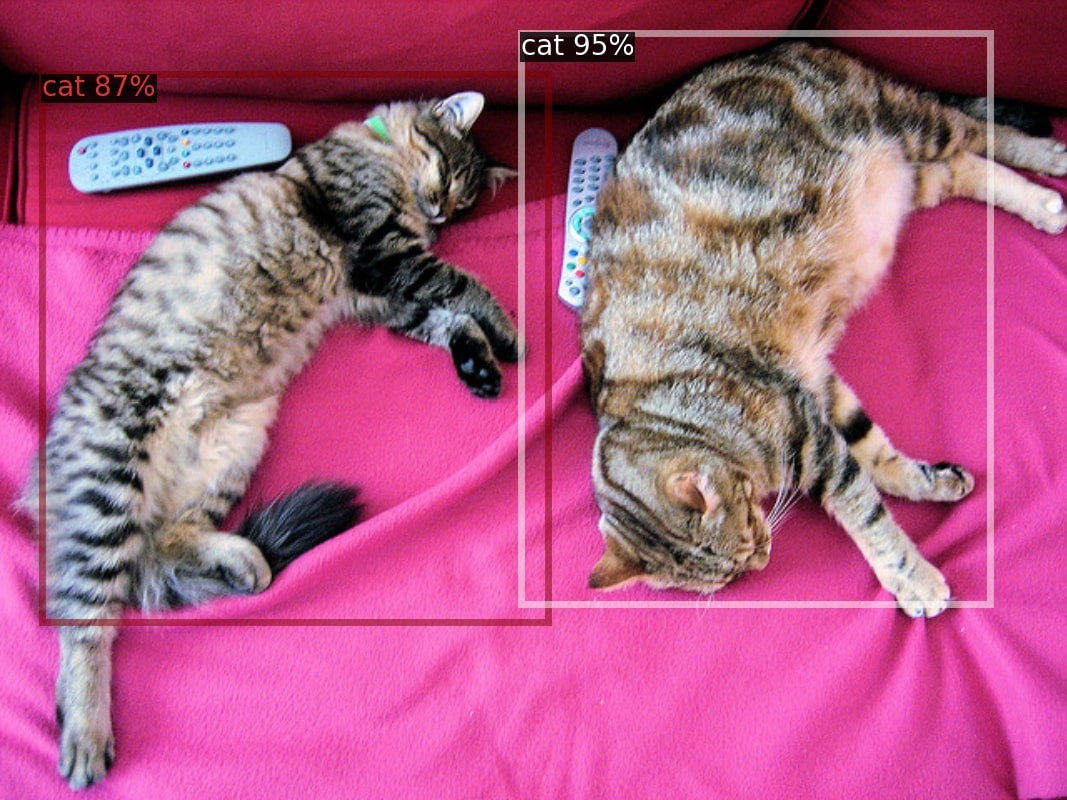}&
\includegraphics[width=0.24\columnwidth]{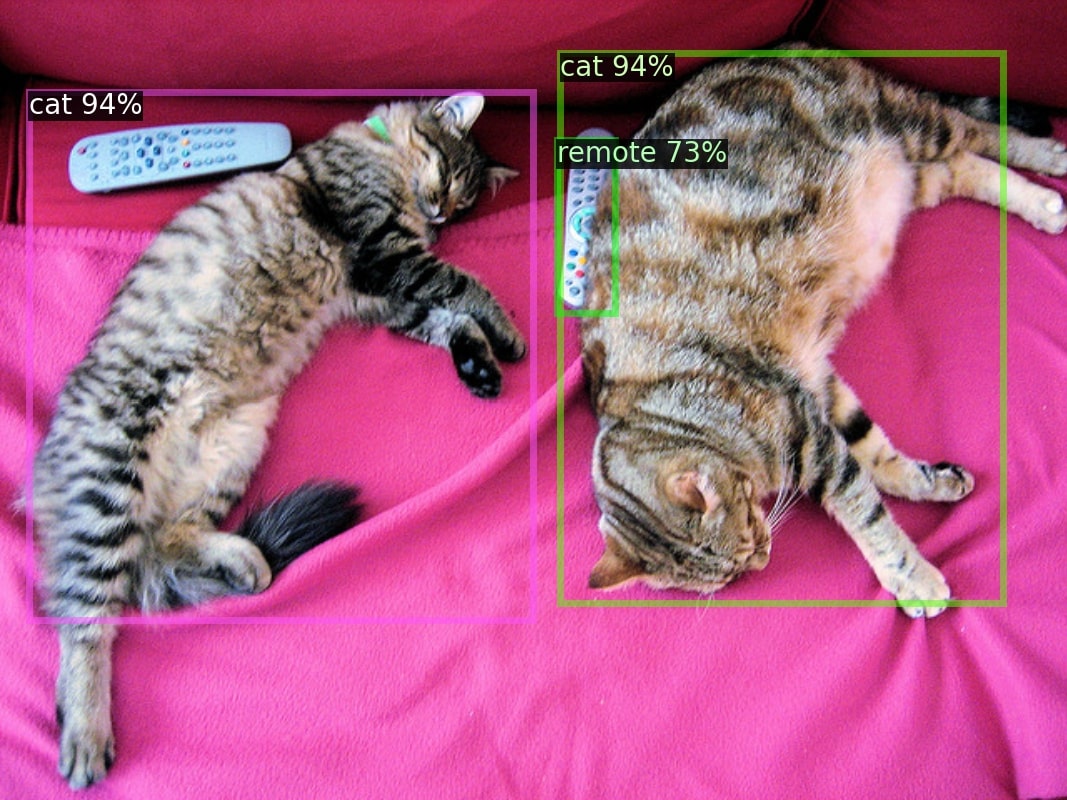} \\

\includegraphics[width=0.24\columnwidth]{./baseline/439.jpg}&
\includegraphics[width=0.24\columnwidth]{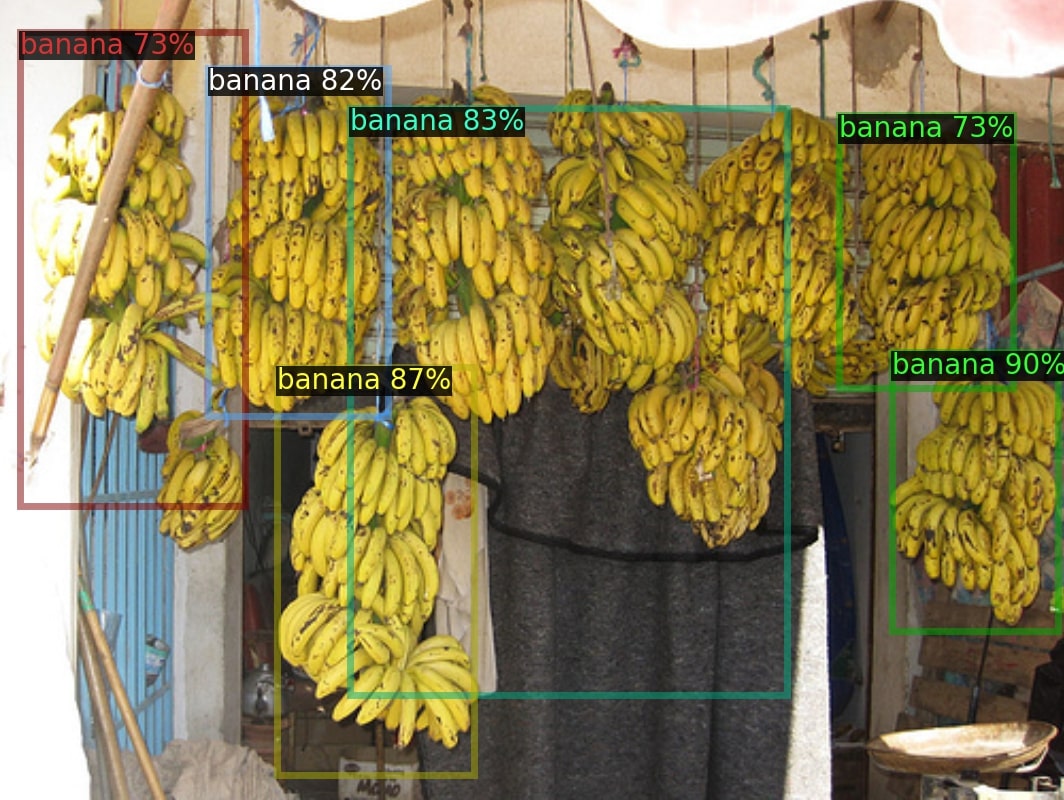} &
\includegraphics[width=0.24\columnwidth]{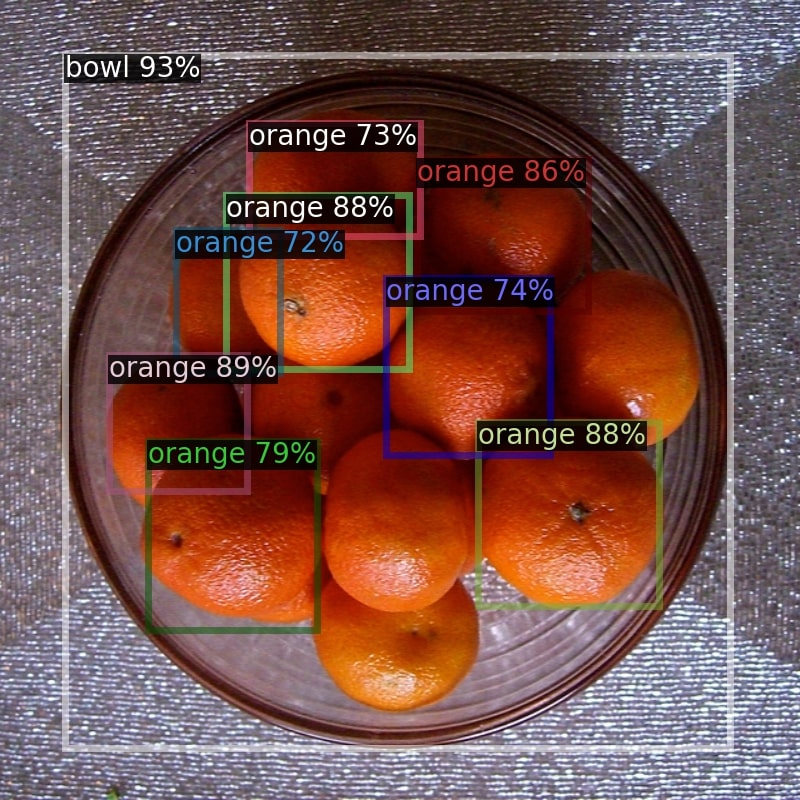}&
\includegraphics[width=0.24\columnwidth]{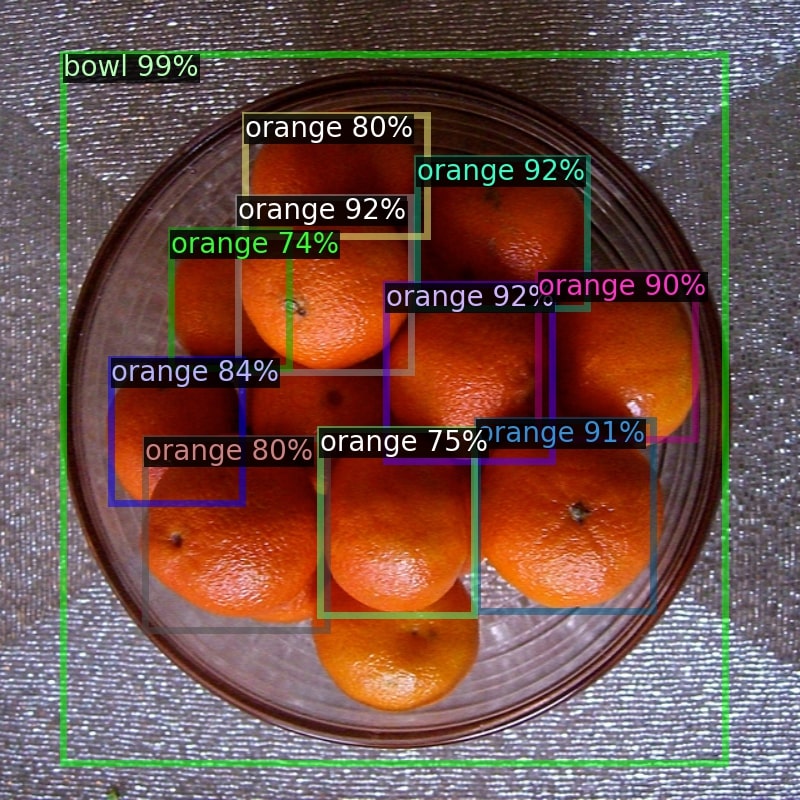} \\

\end{tabular}
\caption{Qualitative results of the baseline~\cite{liu2021unbiased} and our framework~(best viewed with 300\% zoom-in). We set the threshold of classification scores as 0.7 to visualize the results.}
\label{fig:qualitative}
\end{figure}

\begin{figure}[h]
\begin{tabular}{c@{}c@{}c@{}c}
Baseline & Ours & Baseline & Ours \\
\includegraphics[width=0.24\columnwidth]{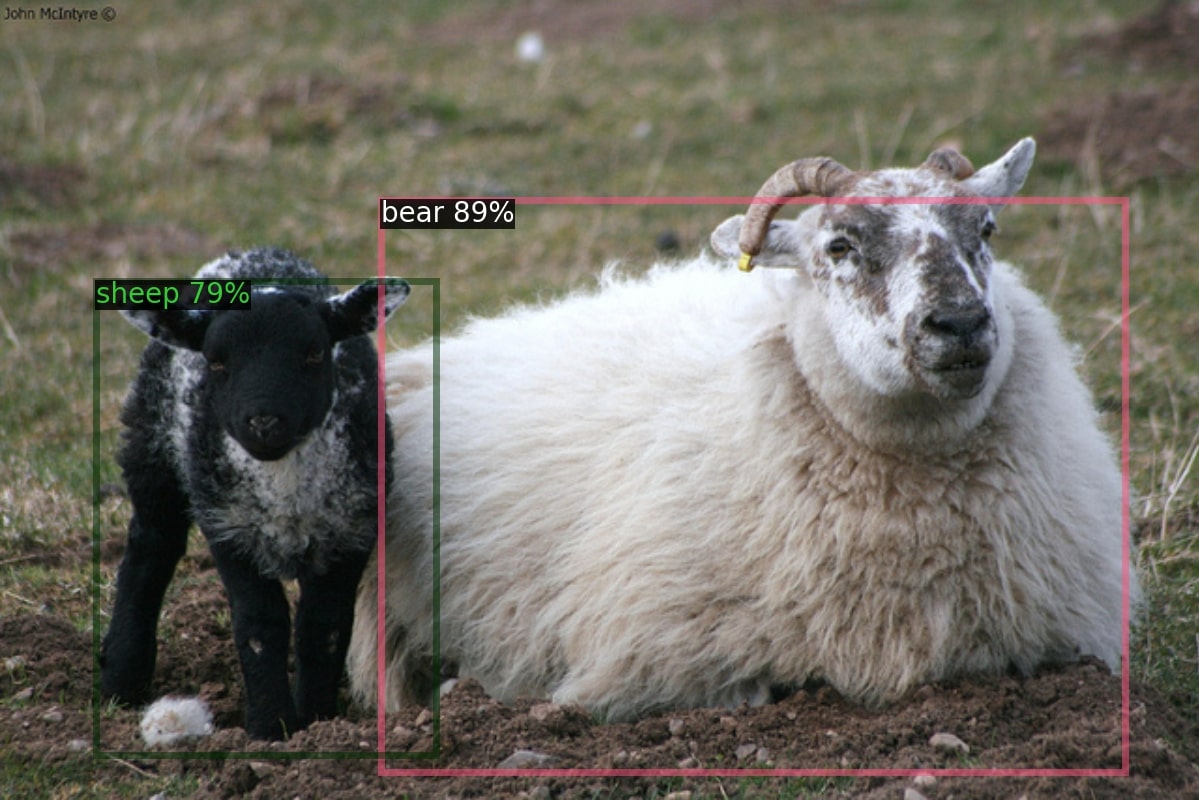}&
\includegraphics[width=0.24\columnwidth]{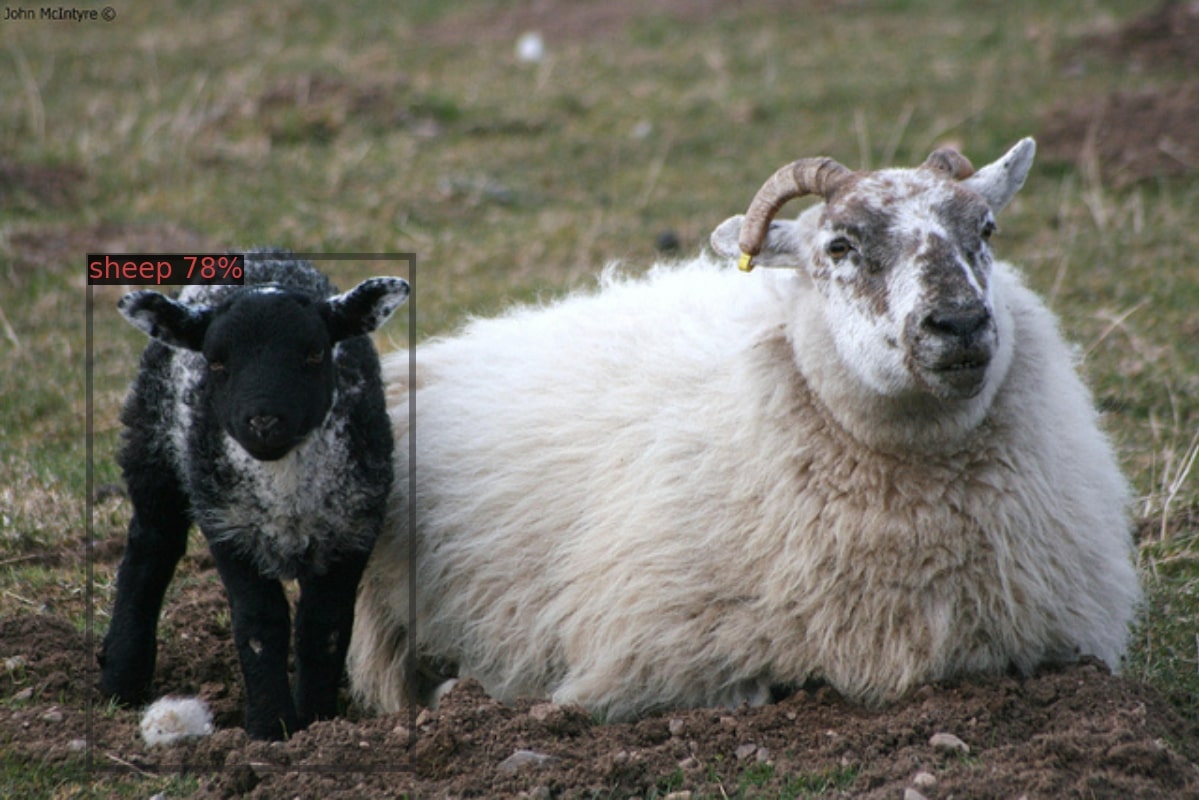} &
\includegraphics[width=0.24\columnwidth]{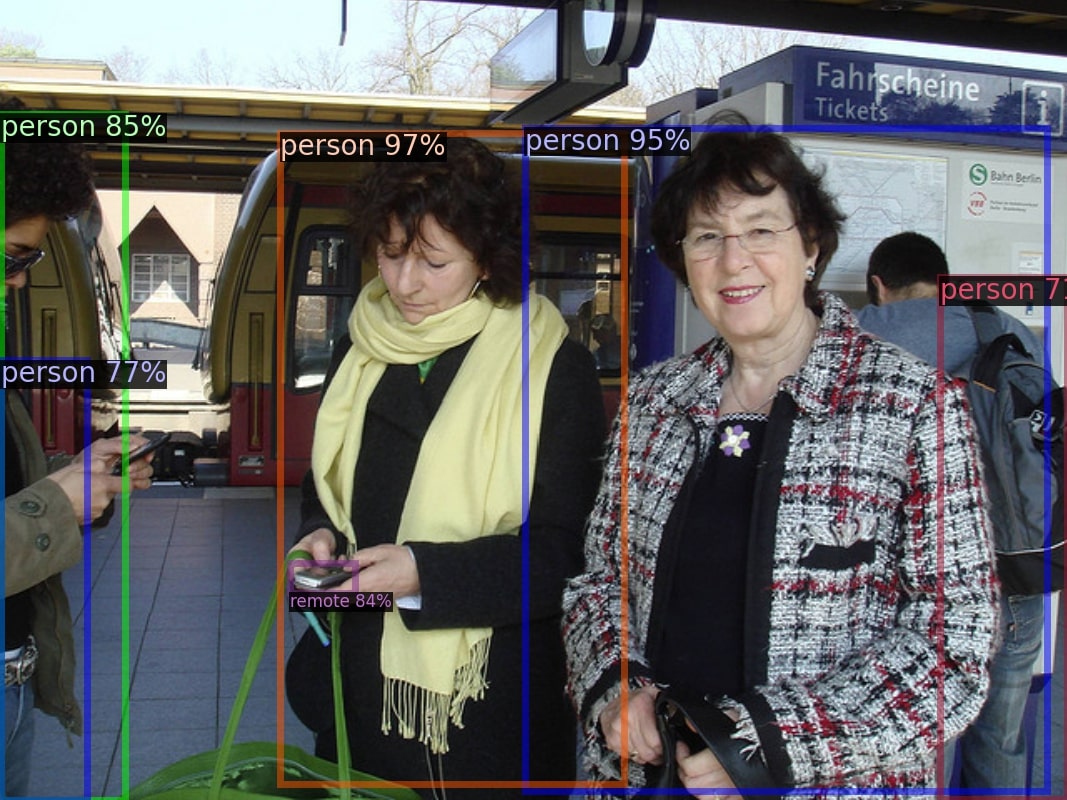}&
\includegraphics[width=0.24\columnwidth]{./ours/638.jpg} \\

\includegraphics[width=0.24\columnwidth]{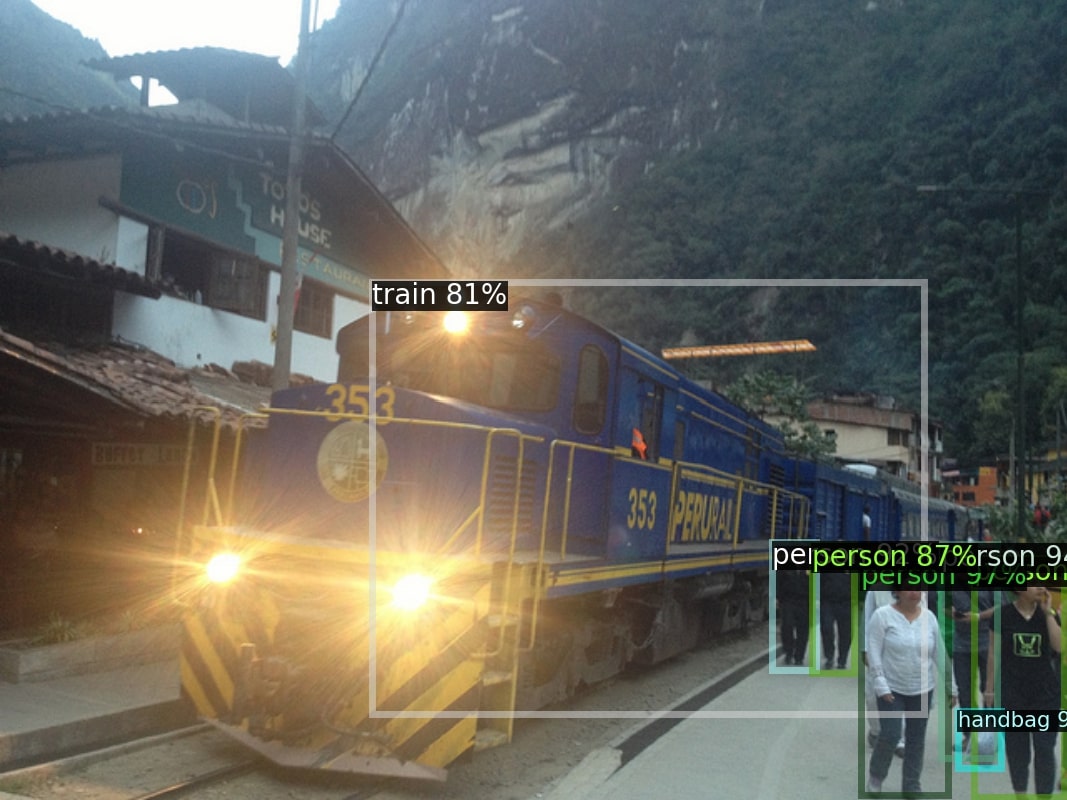}&
\includegraphics[width=0.24\columnwidth]{./ours/652.jpg} &
\includegraphics[width=0.24\columnwidth]{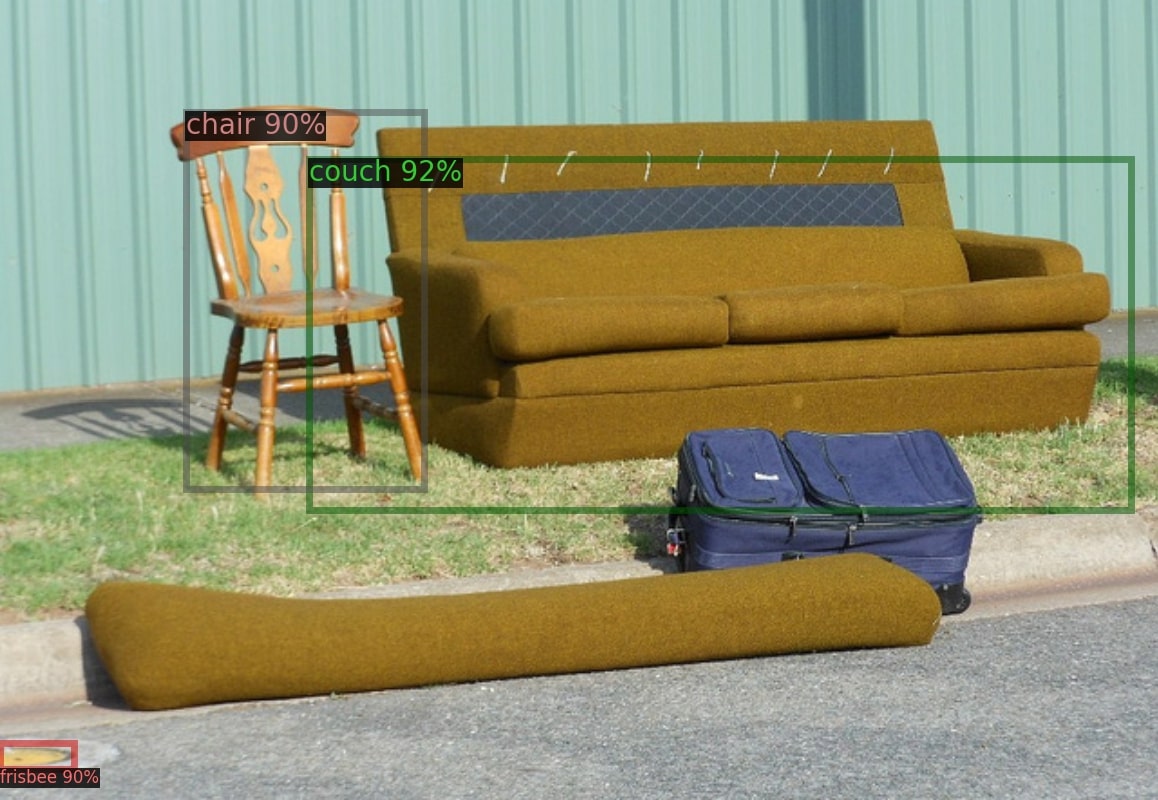}&
\includegraphics[width=0.24\columnwidth]{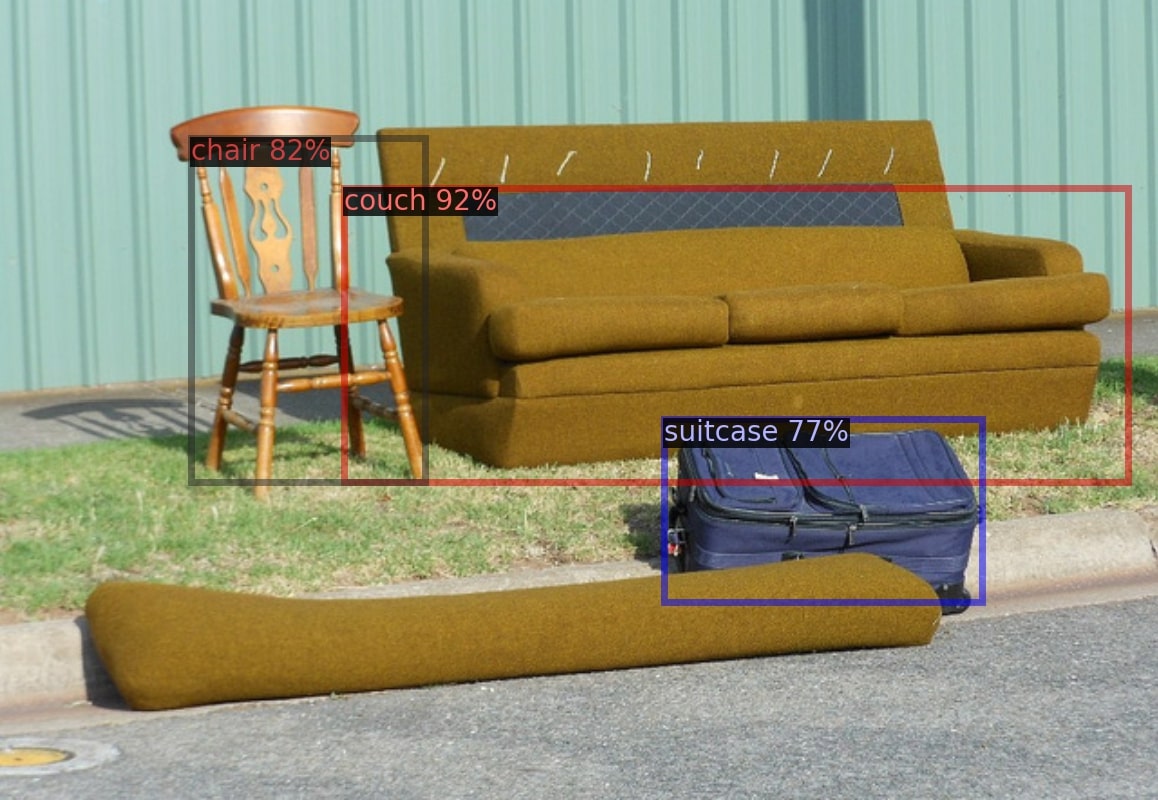} \\

\includegraphics[width=0.24\columnwidth]{./baseline/694.jpg}&
\includegraphics[width=0.24\columnwidth]{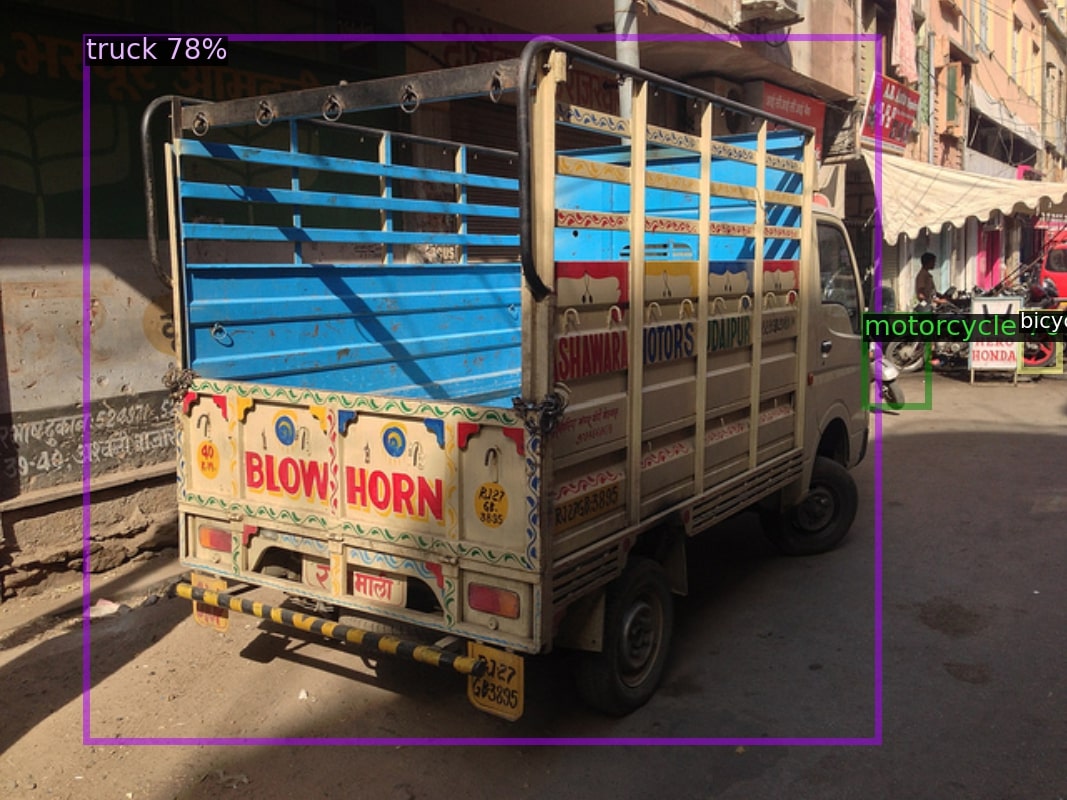} &
\includegraphics[width=0.24\columnwidth]{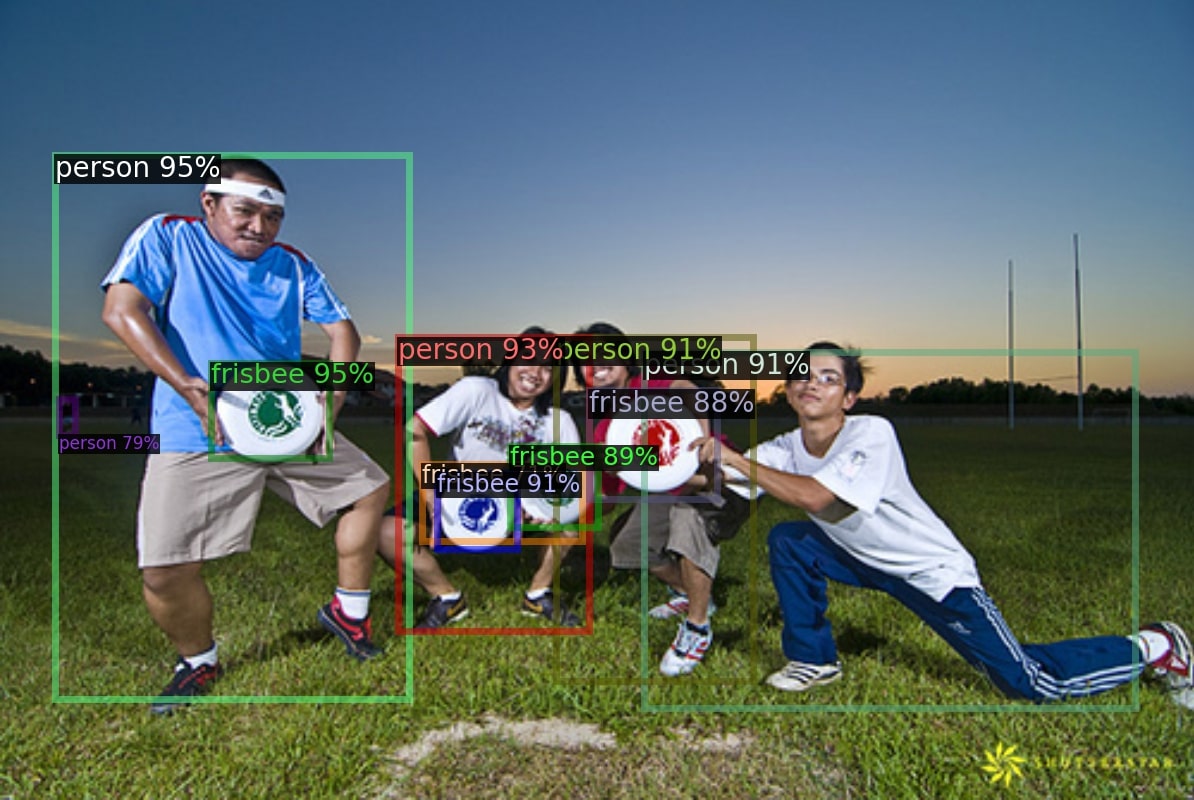}&
\includegraphics[width=0.24\columnwidth]{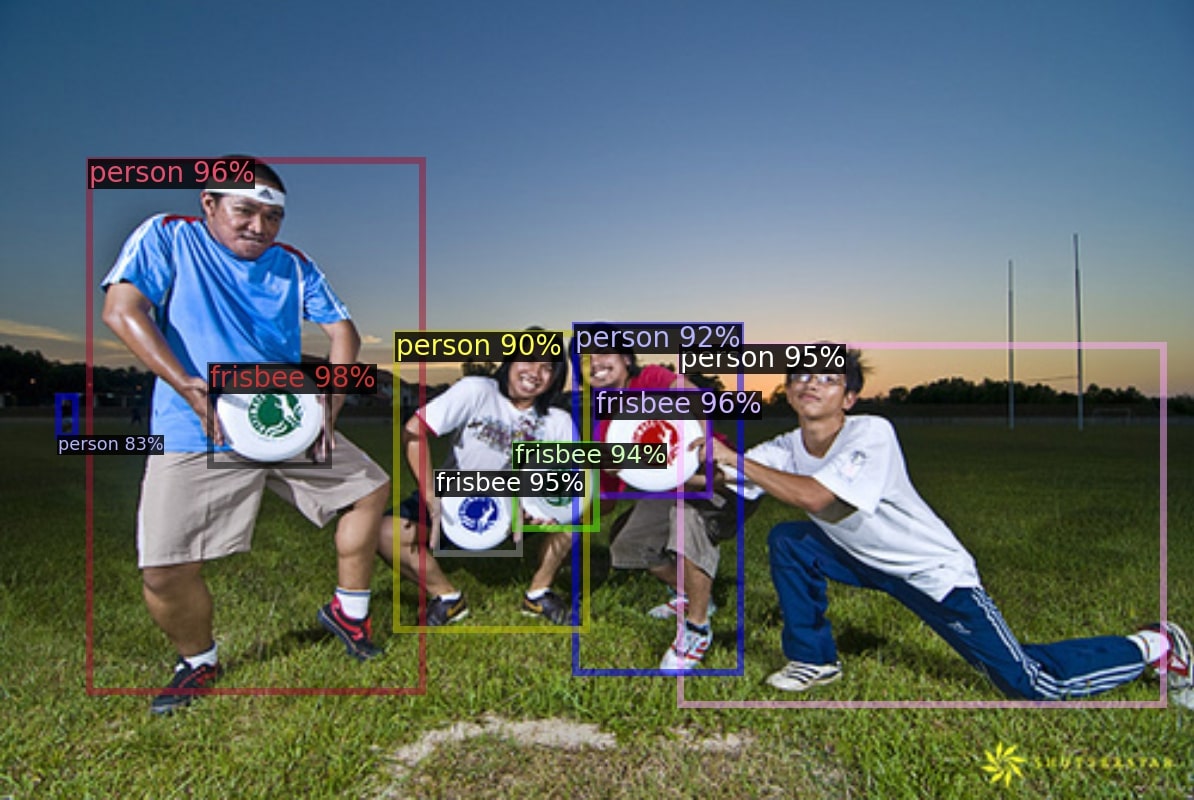} \\

\includegraphics[width=0.24\columnwidth]{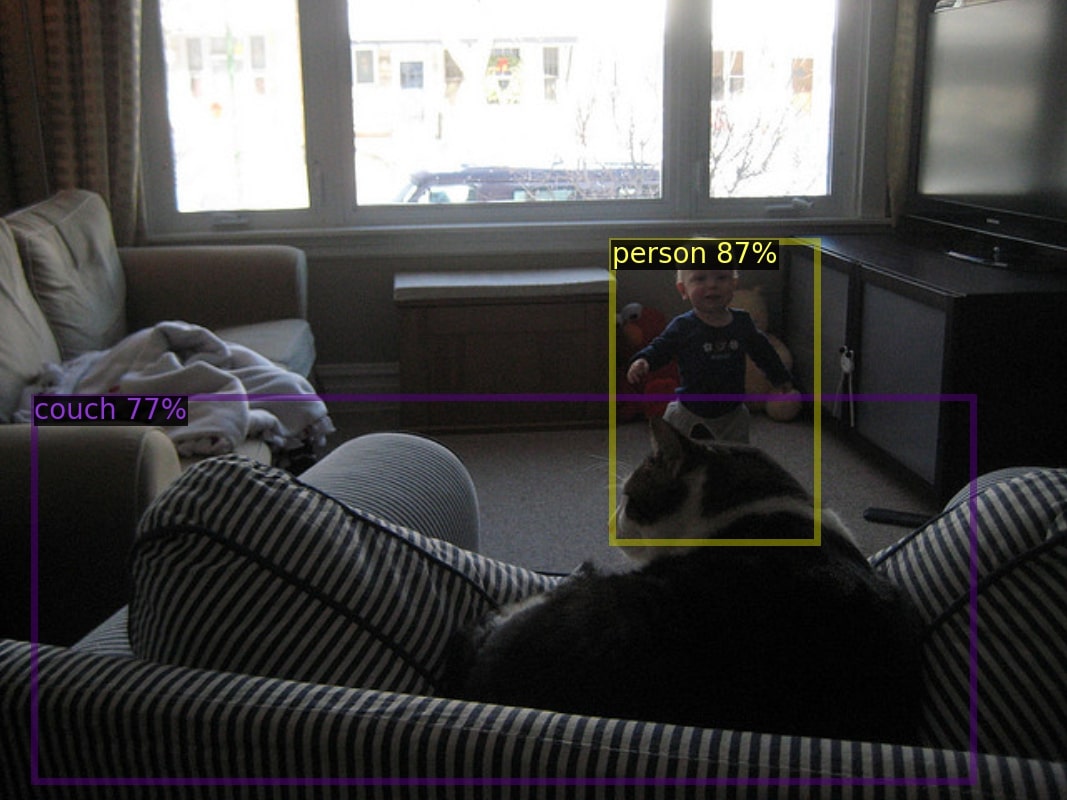}&
\includegraphics[width=0.24\columnwidth]{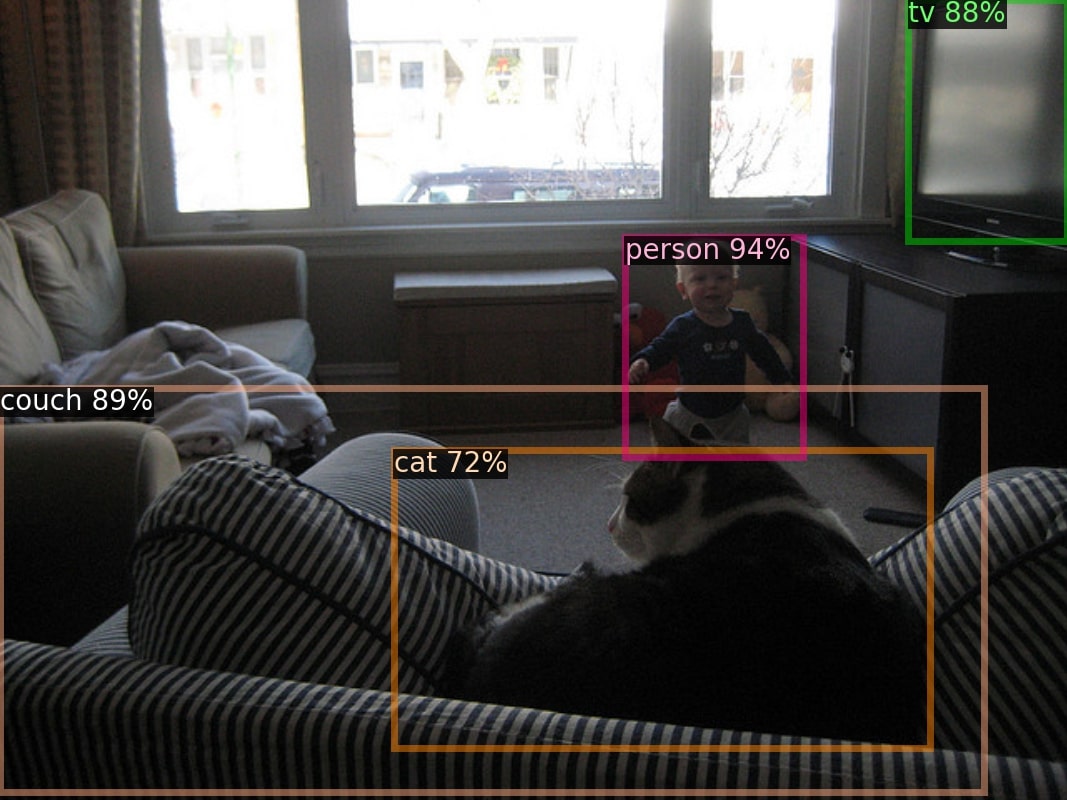} &
\includegraphics[width=0.24\columnwidth]{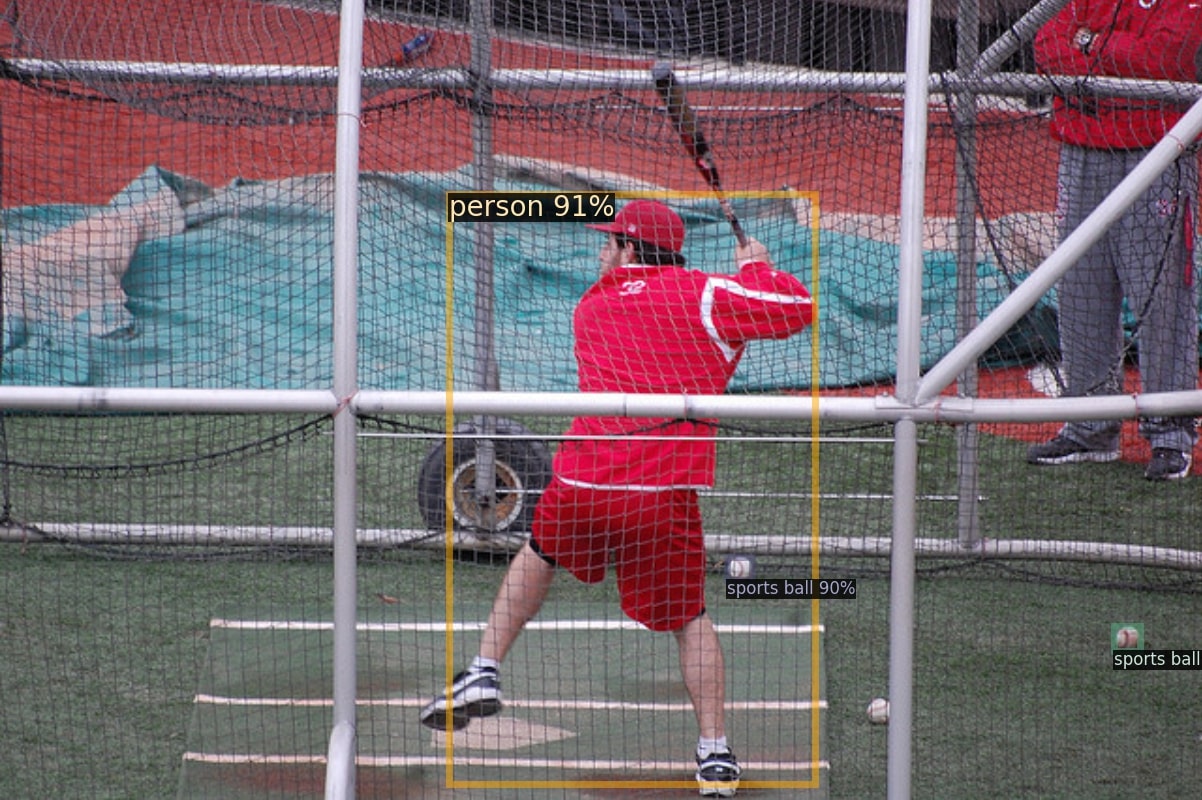}&
\includegraphics[width=0.24\columnwidth]{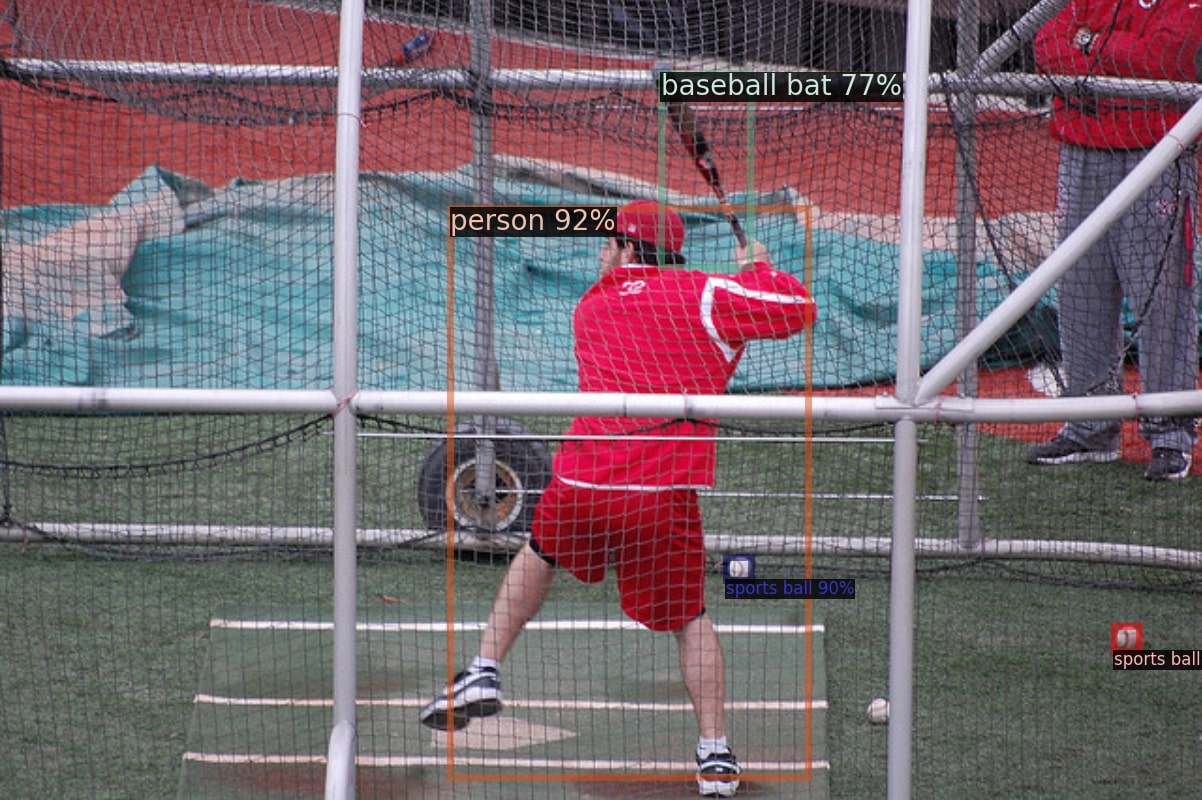} \\

\includegraphics[width=0.24\columnwidth]{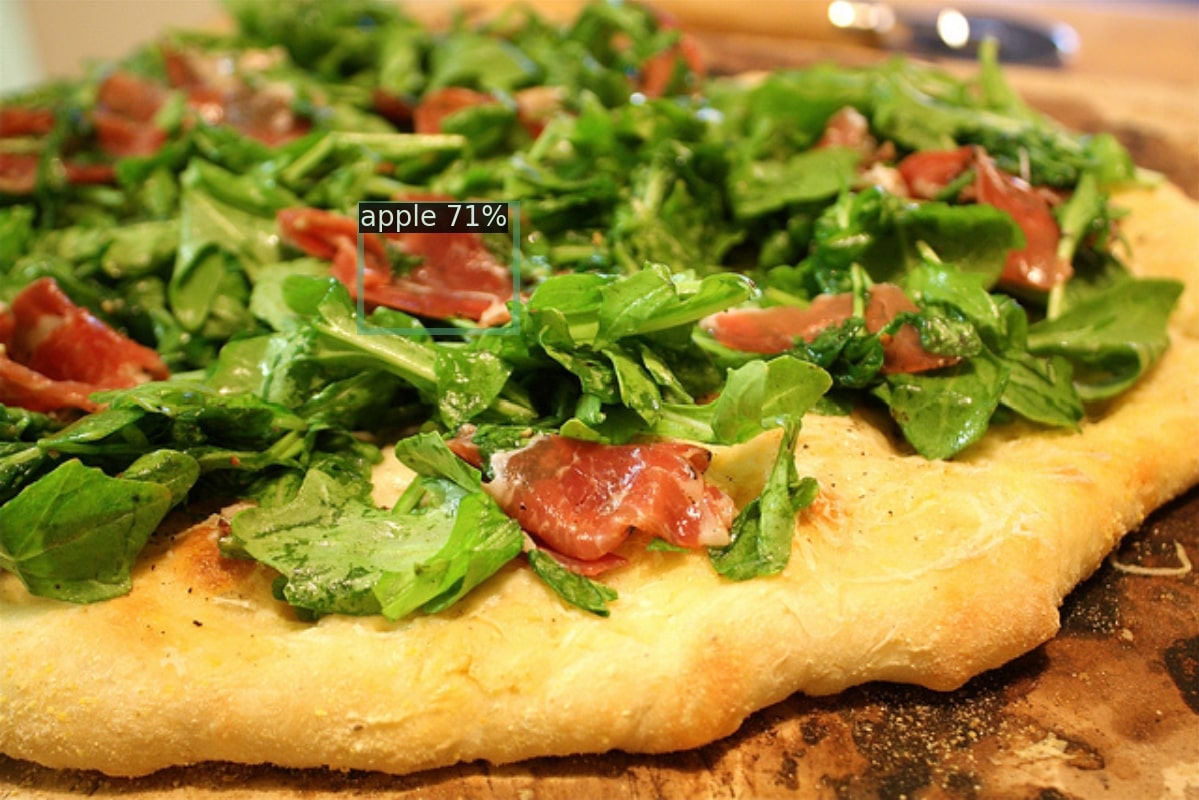}&
\includegraphics[width=0.24\columnwidth]{./ours/859.jpg} &
\includegraphics[width=0.24\columnwidth]{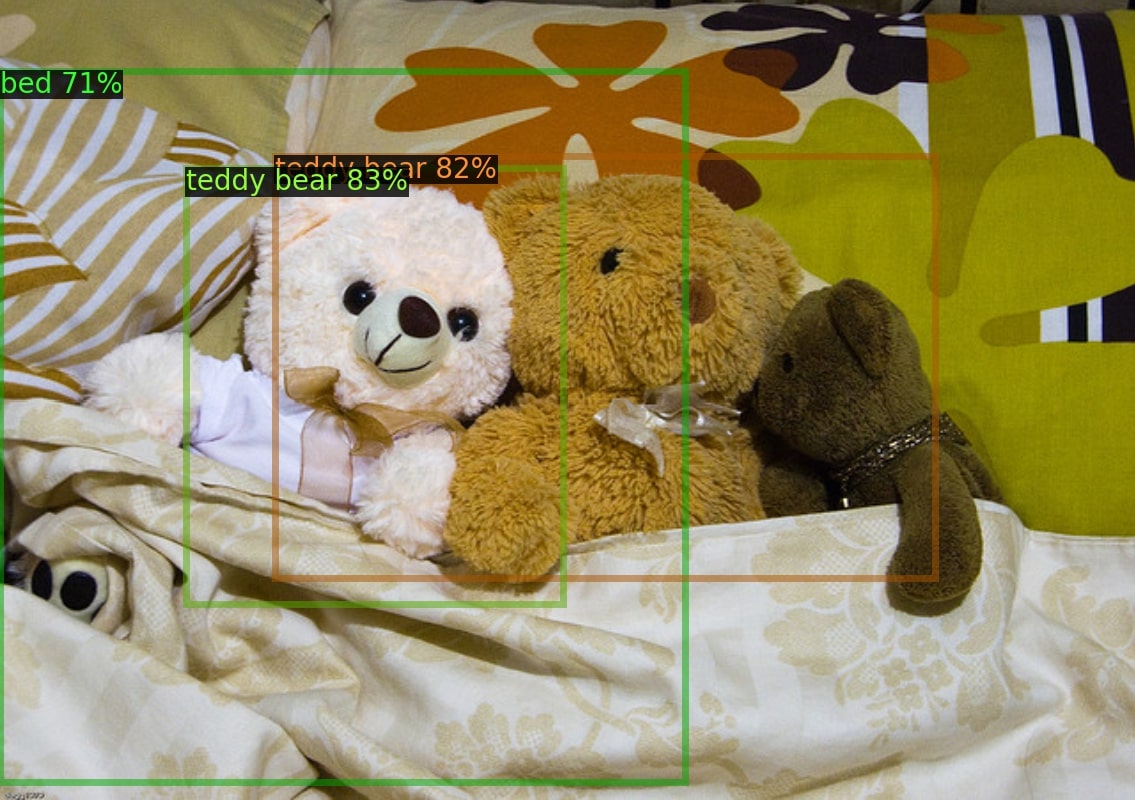}&
\includegraphics[width=0.24\columnwidth]{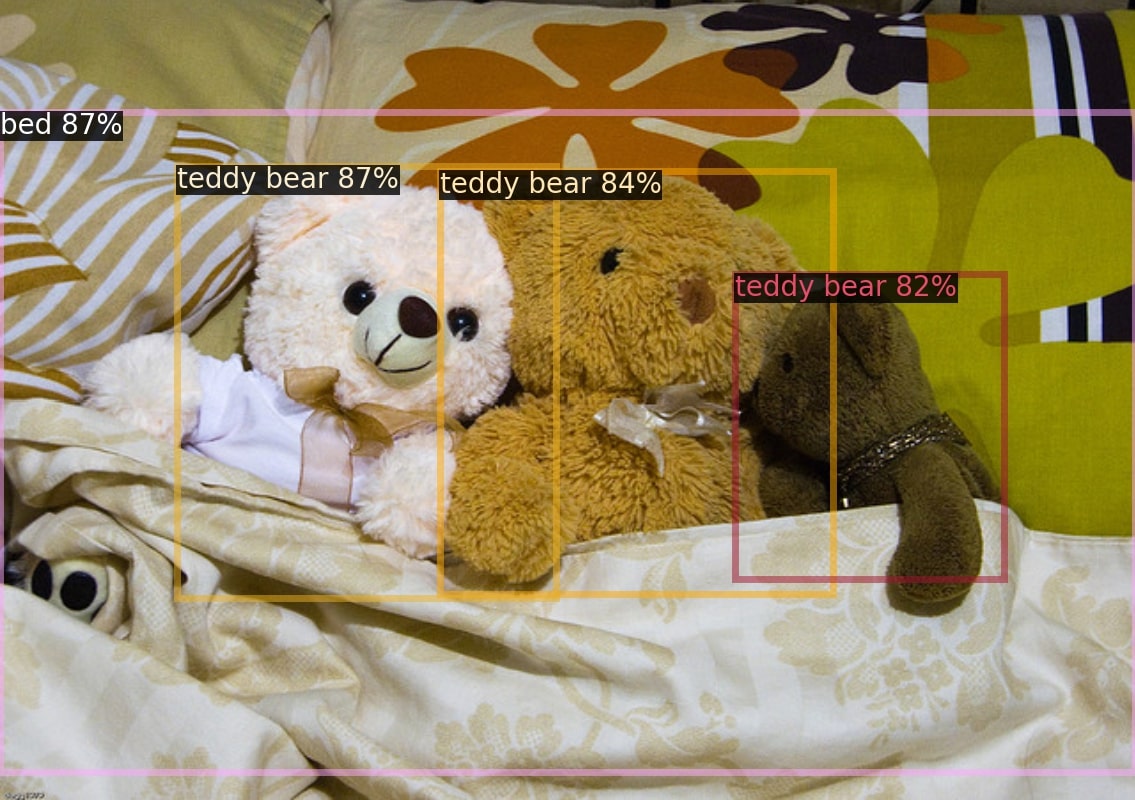} \\

\includegraphics[width=0.24\columnwidth]{./baseline/972.jpg}&
\includegraphics[width=0.24\columnwidth]{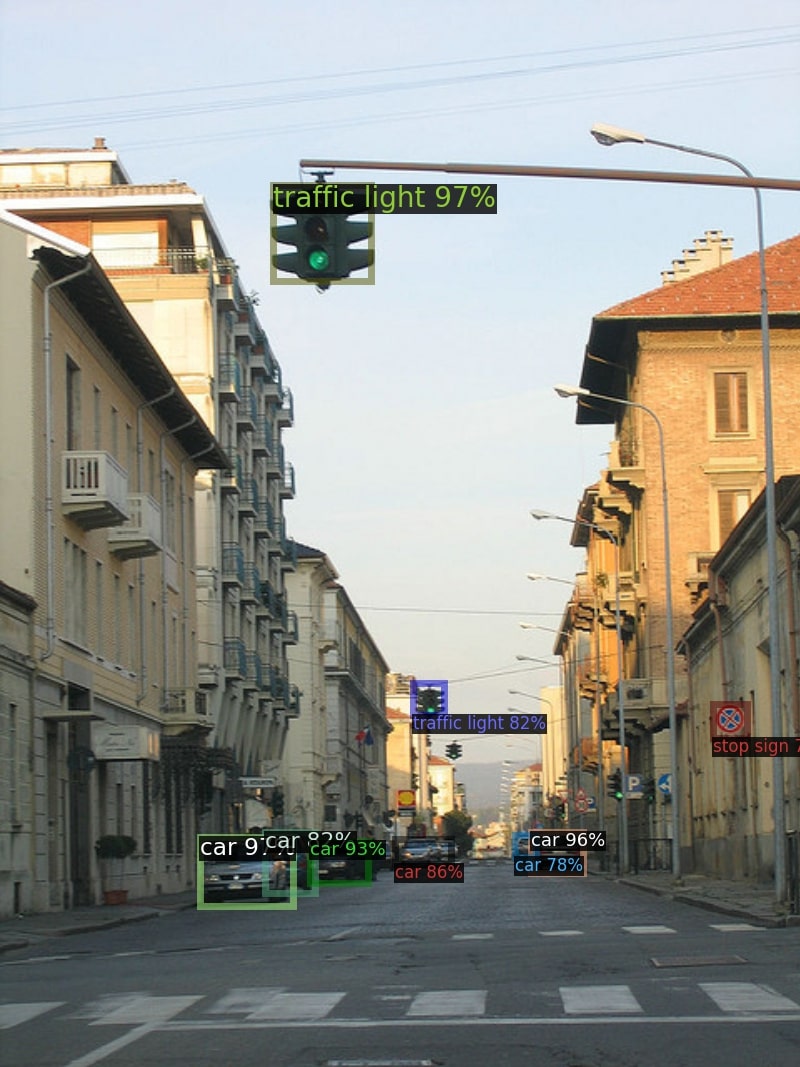} &
\includegraphics[width=0.24\columnwidth]{./baseline/1125.jpg}&
\includegraphics[width=0.24\columnwidth]{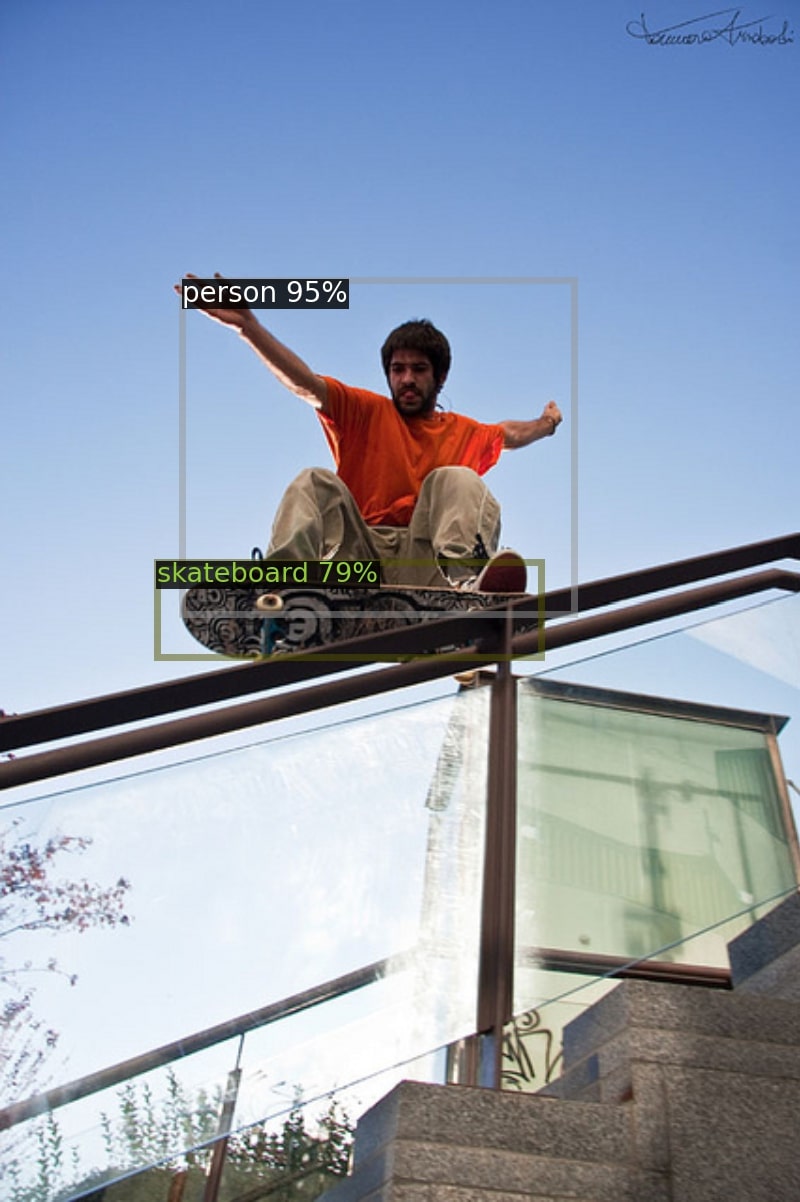} //

\end{tabular}
\caption{Qualitative results of the baseline~\cite{liu2021unbiased} and our framework~(best viewed with 300\% zoom-in). We set the threshold of classification scores as 0.7 to visualize the results. }
\label{fig:qualitative_2}
\end{figure}

\begin{figure}[h]
\begin{tabular}{c@{}c@{}c@{}c}
Baseline & Ours & Baseline & Ours \\
\includegraphics[width=0.24\columnwidth]{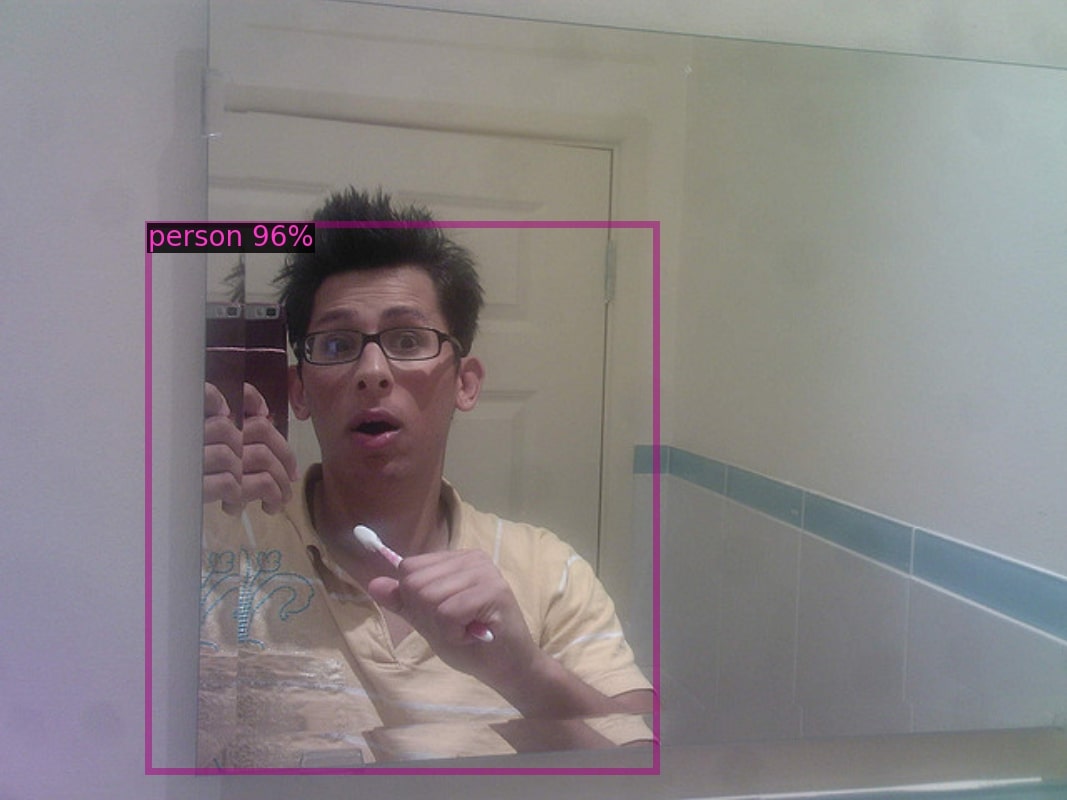}&
\includegraphics[width=0.24\columnwidth]{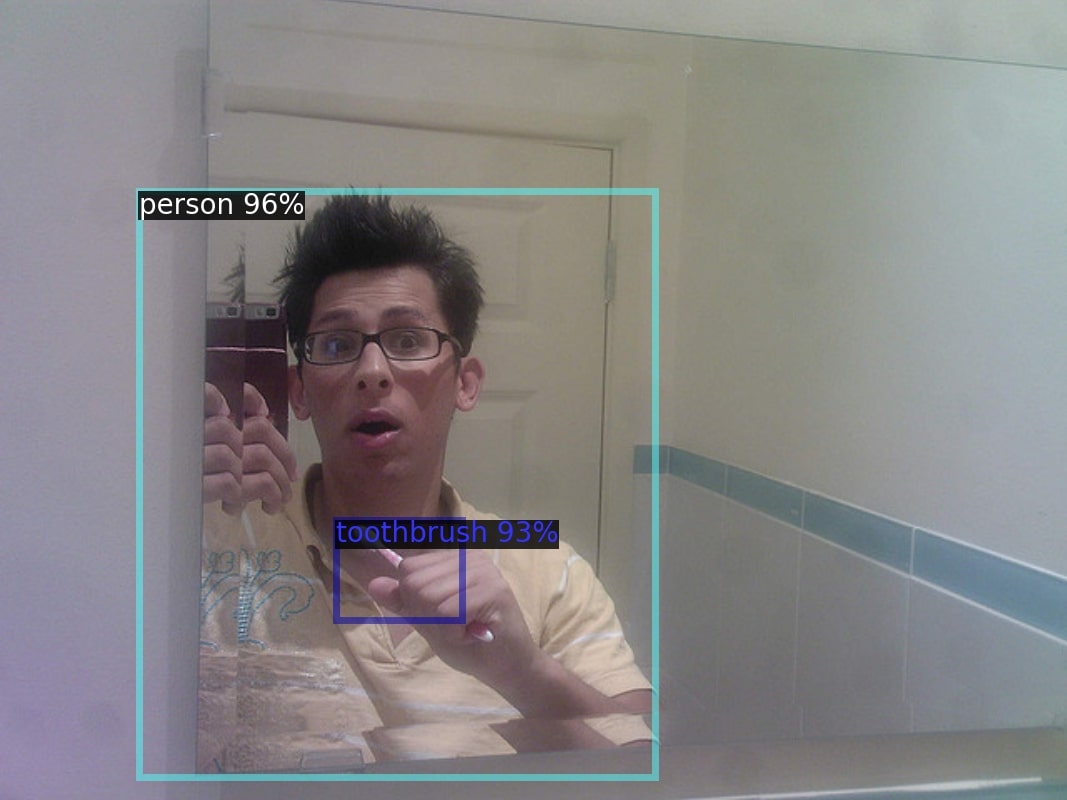} &
\includegraphics[width=0.24\columnwidth]{./baseline/1051.jpg}&
\includegraphics[width=0.24\columnwidth]{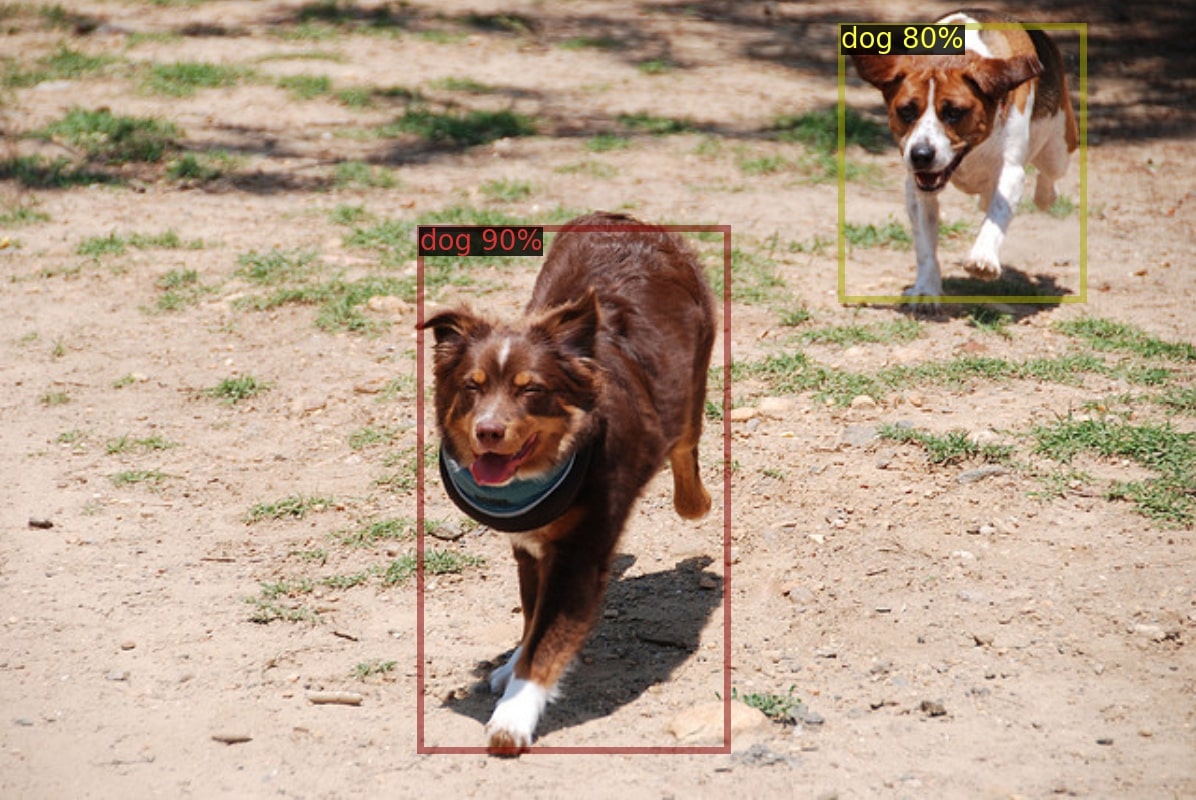} \\

\includegraphics[width=0.24\columnwidth]{./baseline/1062.jpg}&
\includegraphics[width=0.24\columnwidth]{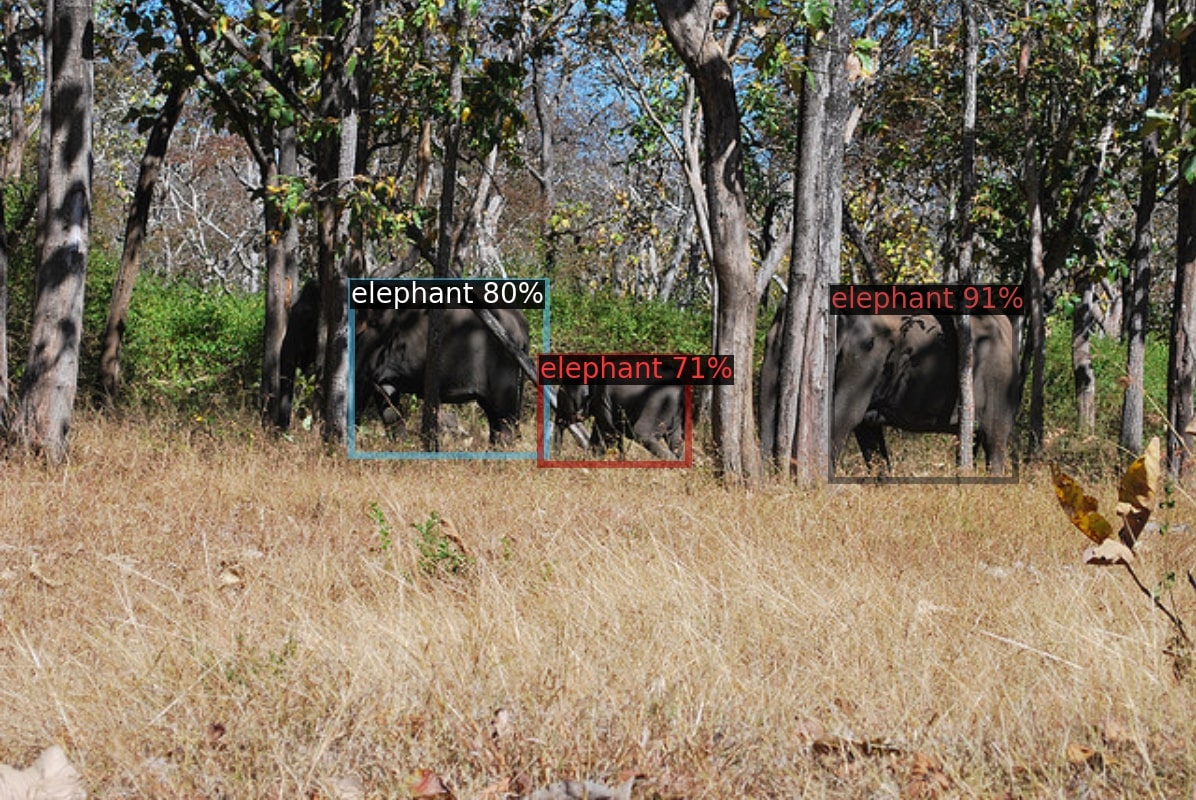} &
\includegraphics[width=0.24\columnwidth]{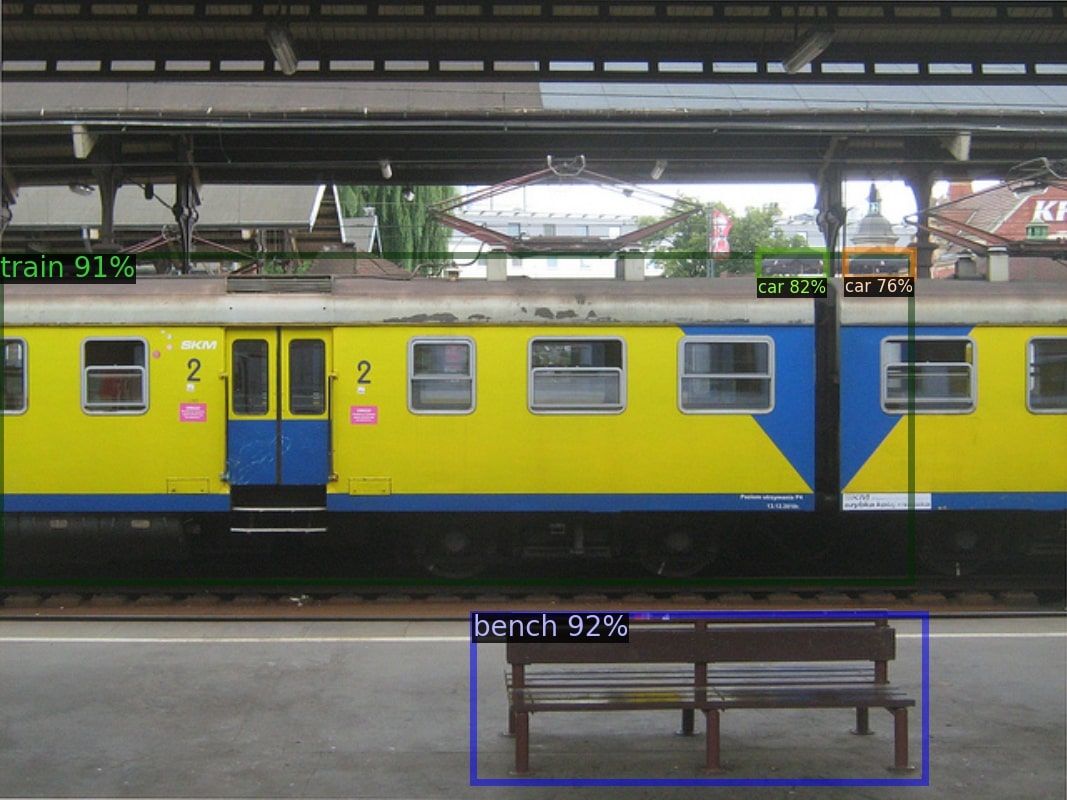}&
\includegraphics[width=0.24\columnwidth]{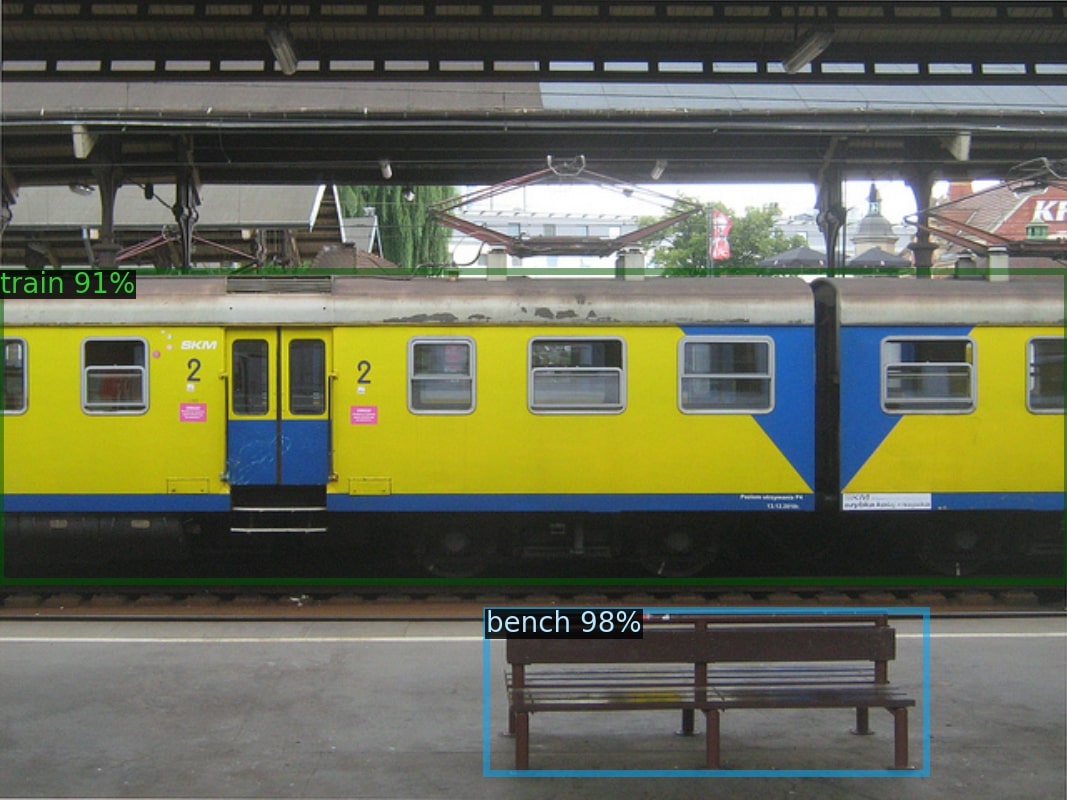} \\

\includegraphics[width=0.24\columnwidth]{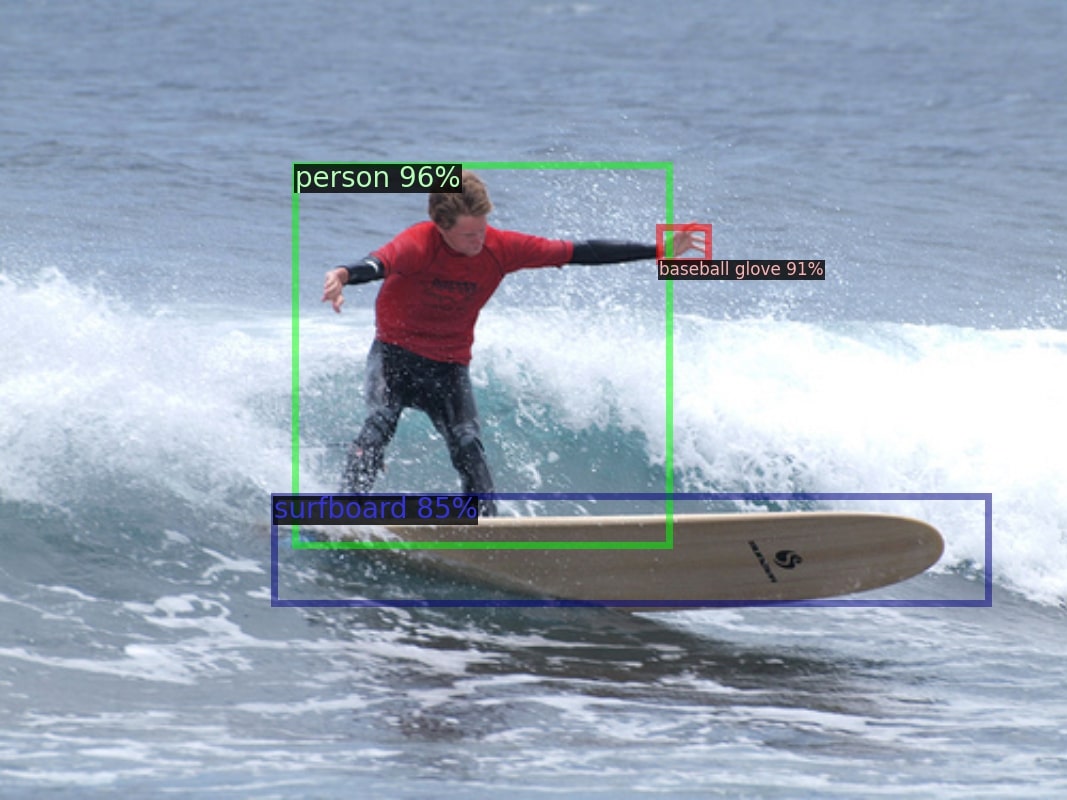}&
\includegraphics[width=0.24\columnwidth]{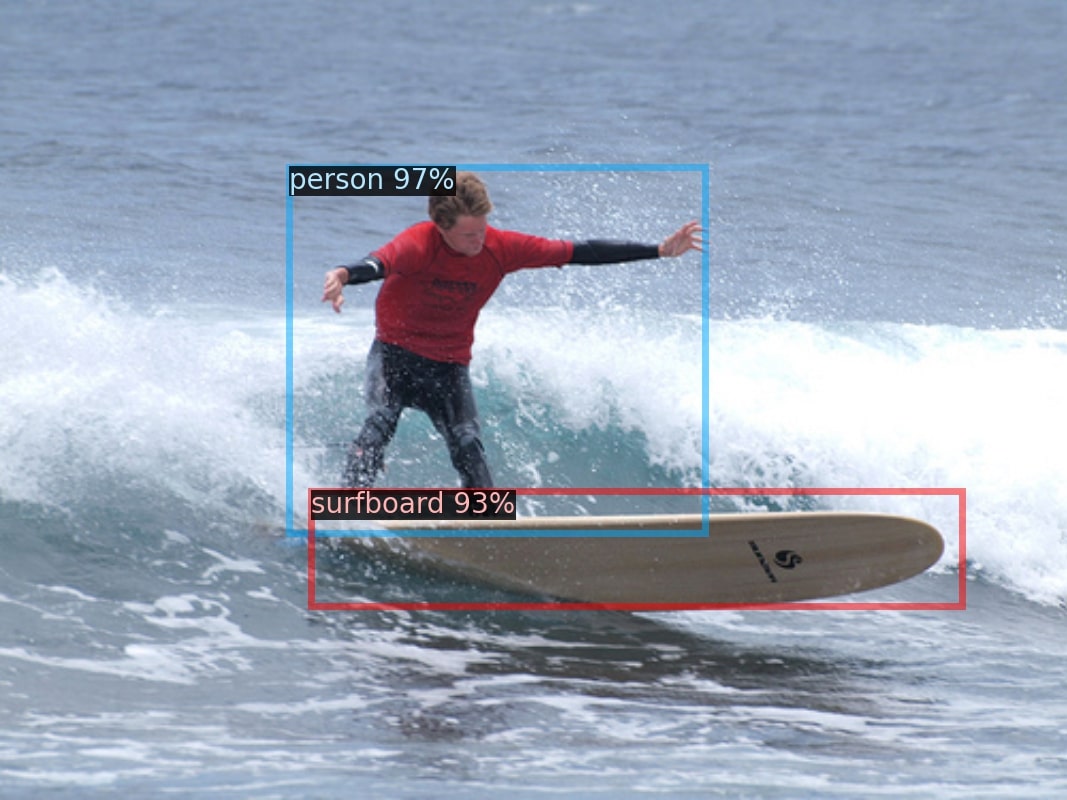} &
\includegraphics[width=0.24\columnwidth]{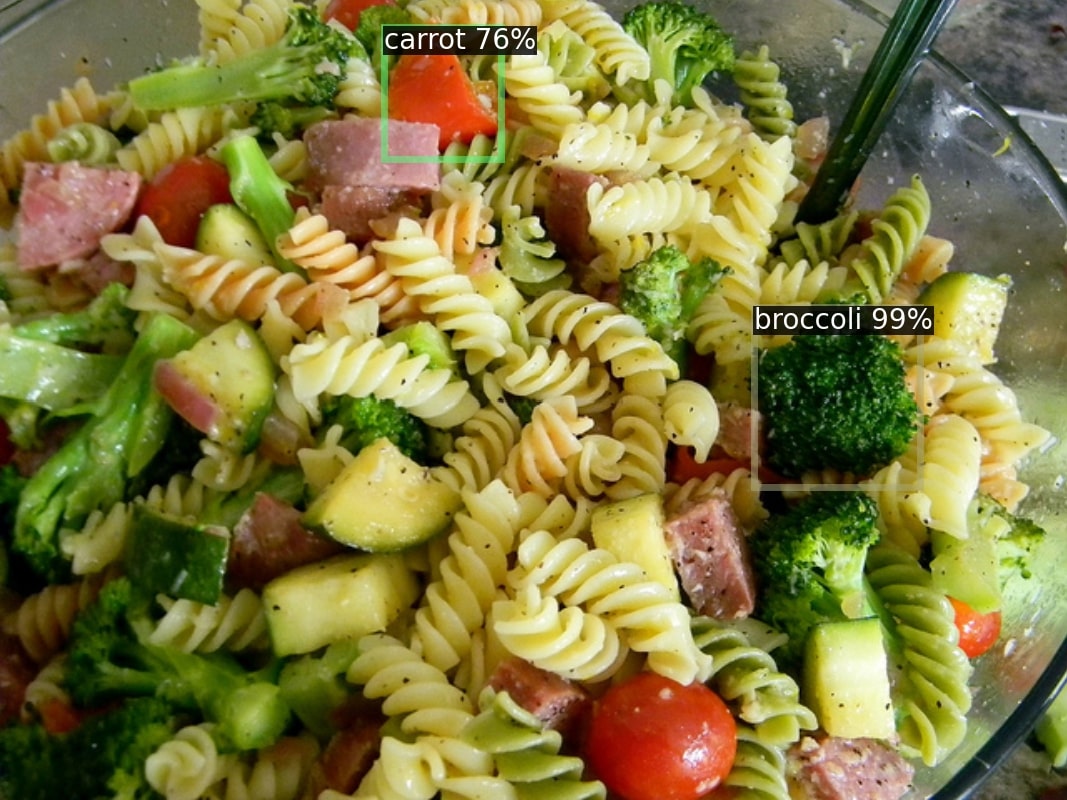}&
\includegraphics[width=0.24\columnwidth]{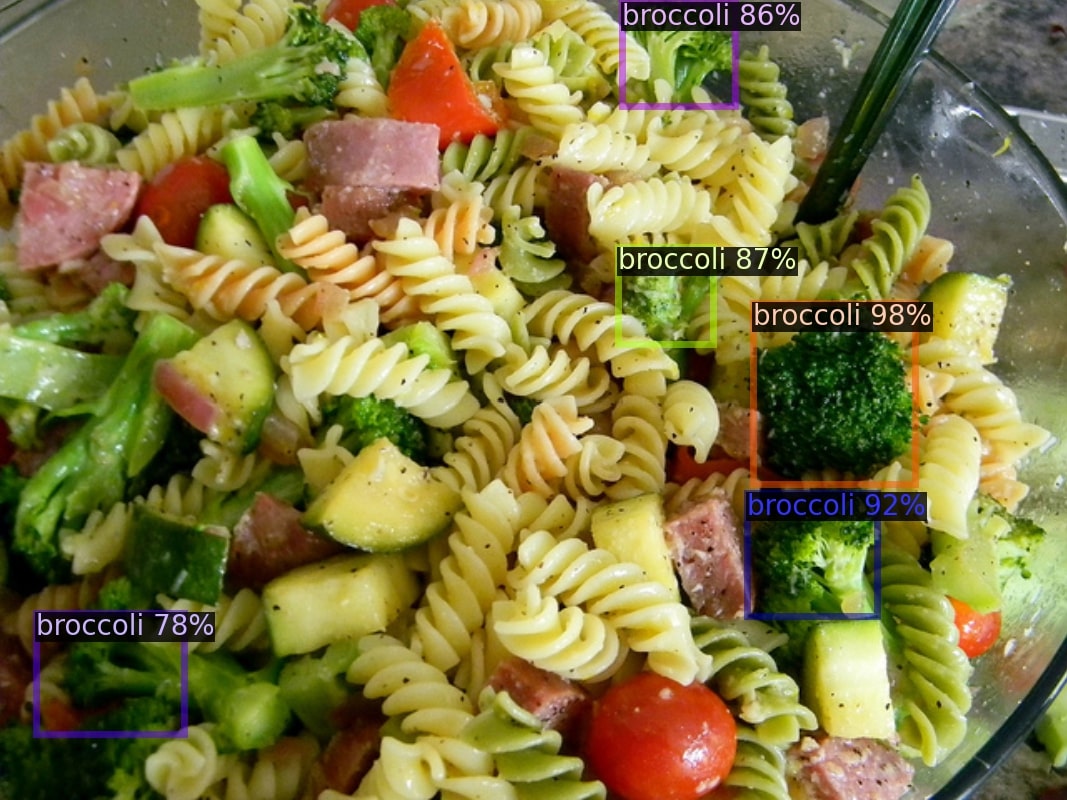} \\

\includegraphics[width=0.24\columnwidth]{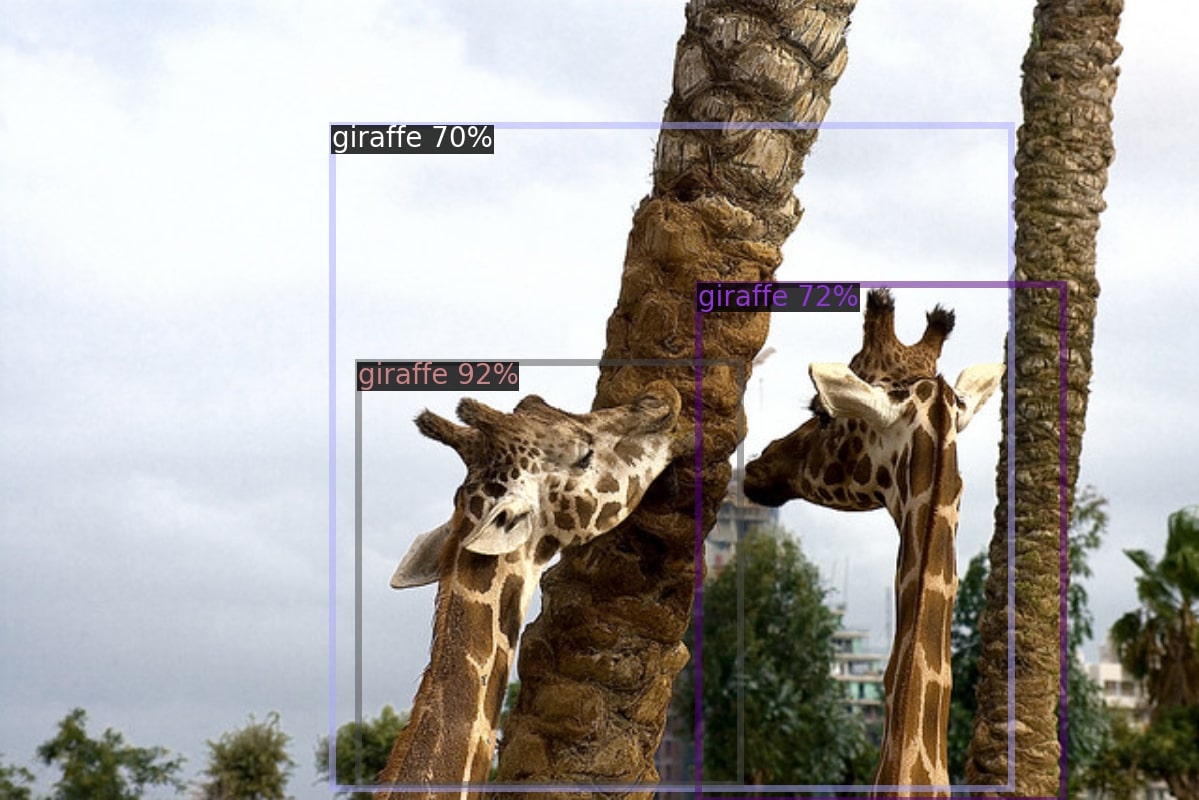}&
\includegraphics[width=0.24\columnwidth]{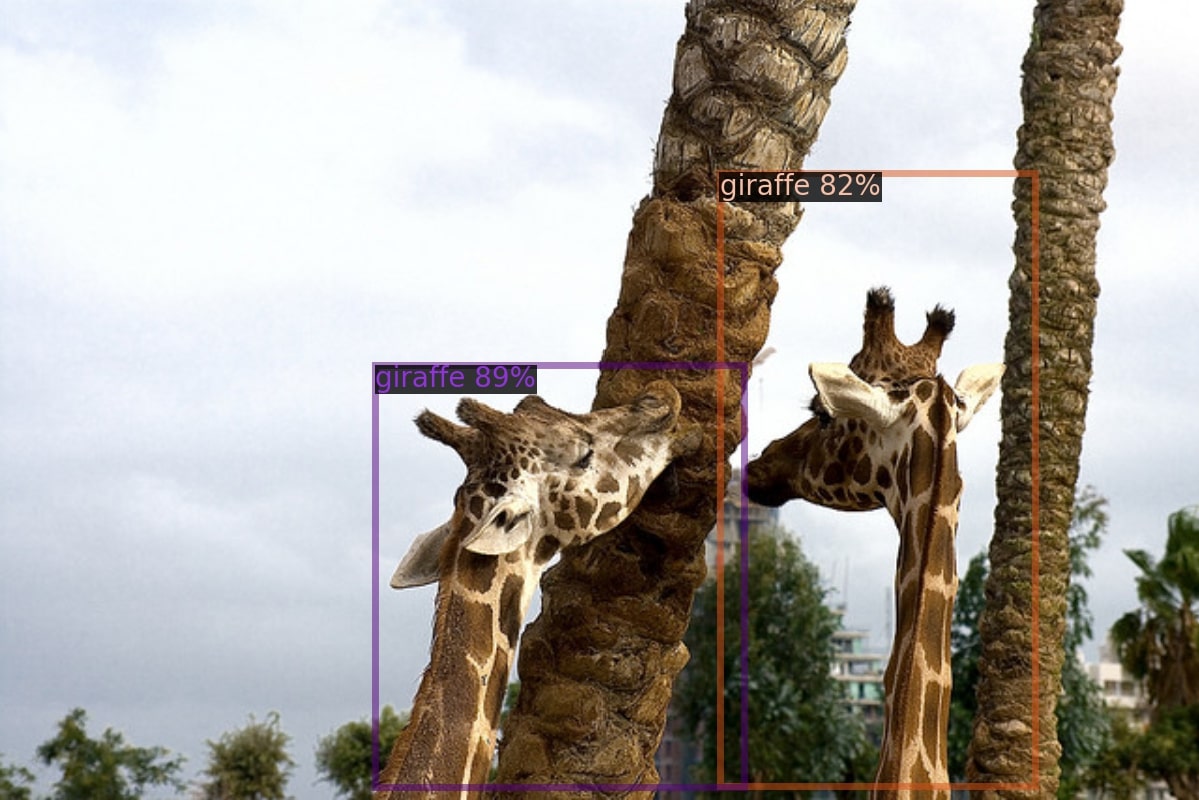} &
\includegraphics[width=0.24\columnwidth]{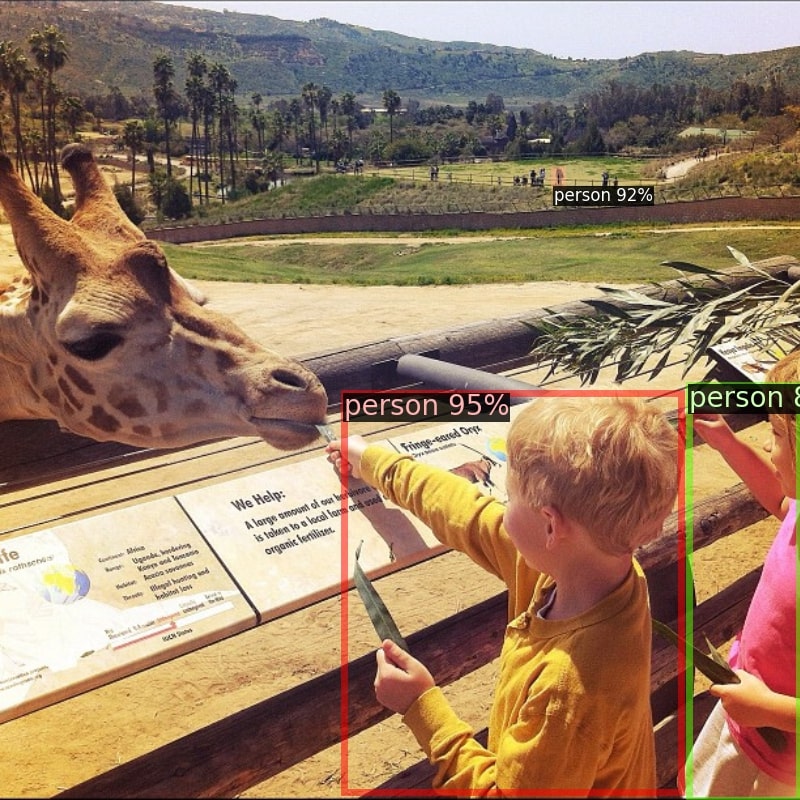}&
\includegraphics[width=0.24\columnwidth]{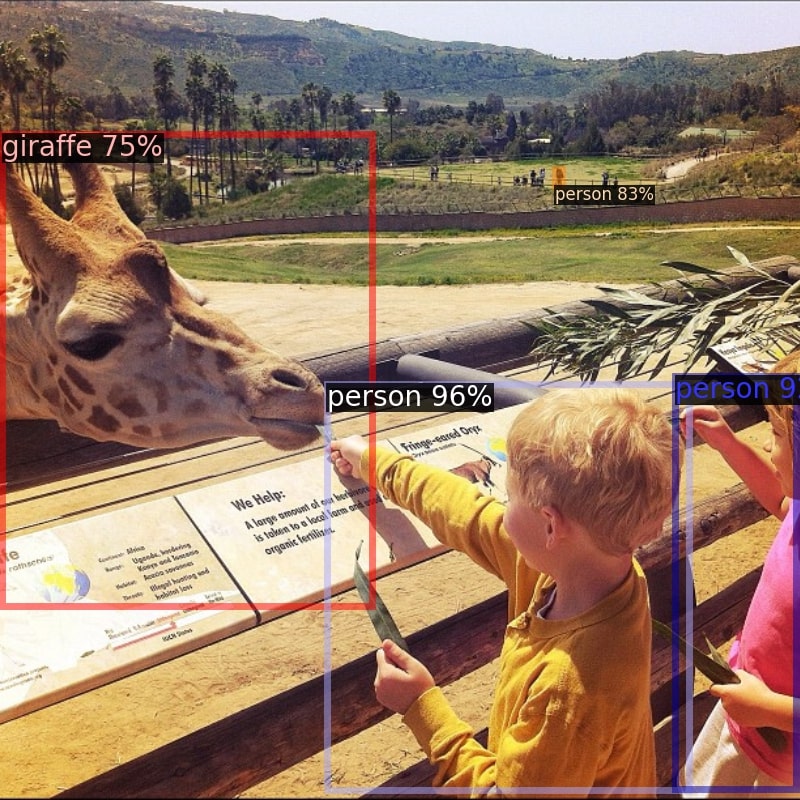} \\

\includegraphics[width=0.24\columnwidth]{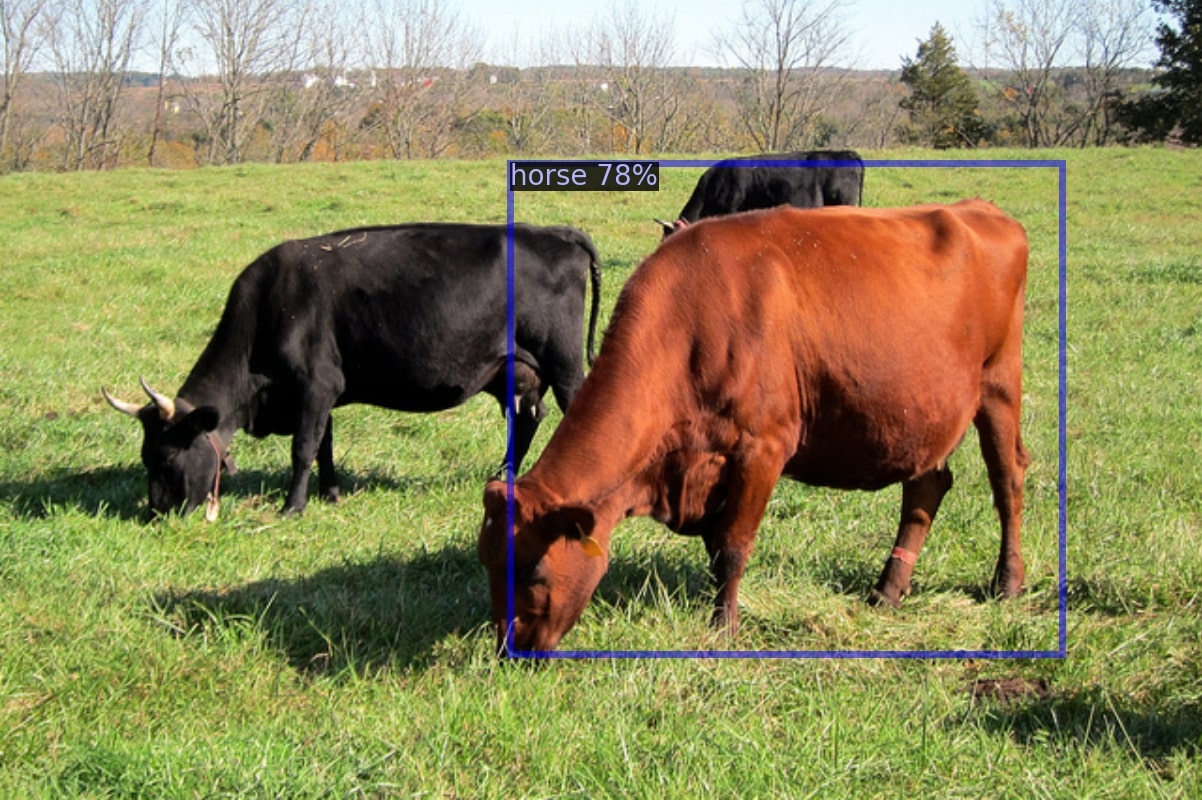}&
\includegraphics[width=0.24\columnwidth]{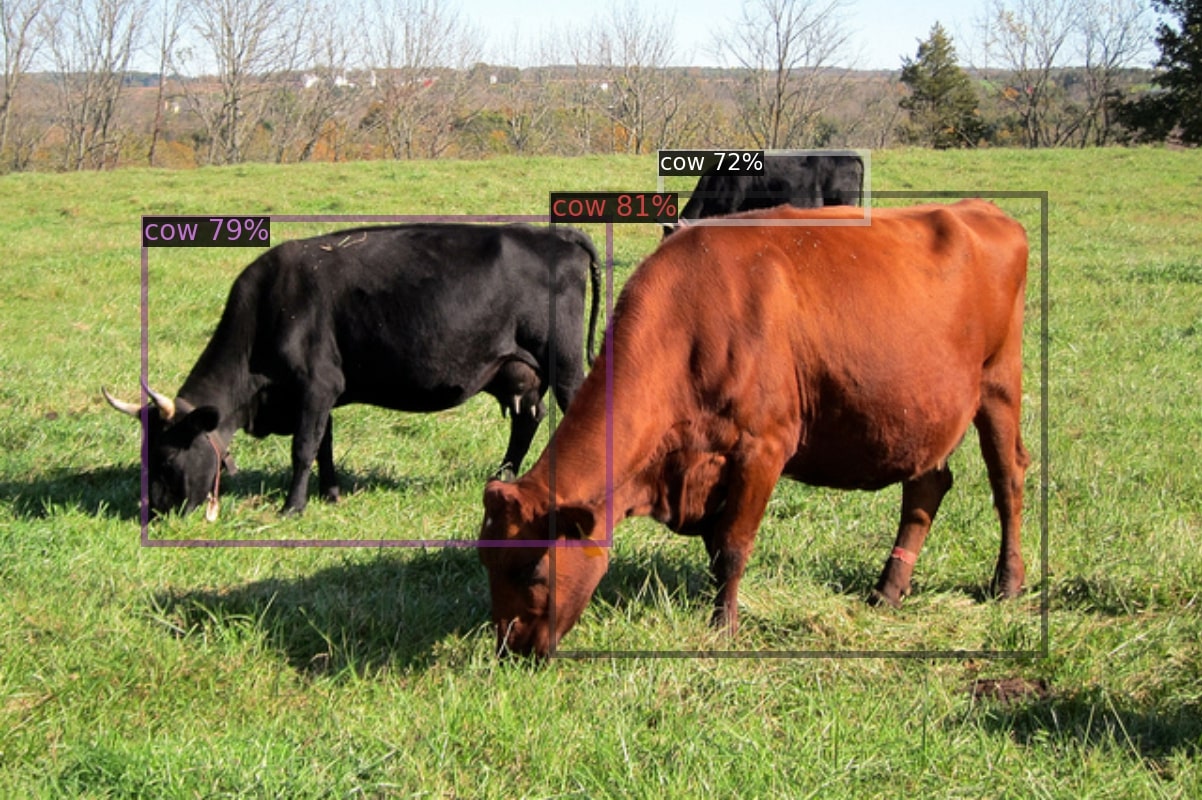} &
\includegraphics[width=0.24\columnwidth]{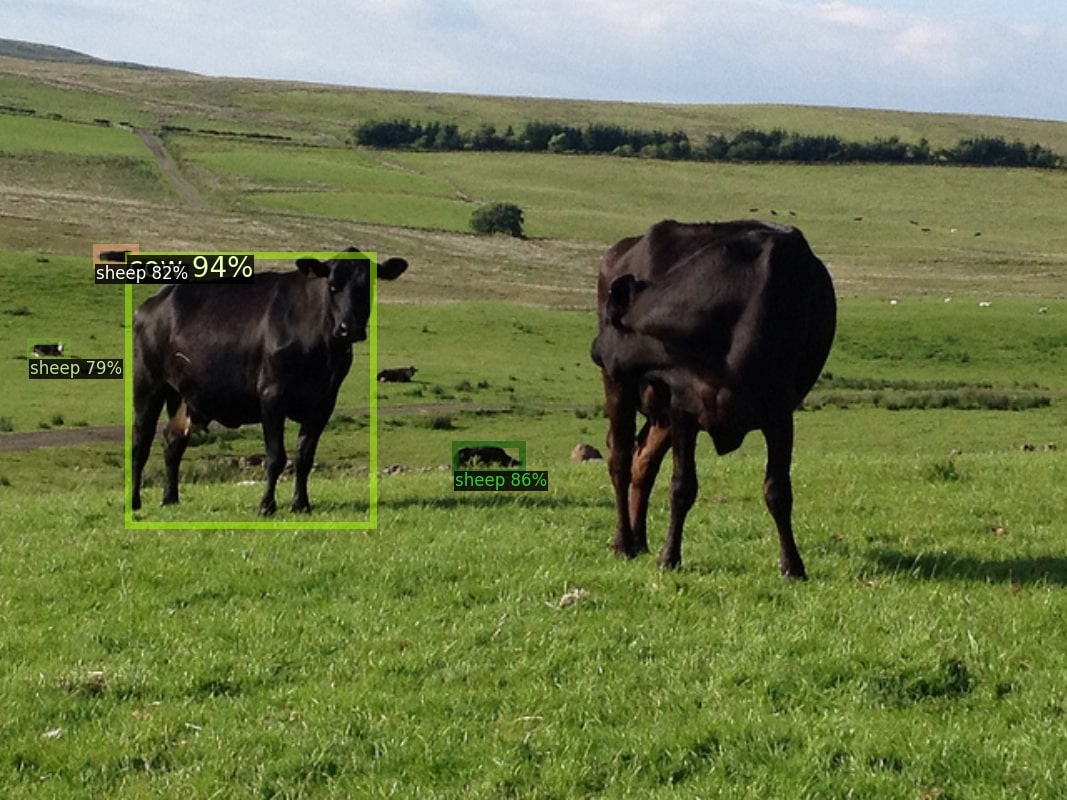}&
\includegraphics[width=0.24\columnwidth]{./ours/2127.jpg} \\

\includegraphics[width=0.24\columnwidth]{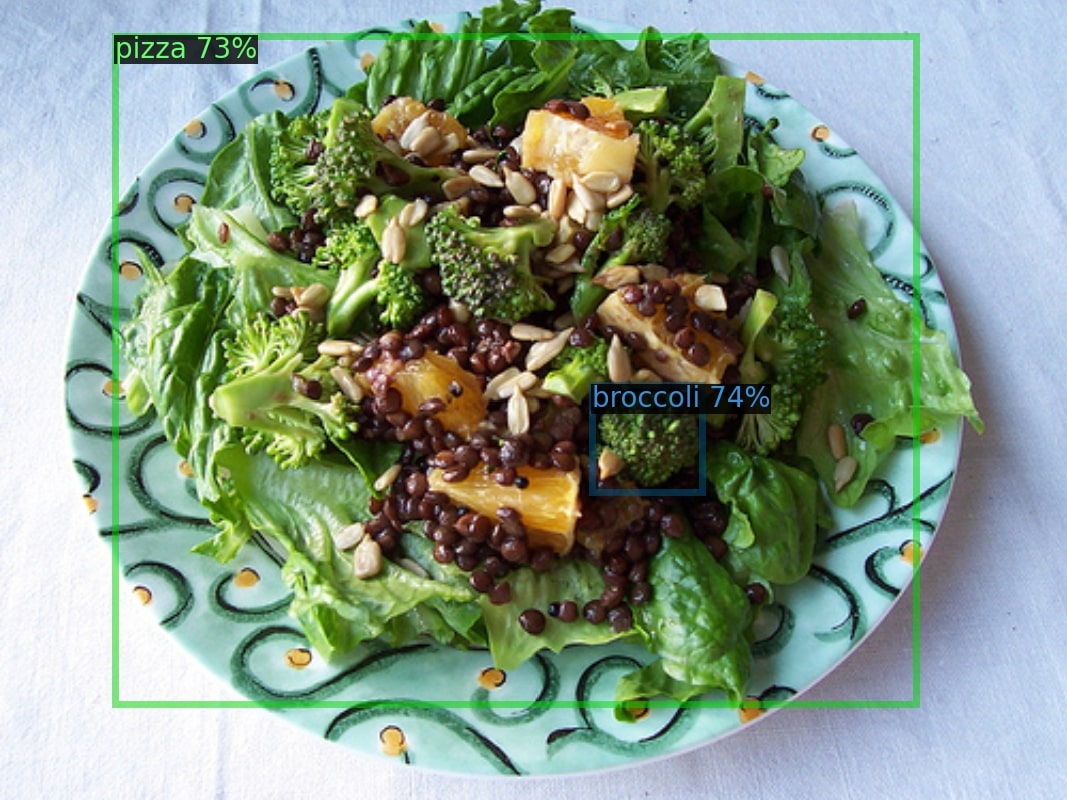}&
\includegraphics[width=0.24\columnwidth]{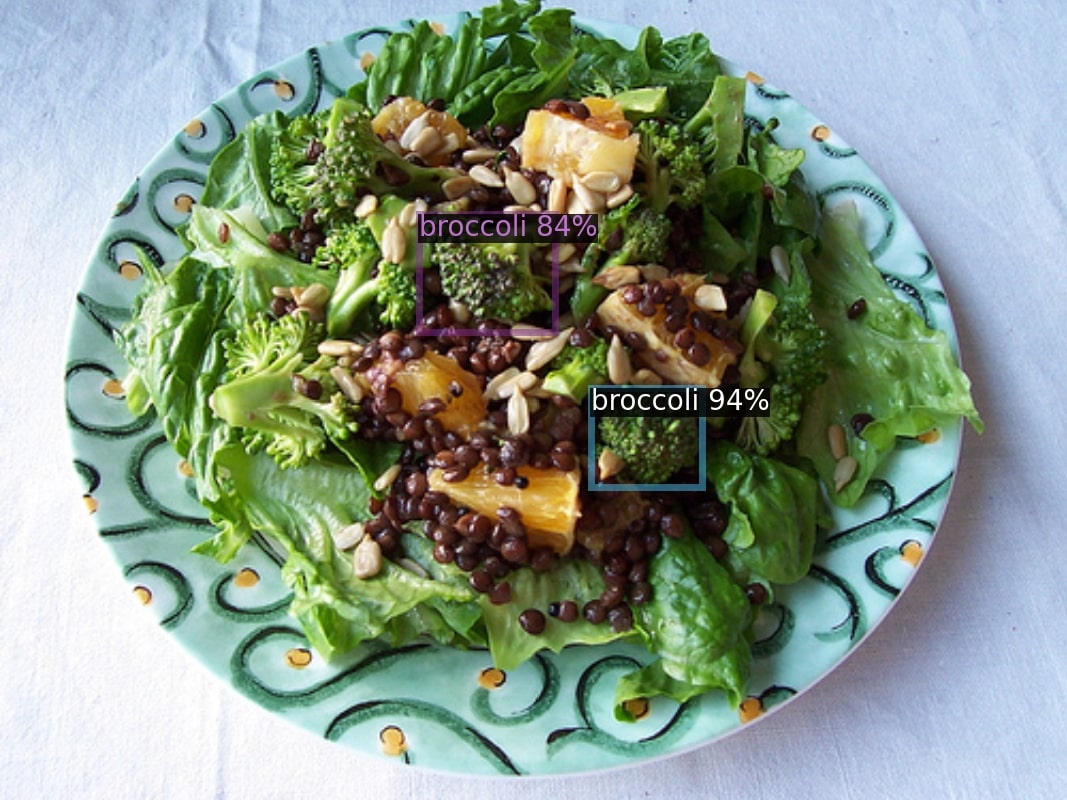} &
\includegraphics[width=0.24\columnwidth]{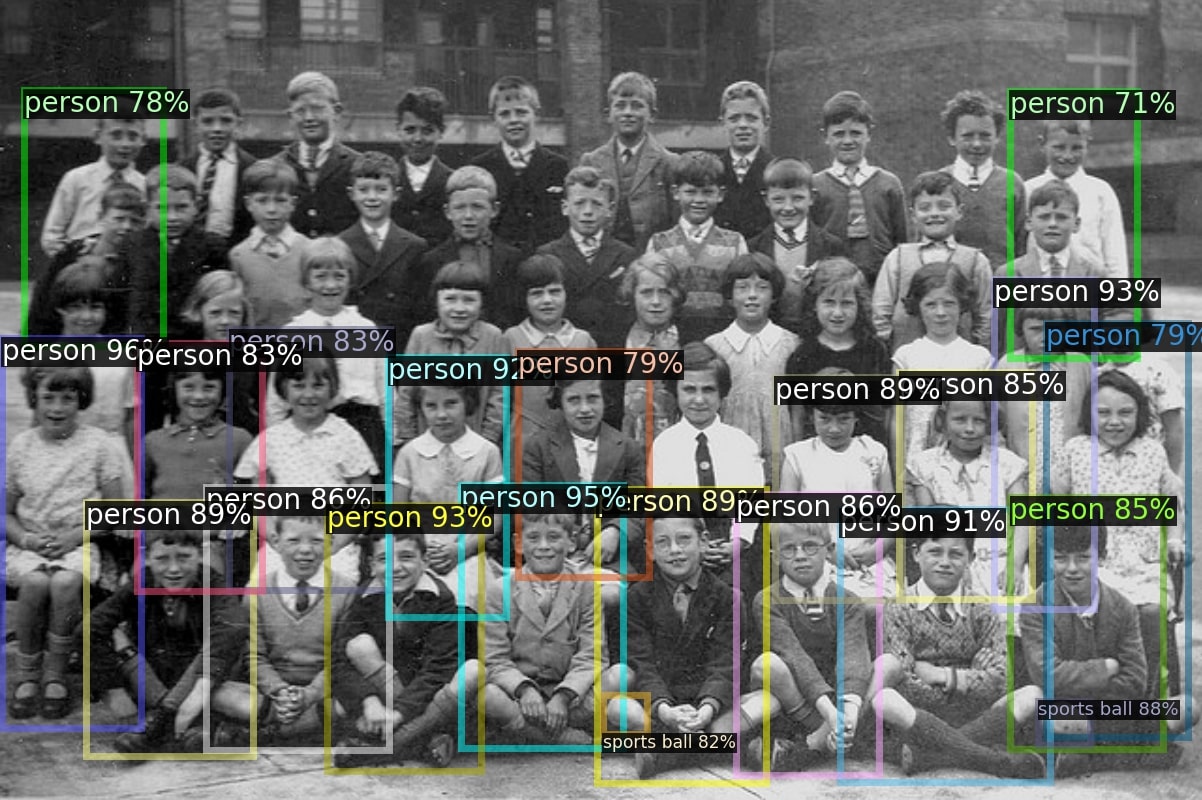}&
\includegraphics[width=0.24\columnwidth]{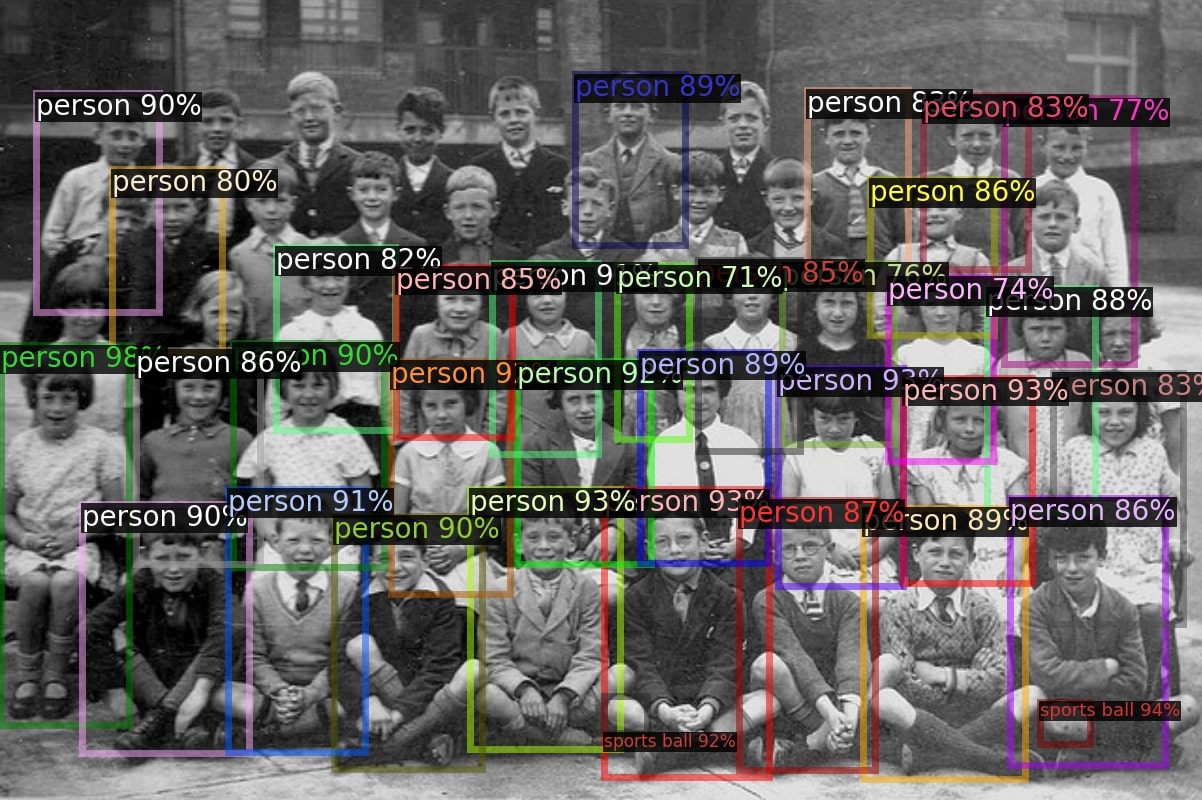} \\

\end{tabular}
\caption{Qualitative results of the baseline~\cite{liu2021unbiased} and our framework~(best viewed with 300\% zoom-in). We set the threshold of classification scores as 0.7 to visualize the results. }
\label{fig:qualitative_3}
\end{figure}

\begin{figure}[h]
\begin{tabular}{c@{}c@{}c@{}c}
Baseline & Ours & Baseline & Ours \\
\includegraphics[width=0.24\columnwidth]{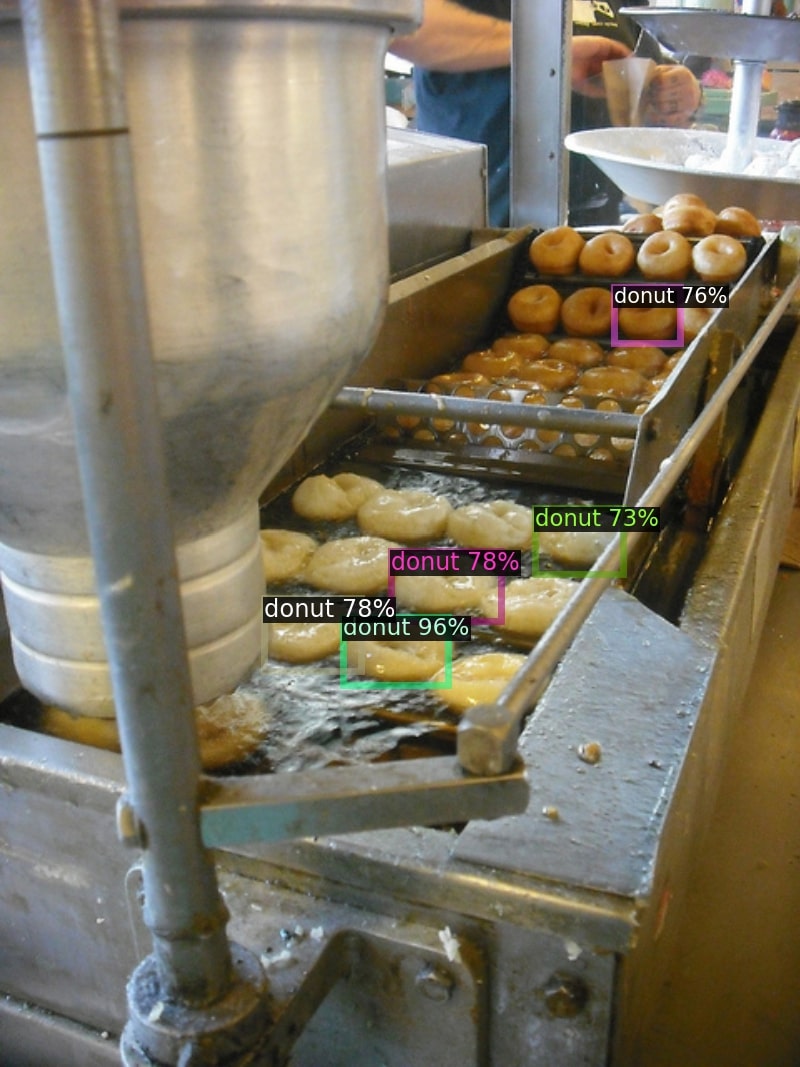}&
\includegraphics[width=0.24\columnwidth]{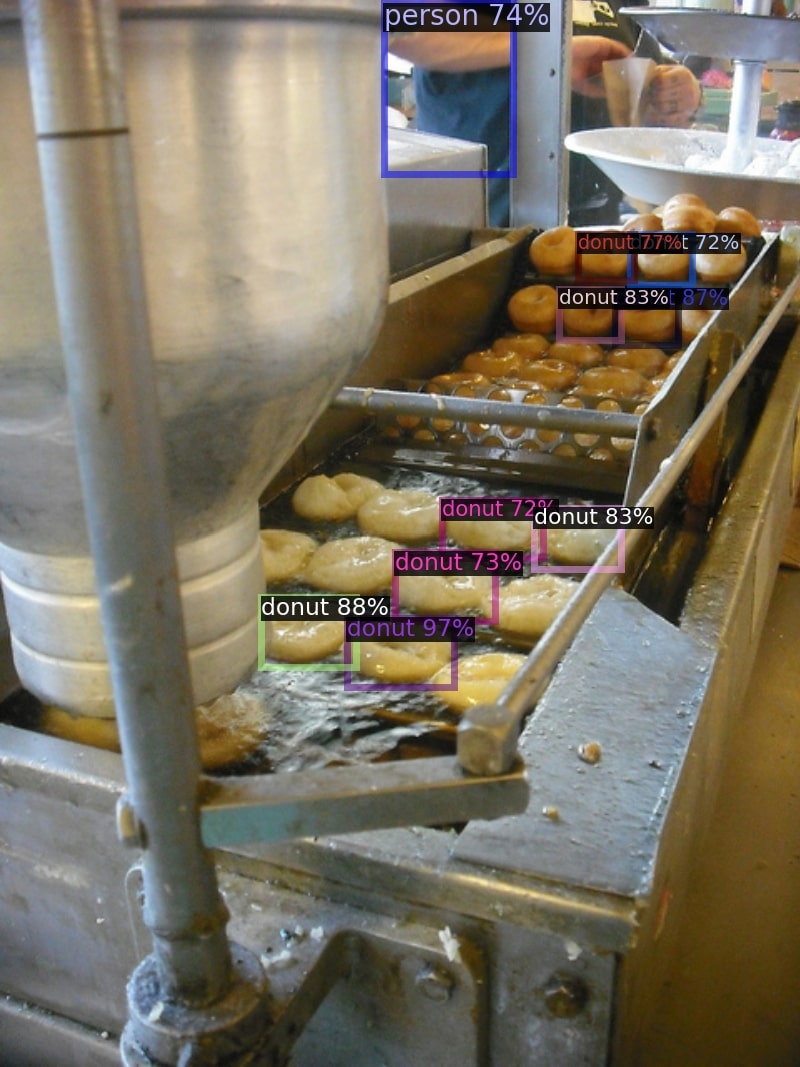} &
\includegraphics[width=0.24\columnwidth]{./baseline/2330.jpg}&
\includegraphics[width=0.24\columnwidth]{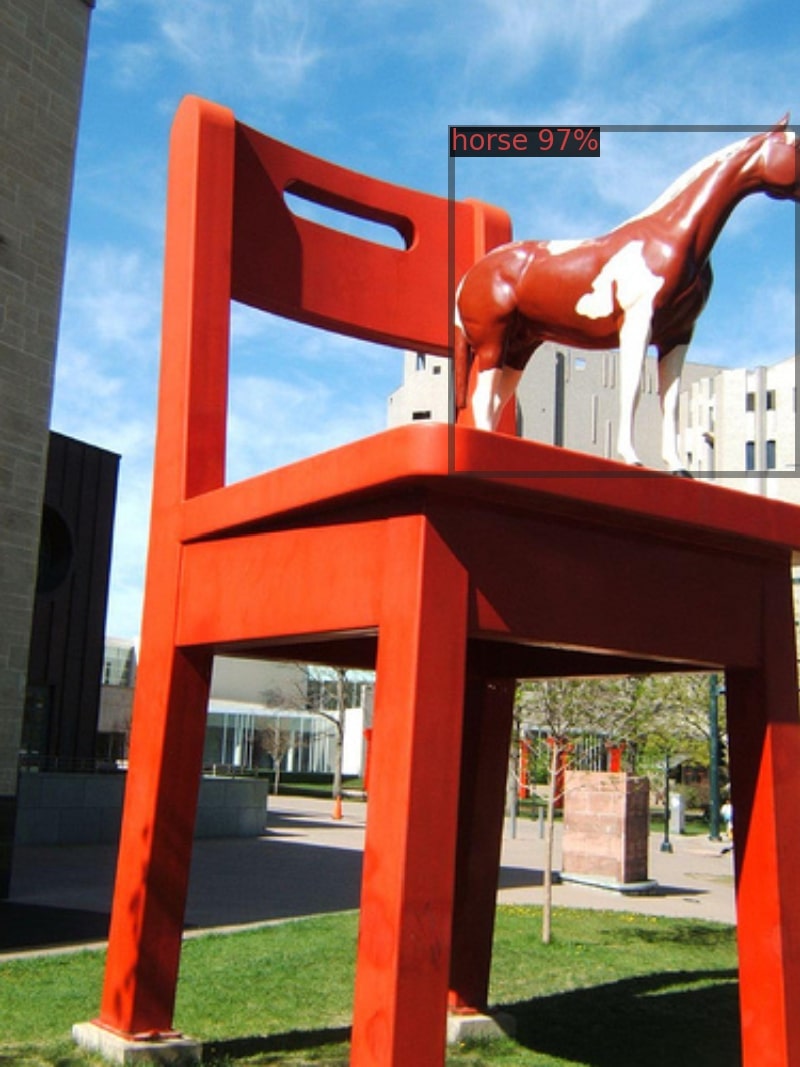} \\

\includegraphics[width=0.24\columnwidth]{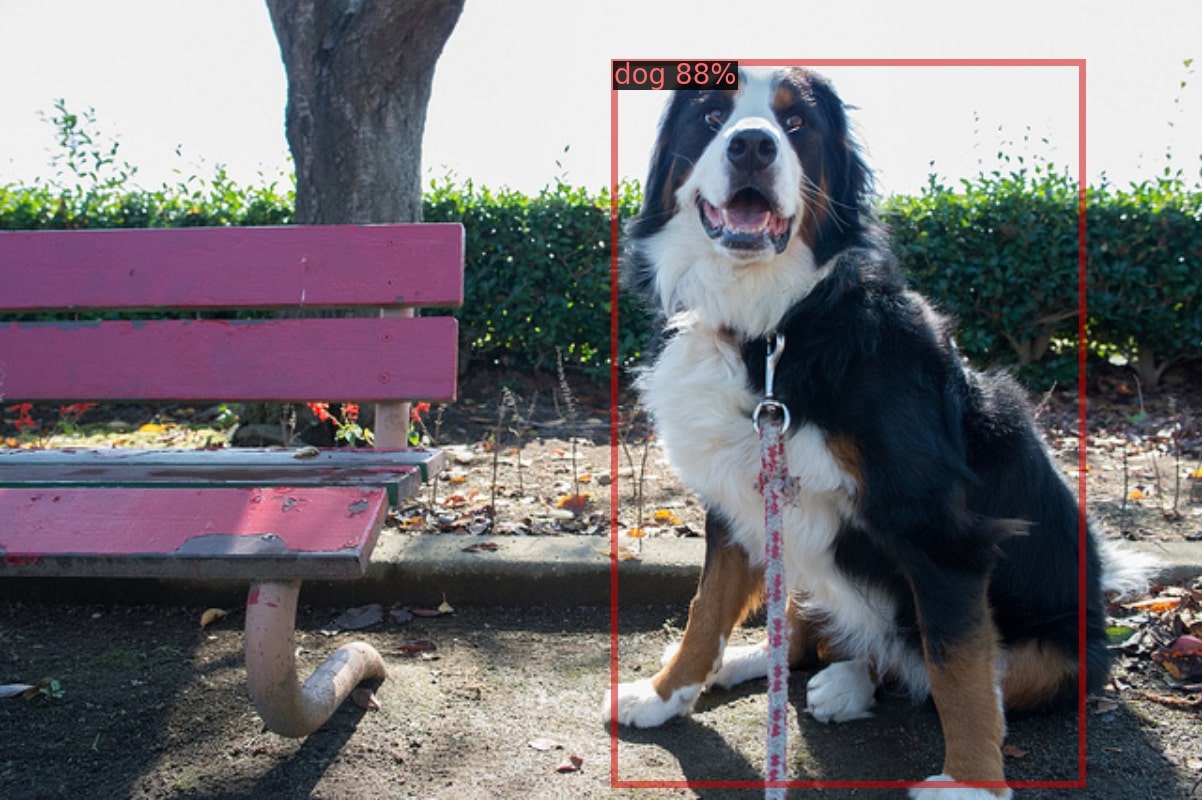}&
\includegraphics[width=0.24\columnwidth]{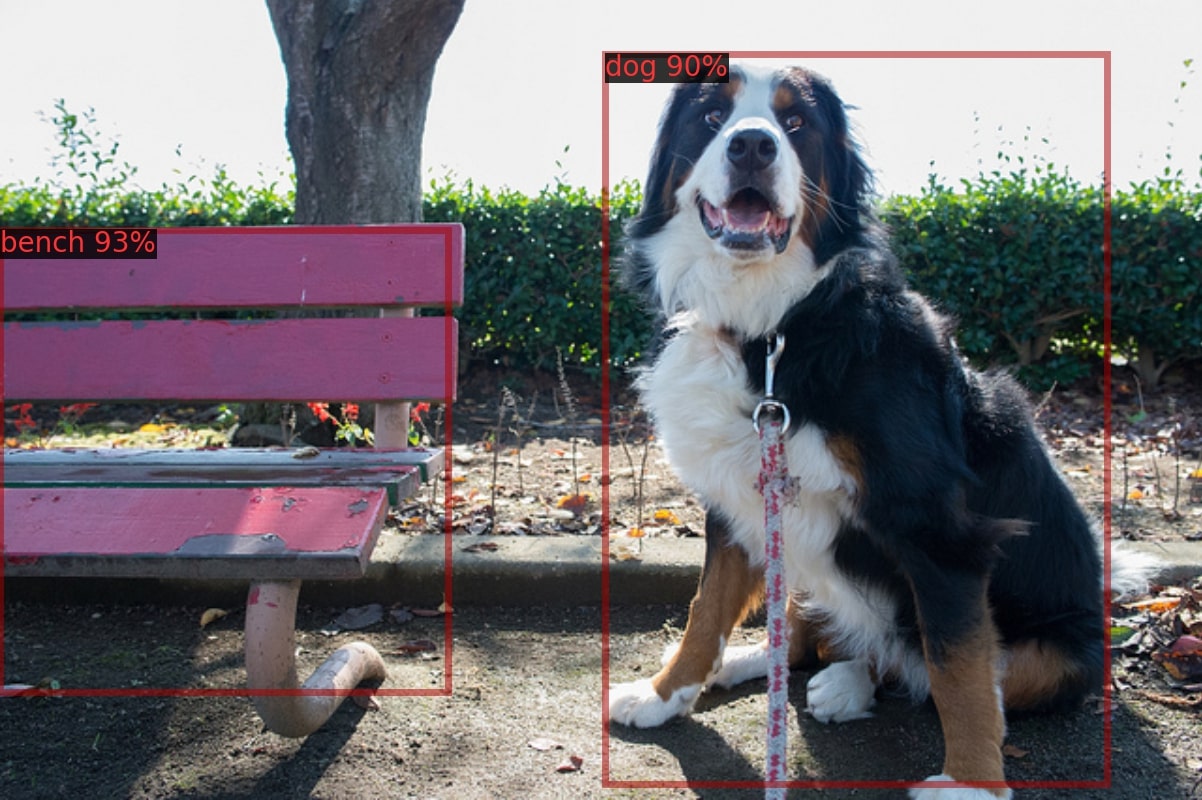} &
\includegraphics[width=0.24\columnwidth]{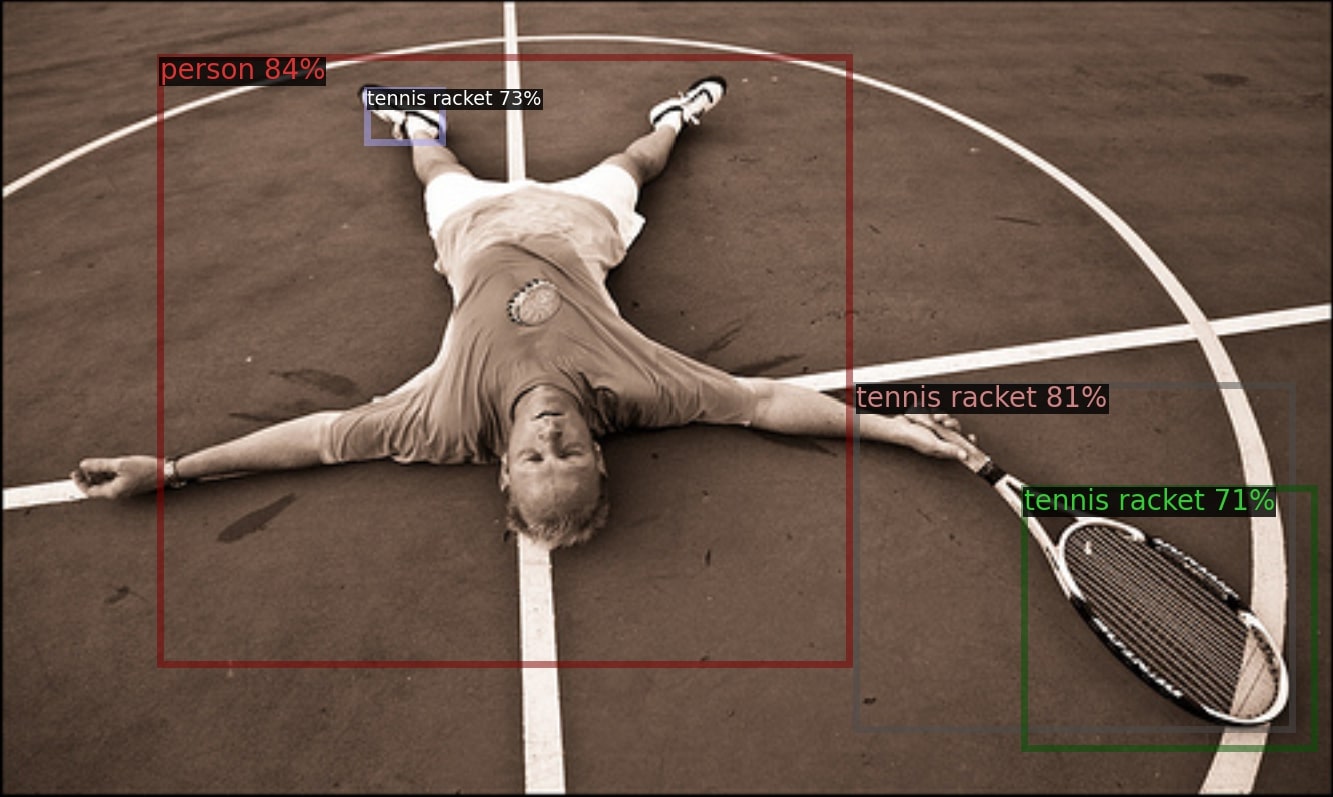}&
\includegraphics[width=0.24\columnwidth]{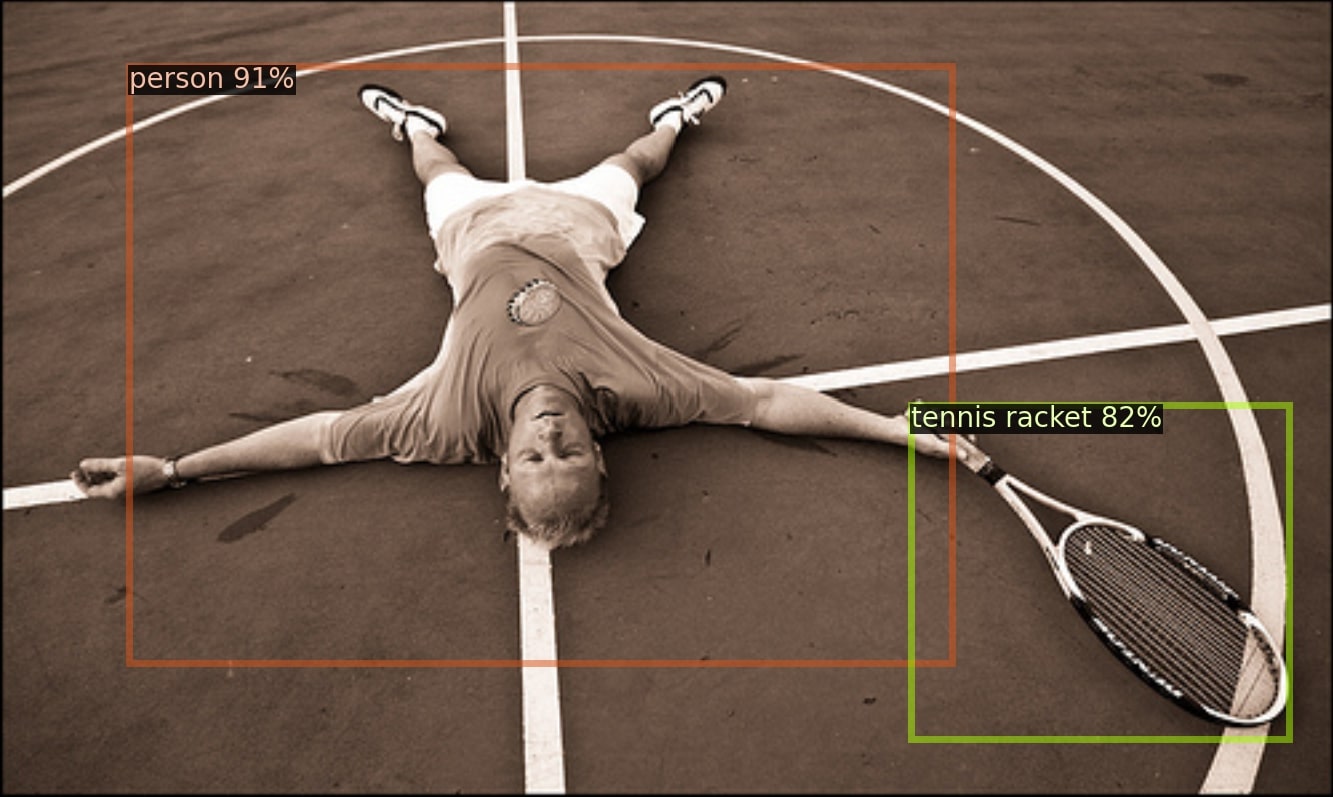} \\

\includegraphics[width=0.24\columnwidth]{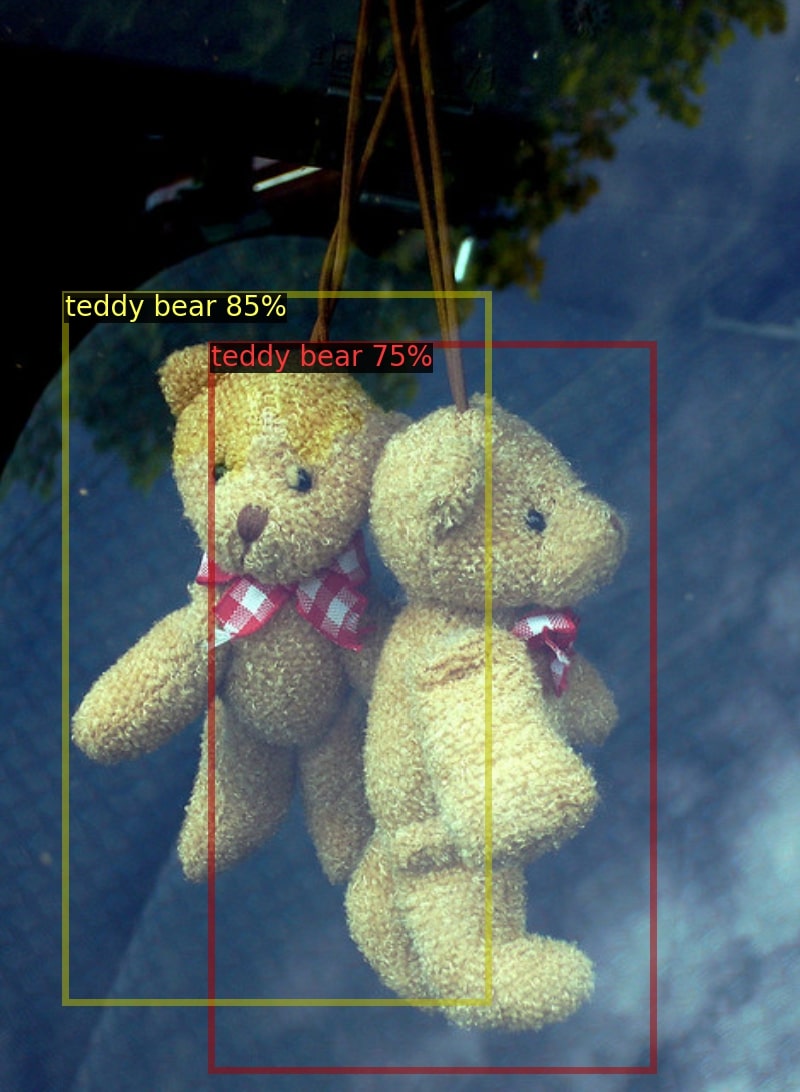}&
\includegraphics[width=0.24\columnwidth]{./ours/2390.jpg} &
\includegraphics[width=0.24\columnwidth]{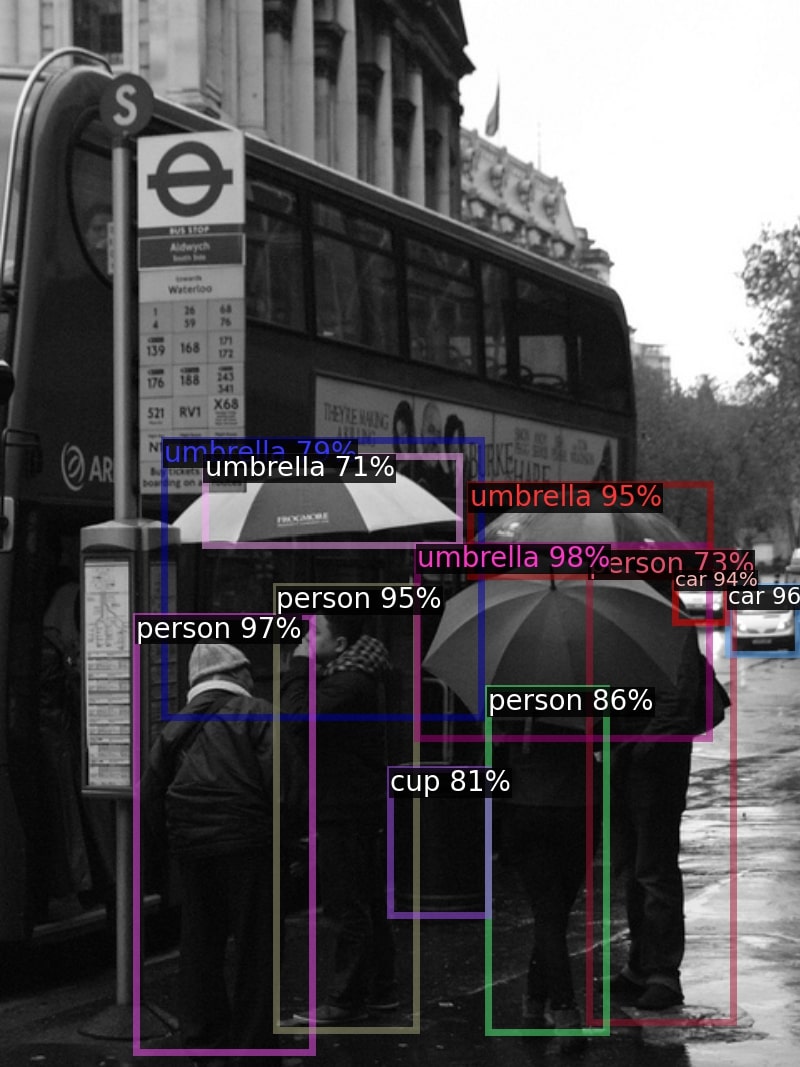}&
\includegraphics[width=0.24\columnwidth]{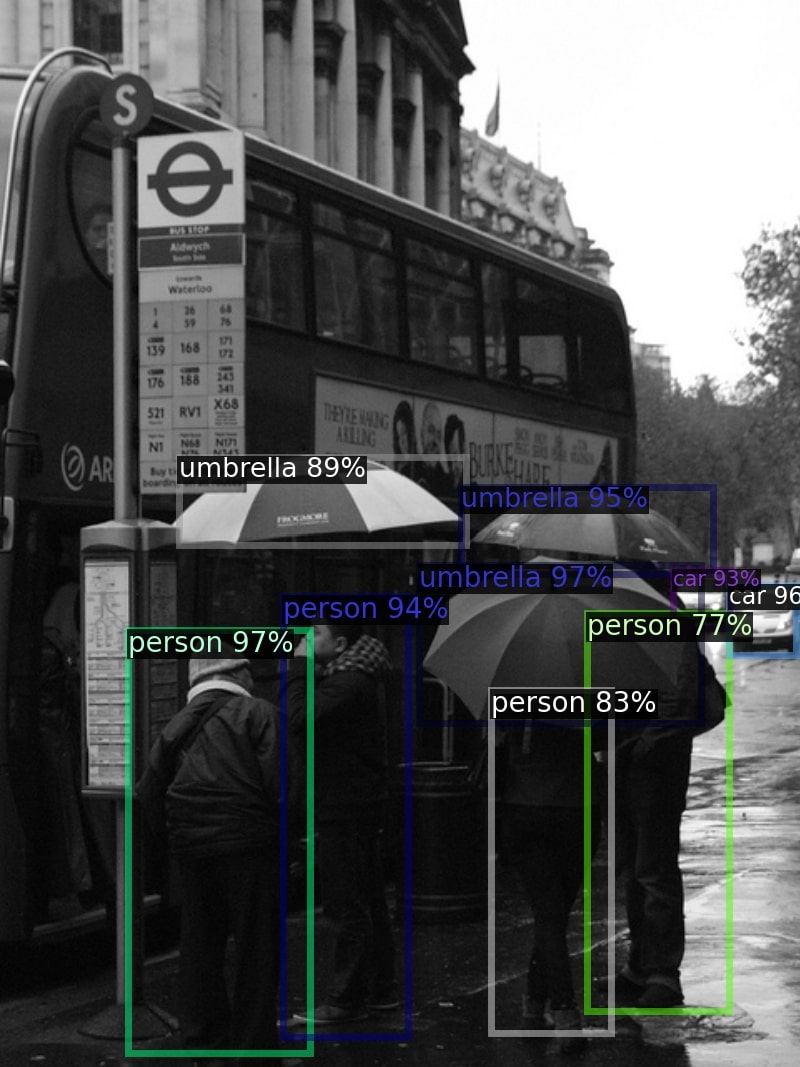} \\

\includegraphics[width=0.24\columnwidth]{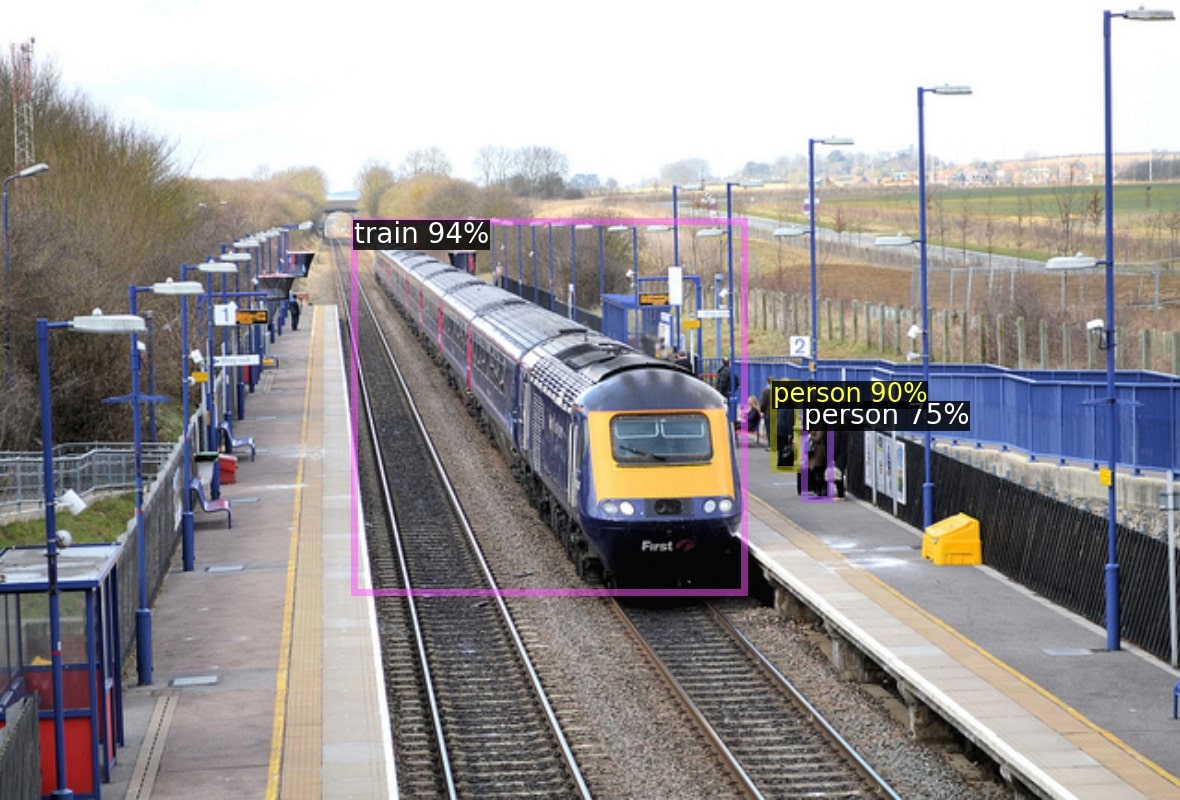}&
\includegraphics[width=0.24\columnwidth]{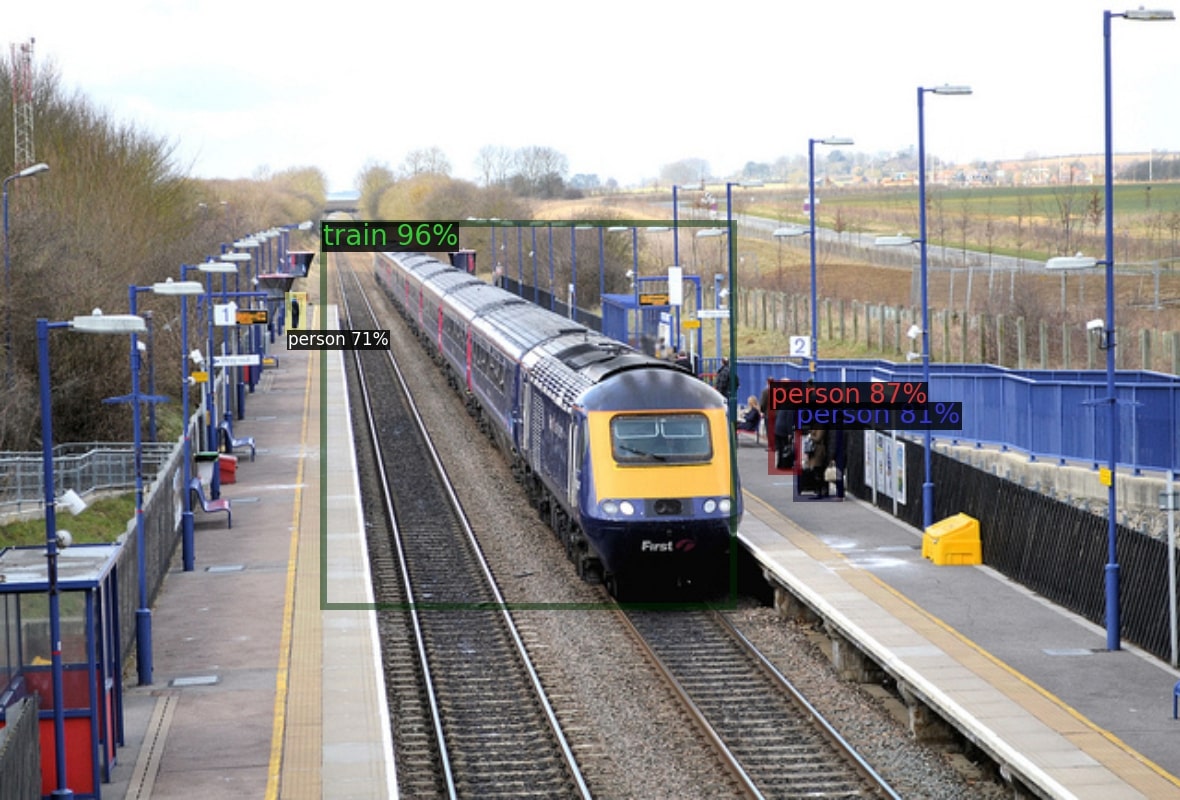} &
\includegraphics[width=0.24\columnwidth]{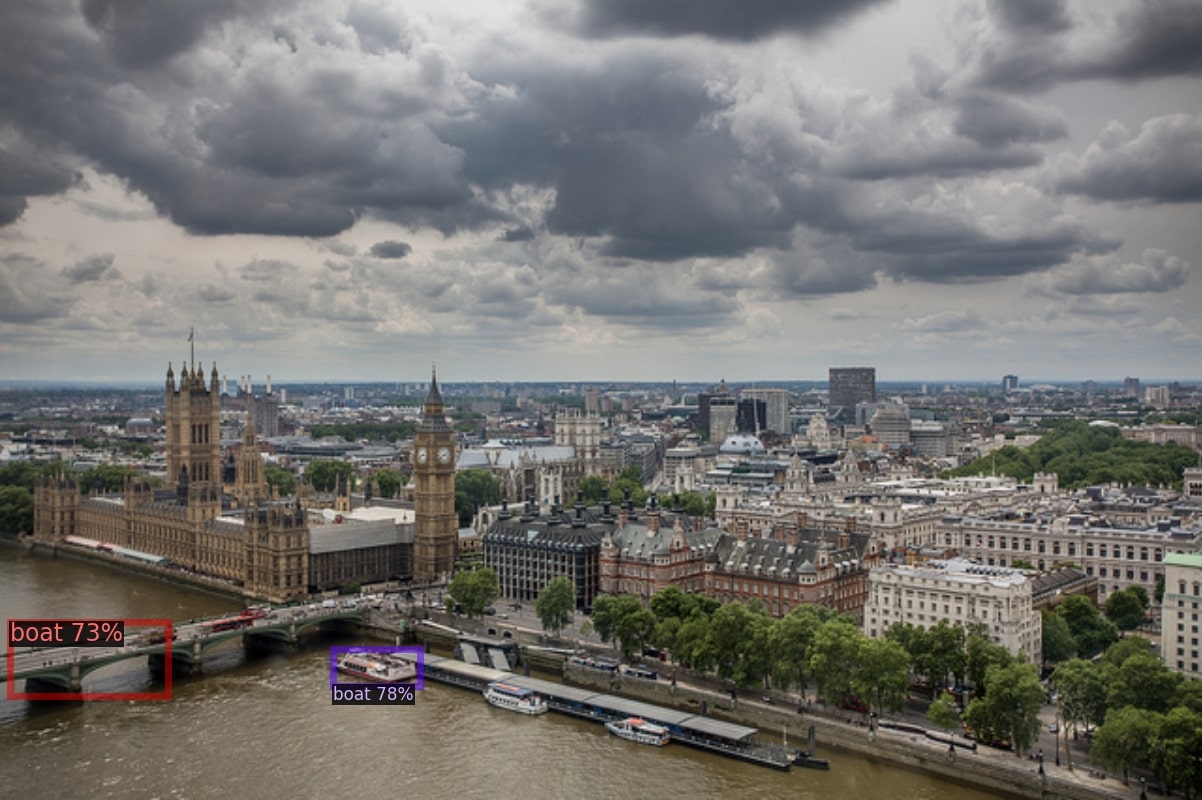}&
\includegraphics[width=0.24\columnwidth]{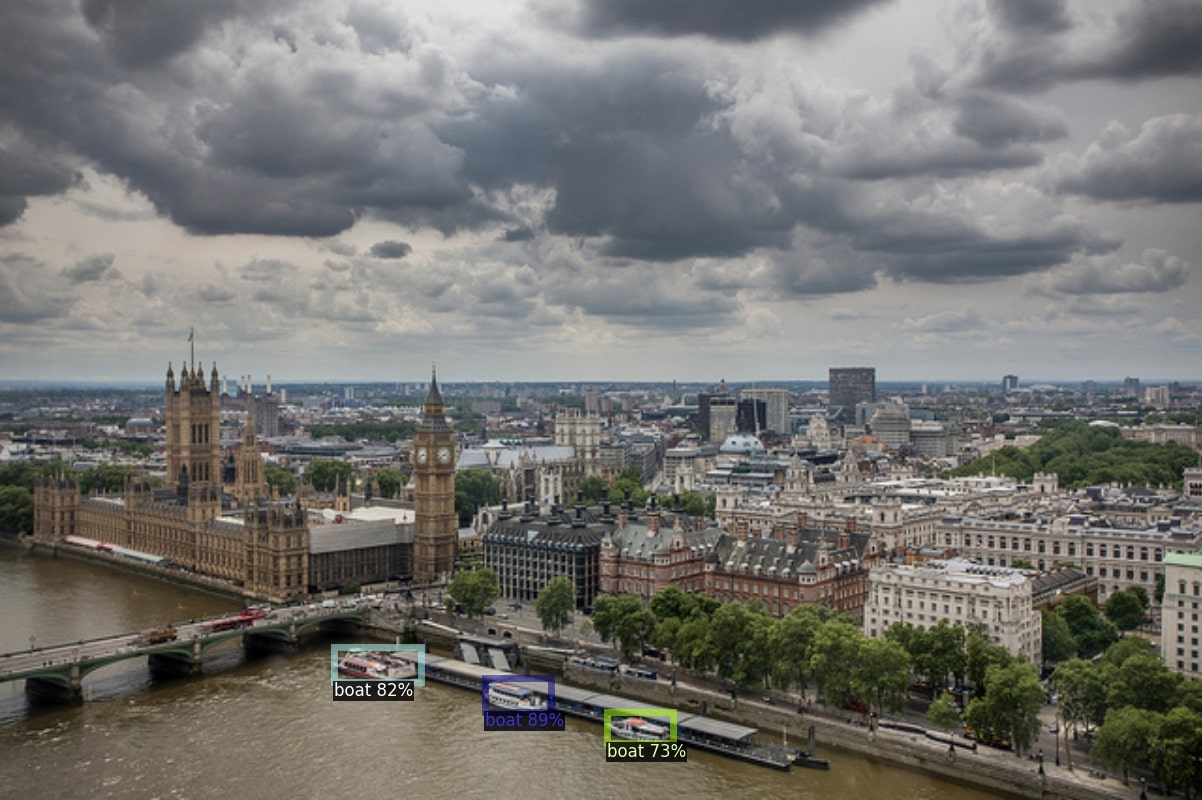} \\
\end{tabular}
\caption{Qualitative results of the baseline~\cite{liu2021unbiased} and our framework~(best viewed with 300\% zoom-in). We set the threshold of classification scores as 0.7 to visualize the results. }
\label{fig:qualitative_4}
\end{figure}

We show the qualitative results of our framework and the baseline~\cite{liu2021unbiased} in Fig.~\ref{fig:qualitative}, Fig.~\ref{fig:qualitative_2}, Fig.~\ref{fig:qualitative_3}, and Fig.~\ref{fig:qualitative_4}. We can observe some advantages of ours. 1) By introducing the OCL, our model can detect more objects and the classification scores are higher than the baseline~\cite{liu2021unbiased}. 2) By introducing the RUPL, our model can predict more accurate bounding boxes.

\section{Implementation Details}

\textbf{Framework details.}
We implement our framework based on Detectron2~\cite{wu2019detectron2} and UBT~\cite{liu2021unbiased}. Following previous works~\cite{STAC, liu2021unbiased}, we use Faster-RCNN~\cite{ren2015faster} with ResNet-50 backbone and FPN~\cite{fpn} as our detector. For OCL, we adopt an asymmetric architecture~\cite{mocov3, byol} in which the teacher network has a projection module parallel to the classification branch, and the student network has an extra prediction module after a projection module. We show the architecture of projection and prediction modules in Tab.~\ref{tab:arch_module}. For RUPL, we add class-aware uncertainty branch parallel to regression branch following He~\etal{}~\cite{box_regression_uncertainty}. 

\begin{table}[h]
    \centering
    \scriptsize
    \begin{tabular}{c|c|c}
    \toprule
    \multicolumn{3}{c}{\textbf{Projection Module}} \\
    \toprule
    Layer & In\_dim & Out\_dim\\
    \midrule
    Linear~(w/o bias) & 1024 & 2048 \\
    Batchnorm1D & 2048 & 2048 \\
    ReLU & - & - \\
    Linear~(w/o bias) & 2048 & 128 \\
    Batchnorm1D~(w/o affine) & 128 & 128 \\
    \toprule
    \multicolumn{3}{c}{\textbf{Prediction Module}} \\
    \toprule
    Layer & In\_dim & Out\_dim \\
    \midrule
    Linear~(w/o bias) & 128 & 2048 \\
    Batchnorm1D & 2048 & 2048 \\
    ReLU & - & - \\
    Linear~(w/o bias) & 2048 & 128 \\
    \bottomrule
    \end{tabular}
    \caption{Architecture for projection and predictions modules.}
    \label{tab:arch_module}
\end{table}

\noindent\textbf{Training details.}
Following the baseline~\cite{liu2021unbiased}, backbone network's weights are initialized by ImageNet-pretrained model, and SGD optimizer is used with constant learning rate. For SGD optimizer, we set a momentum parameter and learning rate as 0.9 and 0.01, respectively. For OCL, we apply random box jittering on predicted bounding boxes~$b^p$, which are predicted by the teacher network taking as input the weakly augmented unlabeled images $\mathcal{A}_0(\bm{x}^u)$, before generating $\bm{z}^t$ and $\bm{z}^s$. We randomly sample values between [-6$\%$, 6$\%$] of the height and width of bounding boxes and then add them to predicted bounding boxes~$b^p$. Box jittering is used to learn generalizable ROI feature representation, similar to random cropping in representation learning~\cite{simclr}. Loss balance parameters $\lambda_{ocl}$ and $\lambda_{unc}$ are set as 0.1 and 0.25, respectively. We set $T_{cls}$, $T_{cont}$, $T_{reg}$, and  $\tau$ as 0.7, 0.7, 0.5, and 0.07, respectively. We set the EMA momentum parameter to 0.9996, which determines the update rate of the teacher network. We provide training settings for each experiment in Tab.~\ref{tab:training setting}. As summarized in Tab.~\ref{tab:augmentation}, we use same augmentation strategies as the baseline~\cite{liu2021unbiased}.
\\

\noindent\textbf{Ablation study: Different localization quality measurements.}
We use the same settings except for settings with respect to localization quality measurements.
For box jittering~\cite{xu2021soft}, we first select pseudo-labels that have classification scores higher than the threshold~($T_{cls}$). To compute the prediction consistency of pseudo-labels, we generate ten randomly jittered bounding boxes for each pseudo-label and forward them to the ROI head to generate refined predictions. We compute the variance of each box's boundaries through refined predictions and normalize them using the height and width of each bounding box. We use the average of normalized variances as the uncertainty of the bounding box.
For predicted IoU~\cite{wang20213dioumatch}, we add a class-aware IoU branch with sigmoid activation parallel to the regression branch. We train the IoU branch with foreground region proposals in the same manner as the uncertainty branch. We compute ground-truth IoUs of region proposals with ground-truth labels and normalize values to [0.0, 1.0]. We use smoothL1 loss~\cite{girshick2015fast} for the training IoU branch and set IoU loss weight as 1.0.

\begin{table}[h]
\centering
\scriptsize
\begin{tabular}{ccccccc}
\toprule
\multirow{2}{*}{Training setting} & \multirow{2}{*}{VOC} & VOC & COCO & COCO & COCO \\
{} & {} & COCO20cls & standard & 35k & additional & {} \\
\toprule
Iteration for pretraining & 12k & 12k & 5k/20k/40k & 12k  & 90k  \\
\midrule
Iteration for training & 60k & 90k & 180k & 180k  & 360k  \\
\midrule
Batch size for labeled data & 32 & 32 & 32 & 16  & 32  \\
\midrule
Batch size for unlabeled data & 32  & 32 & 32 & 16 & 32 \\
\midrule
Unsupervised loss weight~($\lambda_{unsup}$)  & 4 & 4  & 4  & 2  & 2 \\
\bottomrule
\end{tabular}%
\caption{Training settings for each experiment.}
\label{tab:training setting}
\end{table}

\begin{table}[h]
\centering
\tiny
\begin{tabular}{cccccc}
\toprule
\multicolumn{4}{c}{\textbf{Weak augmentation}} \\
\toprule
Process & Probability & Parameters & Details\\
\midrule
Horizontal Flip & 0.5 & - & - \\
\toprule
\toprule
\multicolumn{4}{c}{\textbf{Strong augmentation}} \\
\toprule
Process & Probability & Parameters & Details\\
\midrule
Horizontal Flip & 0.5 & - & - \\
\midrule
\multirow{4}{*}{Color jittering} & \multirow{4}{*}{0.8} & brightness = 0.4 & We uniformly select from [0.6, 1.4] for brightness factor. \\
{} & {} & contrast = 0.4 & We uniformly select from [0.6, 1.4] for contrast factor. \\
{} & {} & saturation = 0.4 & We uniformly select from [0.6, 1.4] for saturation factor. \\
{} & {} & hue = 0.1 & We uniformly select from [-0.1, 0.1] for hue factor. \\
\midrule
Grayscale & 0.2 & - & - \\
\midrule
GaussianBlur & 0.5 & (sigma\_x, sigma\_y)=(0.1, 2.0) & $\sigma_x$ and $\sigma_y$ for gaussian filter are set 0.1 and 0.2., respectively. \\
\midrule
Cutout 1 & 0.7 & scale=(0.05, 0.2), ratio=(0.3, 3.3) & Randomly selected rectangle regions are erased in an image. \\
\midrule
Cutout 2 & 0.5 & scale=(0.02, 0.2), ratio=(0.1, 6)  & Randomly selected rectangle regions are erased in an image. \\
\midrule
Cutout 3 & 0.3 & scale=(0.02, 0.2), ratio=(0.05, 8)  & Randomly selected rectangle regions are erased in an image. \\
\bottomrule
\end{tabular}%
\caption{Details of augmentation strategies.}
\label{tab:augmentation}
\end{table}

\end{document}